\documentclass[review]{elsarticle}
\usepackage[margin=2.9cm]{geometry}

\usepackage{lipsum}                     
\usepackage{xargs}                      

\usepackage{pdflscape}
\usepackage{url}
\usepackage{verbatim}

\usepackage{graphicx}
\usepackage{subfig}
\usepackage{refcount}
\usepackage{footnote}
\usepackage{multirow}
\usepackage{listings}
\usepackage{multicol}
\usepackage{framed}
\usepackage{listings}
\usepackage[skipbelow=\topskip,skipabove=\topskip]{mdframed}
\mdfsetup{roundcorner=0}
\usepackage{etoolbox}
\usepackage{amssymb}
\usepackage{eurosym}
\usepackage{pifont}
\usepackage{algorithmicx}
\usepackage{algorithm}
\usepackage{algpseudocode}
\usepackage{afterpage}
\usepackage[table,xcdraw]{xcolor}

\usepackage{textcomp}
\usepackage{mathtools}
\usepackage{fixltx2e}
\usepackage{textcomp}
\usepackage[normalem]{ulem}
\useunder{\uline}{\ul}{}
\usepackage{amsmath,xparse,mleftright}

\usepackage[colorlinks = true,
            linkcolor = blue,
            urlcolor  = blue,
            citecolor = blue,
            anchorcolor = blue]{hyperref}

\algnewcommand\algorithmicinput{\textbf{Input:}}
\algnewcommand\INPUT{\item[\algorithmicinput]}

\algnewcommand\algorithmicoutput{\textbf{Output:}}
\algnewcommand\OUTPUT{\item[\algorithmicoutput]}

\NewDocumentCommand{\up}{som}{%
  \IfBooleanTF{#1}
    {\upext{#3}}
    {#3\IfNoValueTF{#2}{\mathord}{#2}\uparrow}%
}

\NewDocumentCommand{\down}{som}{%
  \IfBooleanTF{#1}
    {\upext{#3}}
    {#3\IfNoValueTF{#2}{\mathord}{#2}\downarrow}%
}

\usepackage{lineno,hyperref}
\modulolinenumbers[5]
\journal{Expert Systems with Applications}

\biboptions{authoryear}

\usepackage{lipsum}
\makeatletter
\def\ps@pprintTitle{%
\let\@oddhead\@empty
\let\@evenhead\@empty
\def\@oddfoot{}%
\let\@evenfoot\@oddfoot}
\makeatother

\begin{document}

\begin{frontmatter}


\title{On Taking Advantage of Opportunistic Meta-knowledge to Reduce Configuration Spaces for Automated Machine Learning}

\author[mymainaddress]{David Jacob Kedziora}
\ead{David.Kedziora@uts.edu.au}

\author[mymainaddress]{Tien-Dung Nguyen \corref{mycorrespondingauthor}}
\cortext[mycorrespondingauthor]{Corresponding author}
\ead{tiendung.nguyen-2@student.uts.edu.au}
\address[mymainaddress]{Complex Adaptive Systems Lab, University of Technology Sydney, Australia}

\author[mymainaddress]{Katarzyna Musial}
\ead{katarzyna.musial-gabrys@uts.edu.au}

\author[mymainaddress]{Bogdan Gabrys}
\ead{bogdan.gabrys@uts.edu.au}







\begin{abstract}

The optimisation of a machine learning (ML) solution is a core research problem in the field of automated machine learning (AutoML).
This process can require searching through complex configuration spaces of not only ML components and their hyperparameters but also ways of composing them together, i.e.~forming ML pipelines.
Optimisation efficiency and the model accuracy attainable for a fixed time budget suffer if this pipeline configuration space is excessively large.
A key research question is whether it is both possible and practical to preemptively avoid costly evaluations of poorly performing ML pipelines by leveraging their historical performance for various ML tasks, i.e.~meta-knowledge.
This paper approaches the research question by first formulating the problem of configuration space reduction in the context of AutoML.
Given a pool of available ML components, it then investigates whether previous experience can recommend the most promising subset to use as a configuration space when initiating a pipeline composition/optimisation process for a new ML problem, i.e.~running AutoML on a new dataset.
Specifically, we conduct experiments to explore (1) what size the reduced search space should be and (2) which strategy to use when recommending the most promising subset.
The previous experience comes in the form of classifier/regressor accuracy rankings derived from either (1) a substantial but non-exhaustive number of pipeline evaluations made during historical AutoML runs, i.e.~`opportunistic' meta-knowledge, or (2) comprehensive cross-validated evaluations of classifiers/regressors with default hyperparameters, i.e.~`systematic' meta-knowledge.
Overall, numerous experiments with the AutoWeka4MCPS package, including ones leveraging similarities between datasets via the relative landmarking method, suggest that (1) opportunistic/systematic meta-knowledge can improve ML outcomes, typically in line with how relevant that meta-knowledge is, and (2) configuration-space culling is optimal when it is neither too conservative nor too radical.
However, the utility and impact of meta-knowledge depend critically on numerous facets of its generation and exploitation, warranting extensive analysis; these are often overlooked/underappreciated within AutoML and meta-learning literature.
In particular, we observe strong sensitivity to the `challenge' of a dataset, i.e.~whether specificity in choosing a predictor leads to significantly better performance.
Ultimately, identifying `difficult' datasets, thus defined, is crucial to both generating informative meta-knowledge bases and understanding optimal search-space reduction strategies.

\end{abstract}


\begin{keyword}
AutoML \sep
Automated Machine Learning \sep
Opportunistic Meta-knowledge \sep
Search Space Reduction \sep
Configuration Space Reduction \sep
ML Pipeline Composition and Optimisation
\end{keyword}

\end{frontmatter}




\section{Introduction}
\label{sec:chap4_intro}

Many high-level processes go into running a machine learning (ML) application, the full automation of which was first recognised and discussed over a decade ago~\citep{kaga09} with the proposal of an architecture for developing, deploying and continuously adapting potentially complex ML models.
The field of automated machine learning (AutoML), as it has come to be known, has cemented itself in recent years with the intent to further mechanise these operations~\citep{va19,huko19,kemu20,zohu21}.
One of the most important targets of AutoML is ML pipeline composition and optimisation (PCO), which seeks to sequentially combine ML components with tunable hyperparameters into valid and well-performing solutions for an ML problem, i.e.~a dataset~\citep{kaga09,sabu16b,sabu17b,giya18,sabu18,va19,huko19,zohu19,ngma20,kemu20,zohu21}.
An ML component can be considered a data transformation with, if relevant, an associated algorithm that adjusts parameters during ML model training.
Accordingly, ML components encompass classification predictors and other preprocessing operators, e.g.~imputation or feature generation/selection.
Whatever ends up available to an AutoML package, PCO cannot proceed without first defining a searchable configuration space containing the ML components, their hyperparameters, and all the possible ways to connect them. 
Among general-purpose ML PCO methods that are relatively recent, one of the most successful is built into AutoWeka for Multi-Component Prediction Systems (AutoWeka4MCPS)~\citep{sabu17b}, which uses a Sequential Model-based Algorithm Configuration (SMAC) optimisation approach~\citep{thhu13}. 
The main idea of SMAC is to combine two processes: (1) randomly evaluate ML pipelines in the configuration space, i.e.~exploration, and (2) evaluate ML pipelines that are configured similarly to well-performing pipelines from the exploration step, i.e.~exploitation.

While using all available ML components to construct configuration spaces allows a broader range of diverse and possibly better ML solutions to be explored, a significant drawback is the waste of time evaluating poor-quality ML pipelines during AutoML PCO and, crucially, significant deterioration of performance for general-purpose optimisers, like SMAC, in finding effective solutions in large search/configuration spaces~\citep{sabu18}.
Therefore, we pose the following \textbf{research questions: given a fixed amount of time, can a smaller configuration space help AutoML PCO find better ML pipelines?
How can we best select suitable ML components for such reduced configuration spaces?}

Previous studies aimed at reducing configuration space can be divided into two approaches.
The first set of approaches uses expert knowledge to select ML components, hyperparameters and pipeline structures that are most likely to yield good ML pipelines~\citep{sabu18,fekl15,depi17,tsga12,kaga09}. 
The second approach is to constrain configuration spaces using meta-learning techniques ~\citep{olmo16,wemo18,giya18,dbbr18,va19,lebu15,albu15,rihu18,prbo19,wemu20}. 
We recently illustrated the feasibility of the second approach for an `opportunistic' meta-knowledge base (\textit{automl-meta}) derived from prior evaluations of AutoML PCO processes~\citep{ngke21}.
Although its `unplanned' nature produced many statistical drawbacks, as expected, \textit{automl-meta} remained surprisingly useful for informing configuration-space culling strategies.
We drew several conclusions in that work.
First, in general, the best-performing components for one dataset are a good guide for what to include in the search space for a `similar' dataset, although, in the absence of similarity metrics, it is also reasonable to look at the best performers overall.
Second, spending all the PCO time on a configuration space with only one predictor can often be the best choice, but if it misses, then it misses hard.
Third, as a corollary to the second conclusion, the safest bet is to reduce the search space to a handful of predictors, mitigating risk while profiting from the search efficiency boost.
However, statistical variance made it challenging to draw more robust conclusions despite these results.
Essentially, there was experimental scope to further tease out the signal from the noise concerning search-space reduction strategies.

Accordingly, this work is a significant extension of our conference paper~\citep{ngke21}, which conducted the preliminary analysis of opportunistic meta-knowledge for search-space reduction.
Naturally, certain core aspects of the research remain the same.
For instance, while there is future scope to be particularly surgical with pipelines, this work fixates on inclusion/exclusion for pipeline-ending predictors alone, pruning complex multi-dimensional configuration spaces in a blunt but straightforward manner.
Therefore, we continue to assess strategies driven by meta-knowledge that cull search spaces via recommending predictors they perceive as best-performing.
These strategies use different principles, with some leveraging dataset similarity calculated by, in this case, the relative landmarking method~\citep{va19}.
All reduction strategies are evaluated according to the mean error rates of the best ML pipelines found once PCO is applied to their recommended search spaces.
As for the foundational meta-knowledge itself, this takes the form of mean-error statistics and associated performance rankings for 30 Weka predictors, these being compiled from evaluations across 20 datasets; the rankings are both overall and per dataset.
Both the initial compilations, as well as the subsequent reduced-space searches, are performed with the AutoWeka4MCPS package\footnote{https://github.com/UTS-CASLab/autoweka}~\citep{sabu18}, which is accelerated by the ML-pipeline validity checker, AVATAR~\citep{ngma20}, wherever specified.

However, around this core, there are several novel expansions of knowledge.
First of all, we mathematically formalise the extension of the pipeline-inclusive `combined algorithm selection and hyperparameter optimisation' (CASH) problem~\citep{sabu18} to constrained configuration spaces.
This formalism anchors the current investigation in concrete theory, laying the groundwork for more sophisticated search-space control strategies in the future, i.e.~dynamic ones.

Next, when considering the statistical variance encountered in the previous work~\citep{ngke21}, a research question arises: how much of this noise is due to the haphazard evaluations involved in acquiring `opportunistic' meta-knowledge, i.e. \textit{automl-meta}?
After all, many features of this meta-knowledge base contribute to the unpredictability of predictor rankings, e.g.~loose averaging assumptions to account for multi-component pipelines.
Perhaps it is better to instead curate meta-knowledge ahead of time in a `systematic' fashion, where, for every dataset in \textit{automl-meta}, every available predictor is evaluated in single-component fashion the same number of times, e.g.~using a 10-fold cross-validation (CV) strategy for a dataset.
Such a curation would not be exhaustive, as a justified comparison requires the two meta-knowledge bases to be compiled within times that have a similar order of magnitude, but the systematic meta-knowledge base would at least be much more regular.
Thus, another significant addition in this work involves contrasting the impact of \textit{automl-meta} on search-space reduction strategies with that of a systematic meta-knowledge base (\textit{default-meta}), where every predictor is fully cross-validated but only for a single default set of hyperparameters.
Both \textit{automl-meta} and \textit{default-meta} have respective advantages and disadvantages, so this work essentially assesses -- this is a presumptive hypothesis -- whether the careful curation of meta-knowledge outweighs sheer opportunism, given comparable times for collation.

There is yet another crucial angle to consider regarding statistical variance in the performance of search-space reduction strategies, namely the impact of individual datasets.
This impact is pertinent to two distinct phases: (1) when evaluating and ranking predictors to form a meta-knowledge base, and (2) when determining the impact of a reduction strategy and assessing whether it actually helps find a better ML solution for a dataset.
One limitation within the previous work~\citep{ngke21} was the lack of consideration for individual datasets and their particular characteristics, except when determining dataset similarity.
The 20 datasets chosen for the investigation, reused here, were selected due to their diverse representative nature for benchmarking in earlier research~\citep{fekl15,sabu18}.
As such, results were aggregated across the entire set, hoping to identify reduction strategies and ways to drive them that performed universally well.
In contrast, this work presents a substantial investigation into the nature of individual datasets, concluding that their `challenge' impacts how suitable they are for use as meta-knowledge and how they should be solved via AutoML PCO processes.
Specifically, a dataset of minimal challenge, where any applied predictor performs equally and unexceptionally well, will provide predictor rankings that are extremely sensitive to stochastic factors.
Any search-space reduction strategy should be acceptable to \textit{solve} such an `easy' dataset, but their inclusion for meta-knowledge is up for debate.
Without mitigating approaches, these datasets may, via averaging processes, severely dilute the power of a meta-knowledge base.
This is likely to impact conclusions regarding the effectiveness, and other characteristics, of the investigated configuration-space reduction strategies.

Ultimately, this paper has five main contributions:

\begin{itemize}

\item Mathematically formulating the problem of ML pipeline composition and optimisation with constrained configuration spaces. This formulation aims to define the control of configuration space reduction.
With this theoretical backing, we can then use collated \textit{automl-meta} and \textit{default-meta} knowledge bases to generate reduced configuration spaces.
These reduced spaces are used as inputs to AutoML pipeline composition/optimisation processes to demonstrate and evaluate the usefulness of search-space reduction.

\item Exploring the characteristics of ML problems and how these impact both the quality of input meta-knowledge and the final performance of search-space reduction strategies.
The motivation here is the existence of `easy' problems, where most predictors deliver nearly identical performance, stochastically distorting aggregate predictor rankings.
In contrast, `difficult' problems, where only a few predictors statistically deliver better performance than others, are much more helpful at discriminating between the performance of ML components and ultimately testing the investigated configuration-space reduction strategies.
This exploration investigates whether easy and difficult problems, defined as such, can be identified and grouped effectively.

\item Critically assessing the advantages and disadvantages of constructing meta-knowledge bases in opportunistic and systematic fashion, i.e.~\textit{automl-meta} and \textit{default-meta}, respectively.
We consider these assessments in terms of (1) their evaluation time and (2) their sampling coverage of configuration space, i.e.~hyperparameters, predictors and pipelines.

\item Exploring how the results of searching through a reduced configuration space vary under different modes of how that reduction was recommended, e.g.~according to the best predictors over all datasets versus the best predictors for the most similar dataset.

\item Investigating the underlying mechanism of how the performance of an AutoML composition/optimisation process is affected by varying levels of recommended pipeline search space reduction, i.e.~removing all but the `best' $k$ of 30 predictors, for variable $k$, from an ML-component pool.

\end{itemize}

This paper presents these contributions in six sections.
After the Introduction, Section~\ref{sec:chap4_related_work} reviews previous attempts to constrain configuration spaces in the context of AutoML, as well as relevant research grappling with the nuances of benchmarking and dataset characterisation.
Section~\ref{sec:chap4_problem_formulation} formulates the problem of AutoML PCO in reduced configuration spaces as an extension of the CASH problem, i.e.~a constrained optimisation problem.
Section~\ref{sec:chap4_methodology} details the methodology used in this study, i.e.~the different methods of meta-knowledge collation, the various strategies for leveraging this meta-knowledge, and the process of relative landmarking for the strategies that exploit dataset similarity.
Section~\ref{sec:chap4_experiment} presents experiments and analyses assessing whether meta-learned recommendations for culling configuration space are beneficial to the performance of PCO, with particular focus on how severe the reduction should be, what strategies should guide the reduction, what effect opportunistic/systematic meta-knowledge has, and how datasets of varying `challenge' influence the whole process.
The comprehensive nature of this section is necessitated by many AutoML and meta-learning investigations overlooking the nuanced impact of generating/exploiting meta-knowledge on the eventual outcomes.
Finally, Section~\ref{sec:chap4_conclusion} concludes this study.

\section{Related Work}
\label{sec:chap4_related_work}

The growing number of available ML methods with their often complex hyperparameters leads to a very rapid expansion, if not combinatorial explosion, of ML-pipeline configurations and associated search spaces.
Intelligent reduction of these configuration spaces enables ML PCO methods to find valid and well-performing ML pipelines faster within the typical constraints of execution environments and time budgets.
Here, we review two main approaches to reduce configuration spaces in the context of ML PCO.
We also consider the topic of benchmarking, given that the effectiveness of strategies proposed by this work is contingent on the appropriate performance evaluation of predictors.

\textit{Configuration-space reduction via expert knowledge:} 
This approach can be implemented via fixed pipeline templates~\citep{sabu18,fekl15,depi17,tsga12,kaga09} or ad-hoc specifications for constraining search spaces~\citep{olmo16,wemo18,giya18,dbbr18,va19,lebu15,albu15,rihu18,prbo19,wemu20} such as context-free grammars.
Moreover, specific ranges of hyperparameter values that contribute to well-performing pipelines are also predefined in these specifications.
The advantage of this approach is its simple nature, leveraging expert knowledge to reduce configuration spaces by directly restricting the length of ML pipelines, the pool of ML components, and their permissible orderings/arrangements.
However, the disadvantage of this approach is that expert bias might obscure strongly performing ML pipelines outside of the predefined templates.

\textit{Configuration-space reduction via meta-knowledge:} 
This approach reduces configuration spaces by using prior knowledge, compiled mechanically, to avoid wasting time with unpromising ML-solution candidates. 
Frequently, this involves assessing similarity between past and present ML problems/datasets to hone in on the most relevant meta-knowledge available~\citep{lega10,albu15,lebu15,dbbr18,va19}. 
Methods in this category often establish characteristics for datasets that are subsequently used in correlations.
A characteristic can be directly derived from the dataset as a meta-feature, e.g. the number of raw features or data instances. 
Alternatively, two datasets can be compared by the relative performance of landmarkers; see Section~\ref{sec:chap4_landmarkers}.
These landmarkers are ideally simple one-component pipelines, i.e. predictors, of varying types; they estimate the suitability of varying modelling approaches for a dataset.
For instance, the performance of a linear regressor theoretically quantifies whether an ML problem is linear.
An ML problem estimated to be nonlinear will likely not benefit from methods serving a linear dataset. 
In any case, the meta-learning approach can be used to reduce configuration spaces by selecting a number of well-performing ML components ~\citep{albu15,dbbr18,va19,lebu15} or important hyperparameters for tuning~\citep{albu15,rihu18,prbo19,wemu20}. 
For instance, both \textit{average ranking} and \textit{active testing} have previously been used to recommend ML solutions for new datasets~\citep{dbbr18}. 
However, these approaches have not been applied to AutoML yet.
Moreover, these studies limit their scopes by optimising predictors, not multi-component pipelines, and the optimisation method they use is grid search, held not to be as effective as SMAC~\citep{thhu13}. 
Other studies have investigated estimating the importance of hyperparameters~\citep{rihu18,prbo19,wemu20} from prior evaluations.
Specifically, some hyperparameters are more sensitive to perturbation than others; tuning them can contribute to proportionally higher variability in ML-algorithm performance, i.e. error rate. 
As an example, gamma and complexity variable C have been previously identified as the most critical hyperparameters for a support vector machine (SVM)~\citep{rihu18}.
Consequently, the results of these studies can be used to reduce configuration spaces by constraining less-important hyperparameters, either by making their search ranges less granular or outright fixing them as default values.  
This procedure frees up more time to seek the best values for important hyperparameters that have the highest impact on finding well-performing pipelines.
However, a disadvantage of these studies is that the importance of ML-component hyperparameters has only been studied on small sets of up to six algorithms.
This limitation reflects how time-consuming it is to sample hyperparameter space across all available algorithms properly. 

\textit{Benchmarking:}
This concept represents the process of evaluating and comparing algorithms according to fixed standards.
For instance, using common datasets to assess search-space reduction strategies in this work falls under the primary purview of benchmarking.
However, crucially, the construction of a meta-knowledge base in this work is also intrinsically related to the notion of benchmarking, i.e.~generating performance rankings for predictors.
Of course, often treated as a corollary of the no-free-lunch theorem~\citep{adal19}, the performance of a predictor ties closely to the context in which it is applied; without specific ML problems to discriminate between predictors, there is no apriori way to recommend one over another.
Nevertheless, the selection of datasets to benchmark algorithms is often only lightly considered.
Many works simply adopt previous benchmarking datasets to facilitate comparison~\citep{fekl15,olmo16,giva18, sabu18,ngga20}, such as the conference paper preceding this work~\citep{ngke21}.
This approach has its justification, but the nature of a benchmark must be considered more carefully when utilised for a practical meta-knowledge base.
In the literature, the most fundamental question is whether a dataset collection represents all the problems that an algorithm may encounter.
Some studies hope to attain good coverage of possible problem space by working with as many datasets as possible~\citep{zohu21}.
Other studies dive deeper, aiming to characterise the hardness of a dataset~\citep{muma18,loga19}. 
For instance, some previous work uses a fixed threshold to define the difficulty of a dataset, e.g. more than 50\% of algorithms have an error rate higher than 20\%~\citep{muma18}. 
Alternatively, other research classifies difficulty via a set of meta-features measuring linearity, dimensionality and class imbalance~\citep{loga19}.
All these approaches are valid, but their focus is predominantly on a final assessment and recommendation of ML algorithms.
In contrast, within the context of this work, while representativeness is valued during the reduction strategy evaluation phase, we also need to consider the \textit{usefulness} of benchmarking datasets as meta-knowledge.
Because all strategies in this investigation exploit predictor rankings, any datasets that complicate these rankings, e.g.~via sensitivity to stochastic variability, are an issue.
We thus introduce a new metric for `challenge' in this work, which inversely correlates with the number of predictors that are significantly better than all others on a dataset; see Section~\ref{sec:chap4_dataset_understand}.
Based on this metric, we ultimately consider the influence of `easy' datasets on search-space reduction.

\section{Problem Formulation}
\label{sec:chap4_problem_formulation}

To facilitate the control of configuration space reduction in the context of AutoML, we first present the definition of the ML PCO problem~\citep{sabu18}.
Instead of using whole and unchanged configuration spaces, we extend the original formalism to enable the reduction of configuration spaces at any point in time during an AutoML PCO process.
In effect, PCO becomes a constrained optimisation problem.
So, to detail the core problem that this research tackles, we recall the fundamentals of AutoML-based solution optimisation in Section~\ref{Sec:TheoryGeneral}, then we formalise the notion of configuration space in Section~\ref{Sec:TheoryConfig}, and we finally present equations for the constrained optimisation in Section~\ref{Sec:TheoryOptim}.

\subsection{General Background}
\label{Sec:TheoryGeneral}

Broadly stated, an ML model is a mathematical object that attempts to approximate a desired function.
It is typically paired with an ML algorithm that, via the process of training, feeds on encountered data to adjust certain variables, i.e. model parameters, and improve the accuracy of the approximation.
This pairing of ML model and algorithm -- we call this ML component a predictor in this work -- contains other variables, i.e. hyperparameters, that stay fixed throughout the training process.
Hyperparameter optimisation (HPO) is thus the process of finding values for these training constants that optimise the performance of the trained model, usually via some iterative approach.
Even at this level, the task is not trivial; hyperparameter space can involve many continuous or discrete dimensions with varying ranges.
When HPO extends to variable ML components, a core facet of AutoML, configuration space becomes even more complex, involving so-called `conditional' hyperparameters.
For instance, the polynomial degree of an SVM kernel is only non-null if the type of SVM kernel is set to polynomial.
Moreover, the incorporation of pipeline structure in AutoML search space complicates matters.
This extension involves broadening the notion of an ML component to include data preprocessors or even meta-methods that somehow augment a predictor, e.g.~an ensembling of a base learner.

Conceptually, ML components can be pruned from a configuration space to leave a substantially smaller `active' subspace.
Assuming that this subspace contains well-performing ML components, an optimiser like SMAC will exploit and explore this subspace more effectively and efficiently.
In other words, SMAC spends more of the allocated, often limited, time budget on HPO of well-performing ML components rather than exploring ML pipelines with bad-performing ML components.

\subsection{Configuration Space Modelling}
\label{Sec:TheoryConfig}

To discuss the search space of ML pipelines, it is worth considering that each potential constituent within a pipeline is an instance of an ML component with a specific set of hyperparameter values.
Specifically, if $\mathcal{A}$ is a pool of available ML components, a potential pipeline constituent indexed by $c$ and $r$ can be represented as
\begin{equation}
\small{
x_{c,r}=(\mathcal{A}_c,\lambda_{c,r}),
\label{eq:treenode}
}
\end{equation}
where $\mathcal{A}_c\in\mathcal{A}$ is an ML component and $\lambda_{c,r}$ is a set of hyperparameter values for component $\mathcal{A}_c$.
The $r$ index allows for two of the same component, e.g.~an SVM, to be distinguished by their hyperparameter values.

Given this space, an ML pipeline $p$, also called a configuration, can be represented as
\begin{equation}
\small{
p = (g,\vec{\mathcal{A}}, \vec{\mathcal{\lambda}}),
\label{eq:pipeline}
}
\end{equation}
where $\vec{\mathcal{A}}$ is a vector of ML components in $\mathcal{A}$, $\vec{\mathcal{\lambda}}$ is a vector of hyperparameter-value sets corresponding to $\vec{\mathcal{A}}$, and $g$ is a sequential pipeline structure that defines how the components are connected.
We additionally define configuration space $\mathcal{T}$ to contain all possible ML pipelines $p$.
This configuration space can be reduced in many ways, e.g.~removing individual instances of $p$.
However, in the context of this work, removing an entire component $\mathcal{A}_c$ from a pool makes a much more dramatic reduction of $\mathcal{T}$.
Of course, the danger for this smaller search space is that the removal of $\mathcal{A}_c$ might have eliminated an optimal pipeline $p$ that contains $\mathcal{A}_c$ within $\vec{\mathcal{A}}$.

\subsection{The Constrained Optimisation Problem}
\label{Sec:TheoryOptim}

The PCO process aims to find the best-performing ML pipeline $(g,\vec{\mathcal{A}}, \vec{\mathcal{\lambda}})^{*}$, which involves the optimal combination of pipeline structure, selected components, and associated hyperparameters.
The pipeline-search problem~\citep{sabu18} can thus be written as
\begin{equation}
\small{
(g,\vec{\mathcal{A}}, \vec{\mathcal{\lambda}})^{*} = \mathrm{arg\,min}\frac{1}{j}\sum_{i=1}^{j} \mathcal{L}((g,\vec{\mathcal{A}},\vec{\mathcal{\lambda}}),\mathcal{D}^{(i)}_{train},\mathcal{D}^{(i)}_{valid}),
}
\label{eq:pipeline_problem}
\end{equation}
where $\mathcal{D}_{train}$ and $\mathcal{D}_{valid}$ are training and validation datasets.
Specifically, this equation minimises the k-fold CV error -- we index by $j$ to avoid confusion with another $k$ variable in this paper -- of loss function $\mathcal{L}$.

Importantly, standard PCO does not consider changes to the configuration space $\mathcal{T}$.
If we reduce this configuration space by selecting promising ML components to form the configuration subspace $\mathcal{T}^{*}\subset \mathcal{T}$, the problem of PCO can be reformulated as a constrained version of Eq.~(\ref{eq:pipeline_problem}), as follows:  
\begin{equation}
\small{
(g,\vec{\mathcal{A}}, \vec{\mathcal{\lambda}})^{*} = \mathrm{arg\,min}\frac{1}{j}\sum_{i=1}^{j} \mathcal{L}((g,\vec{\mathcal{A}}, \vec{\mathcal{\lambda}}) \in \mathcal{T}^{*},\mathcal{D}^{(i)}_{train},\mathcal{D}^{(i)}_{valid}).
}
\label{eq:extended_pipeline_problem}
\end{equation}
In this work, we define a function,
\begin{equation}
\small{
h(\mathcal{T}) = \mathcal{T}^{*},
}
\label{Eq:Strategy}
\end{equation}
that is responsible for passing the subset of a configuration space to Eq.~(\ref{eq:extended_pipeline_problem}).
Specifically, $h$ represents a culling strategy based on meta-learning; see Section~\ref{Sec:Strategies}. 
As such, it is dependent on the meta-knowledge base that feeds the strategy, which is itself dependent on the curation method, i.e.~opportunistic versus systematic.

In principle, subspace control, i.e.~$h$, can be applied at varying times:
\begin{itemize}
    \item \textit{Initialisation phase of AutoML}.
    Meta-knowledge about previously solved problems can be used to constrain the configuration space ahead of applying PCO. 
    This approach is the focus of this work, and the concrete culling strategies used, i.e.~$h$, are detailed in Section~\ref{Sec:Strategies}.
    \item \textit{Operational phase of AutoML}.
    Information can be acquired during an AutoML run that may recommend refining configuration space.
    This acquisition can be externally sourced, e.g.~if numerous optimisations are occurring on a cloud-based AutoML, a meta-knowledge base may be updated with pertinent information to the problem at hand.
    However, any dynamically acquired information is more realistically going to arise internally, i.e.~from evaluating pipelines for the problem at hand.
    This development could potentially improve the understanding of where the new problem sits with respect to existing meta-knowledge, thus improving similarity-based strategies for subspace control; see Section~\ref{sec:chap4_landmarkers}.
    That stated, we do not focus on dynamic subspace control in this work, other than via the use of AVATAR~\citep{ngga20}, which leverages a one-off compilation of expert knowledge to discard invalid pipelines on the fly.
\end{itemize}

\section{Meta-learning Methodology for Configuration Space Reduction}
\label{sec:chap4_methodology}

At its most basic, the core goal of this work is to find whether a search-space reduction strategy driven by meta-learning, i.e.~the $h$ function in Eq.~(\ref{Eq:Strategy}), could significantly boost the outcomes of AutoML, i.e.~the minimisation of loss function $\mathcal{L}$ in Eq.~(\ref{eq:extended_pipeline_problem}), within a fixed amount of optimisation time.
Accordingly, we describe the methodology that supports our investigation into this topic.
Section \ref{sec:chap4_meta_construction} begins by detailing the construction of meta-knowledge bases, highlighting the opportunistic versus systematic approach.
Section \ref{Sec:Strategies} then describes all the configuration-space reduction strategies employed in this work, elaborating how they leverage the compiled meta-knowledge bases.
Finally, Section~\ref{sec:chap4_landmarkers} covers the specifics of relative landmarking, which is used by some of the strategies to identify similar datasets.

\subsection{The Meta-knowledge Base}
\label{sec:chap4_meta_construction}

Meta-knowledge is arguably most appealing when acquired without undue effort, perhaps as the side-effect of some other process.
Accordingly, being able to reuse information from past AutoML runs is a core inspiration to this work and its antecedent~\citep{ngke21}.
This research aims to assess the utility of opportunistic meta-knowledge for configuration-space reduction.
However, this time we attempt to ablate one critical factor that may contribute to statistical variance for the performance of search-space culling strategies: the haphazard nature of opportunistic evaluations.
In particular, we introduce another meta-knowledge base curated from systematic evaluations to see if its regularity proves more beneficial to the reduction strategies we employ.

\begin{figure*}[h] 
    \centering
    \includegraphics[width=0.75\linewidth]{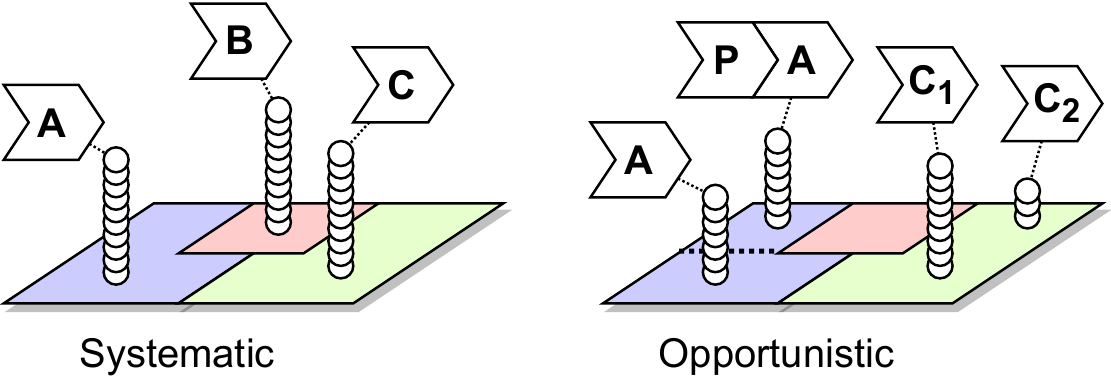}
    \vspace*{-1mm}
     \caption{Schematic for how a configuration space is explored systematically versus opportunistically when generating meta-knowledge. Each circle represents an evaluation of a specific configuration (for predictor A, B or C) on one fold of a dataset. Predictor subscripts emphasise evaluations for different hyperparameter values, while component P represents a preprocessor. Further details in text.}
    \label{Fig:Config}
\end{figure*}

Both compilations are derived by using SMAC-based AutoWeka4MCPS~\citep{sabu18} to evaluate ML solutions involving 30 predictors on 20 datasets; see Section~\ref{sec:chap4_settings} for specifics on the predictors/datasets.
However, Fig.~\ref{Fig:Config} captures the difference in how the evaluations are made for both the systematic and opportunistic meta-knowledge bases, named \textit{default-meta} and \textit{automl-meta}, respectively.
In all cases, there is one fixed maximal configuration space, $\mathcal{T}$, that can be partitioned into 20 subspaces according to the final component in an ML pipeline, i.e.~a predictor.
Technically, AutoWeka4MCPS applies a hard-coded limit of seven on the length of an ML pipeline, but, with ML components in the form of missing value replacement, outlier detection and removal, transformation, dimensionality reduction, sampling, ML prediction and more~\citep{sabu18}, $\mathcal{T}$ contains over 800 billion ML pipelines.

In the systematic case, the sampling is trivial.
Per dataset, each predictor is evaluated on ten folds of the dataset within the 10-fold CV scheme, represented by ten circles per predictor on the `systematic' side of Fig.~\ref{Fig:Config}.
However, they are only evaluated for their default hyperparameter values, as supplied by WEKA.
Other sections of configuration space, especially those that involve multi-component pipelines, are certainly not explored.
Unsurprisingly, there has been debate around the quality of default hyperparameters in the past.
Some have raised concerns about high-variance error due to their lack of mechanical optimisation~\citep{buga12}.
On the other hand, default hyperparameters have often been fine-tuned by algorithm designers, i.e.~expert knowledge, to be as generally functional as possible, so others assert that they show relatively good performance for a large number of problems~\citep{wemu20}.
Either way, the core benefit is that every predictor has been evaluated an equal number of times, allowing for clean statistical analyses and regular representation.
In fact, for this reason, \textit{default-meta} is also used in Section~\ref{sec:chap4_dataset_understand} to quantify the challenge of a dataset.
Of course, the time taken to curate \textit{default-meta} is dependent on the complexity of the predictors involved; in this work, it is not uncommon for the sampling of a dataset to take over ten hours.

In contrast, the opportunistic case assumes five two-hour AutoML runs were applied in the past for each dataset.
As Fig.~\ref{Fig:Config} indicates, the haphazard nature of these exploration/exploitation paths through $\mathcal{T}$ means that a hyperparameter configuration may not be cross-validated across all ten folds of a dataset.
Worse yet, a predictor, particularly a heavyweight one, e.g.~B in Fig.~\ref{Fig:Config}, may not be sampled at all within the time budget, meaning that an assumption must be made about its performance, e.g.~100\% loss.
However, in practice, the ML predictors supplied by WEKA are primarily lightweight, meaning that two-hour runs are often enough for hundreds of evaluations per predictor.
A priori, it is not clear whether this increase in effective sampling is overall positive or negative.
As the `opportunistic' side of Fig.~\ref{Fig:Config} shows, many more hyperparameter values can be assessed, including within regions of configuration space that involve multiple components.
This sampling can reveal particularly well-performing solutions that default hyperparameters may not access.
On the other hand, it is not easy to provide a concise metric for the performance of a predictor on a dataset.
In this work, we average the error across every single-fold evaluation that involves the predictor, exploiting the increased representation in sampling, but it can be debated how appropriate such aggregation is, especially for multi-component pipelines.

Ultimately, neither meta-knowledge base is exhaustive, and both have advantages and disadvantages.
The systematic \textit{default-meta} is equal-opportunity but limited, reliably so, while the opportunistic \textit{automl-meta} can acquire a lot more information, or a lot less of it, all subject to the erratic whims of a SMAC run.
Thus, by using both to drive search-space reduction strategies, we can examine the impact of opportunistic variability on PCO performance.

\subsection{Reduction Strategies}
\label{Sec:Strategies}

In this work, we investigate 33 strategies for defining a configuration space, i.e.~function $h$ in Eq.~(\ref{Eq:Strategy}).
The performance of each strategy is ultimately tested per dataset by, done five times over, examining the best solution found in the resulting configuration space by AutoML within two hours; see Section~\ref{sec:results}.
Three of these strategies can be considered as a scientific `control' and are listed as follows:
\begin{itemize}
    \item \textbf{baseline}:
    The complete configuration space $\mathcal{T}$ is untouched, constructed from all preprocessing and predictor components available to AutoWeka4MCPS~\citep{sabu18}.
    This strategy solves the traditional PCO problem in Eq.~(\ref{eq:pipeline_problem}).
    
    \item \textbf{avatar}:
    Virtually identical to \textbf{baseline}, but invalid pipelines are excluded from $\mathcal{T}$ on the fly by the AVATAR upgrade~\citep{ngma20}, thus avoiding unnecessary evaluations and accelerating the AutoML solution search.
    AVATAR is technically a dynamic form of search-space control but is driven by a static compilation of expert knowledge.
    
    \item \textbf{r30}:
    In an extreme case, the pipeline structure of an ML solution, per dataset, is fixed to the best that was found after a previous 30 hours of optimisation by AutoML~\citep{sabu18}.
    The two hours in this `continuation' experiment are solely dedicated to training the selected predictor/pipeline and optimising hyperparameters that have been re-initialised to their default values.
    Accordingly, this strategy, as a form of ablation analysis, simulates a \textbf{baseline} PCO where an ideal pipeline structure is immediately found, thus attempting to remove the influence of pipeline search on the statistical variance of results.
\end{itemize}

\begin{figure*}[h] 
    \centering
    \includegraphics[width=\linewidth]{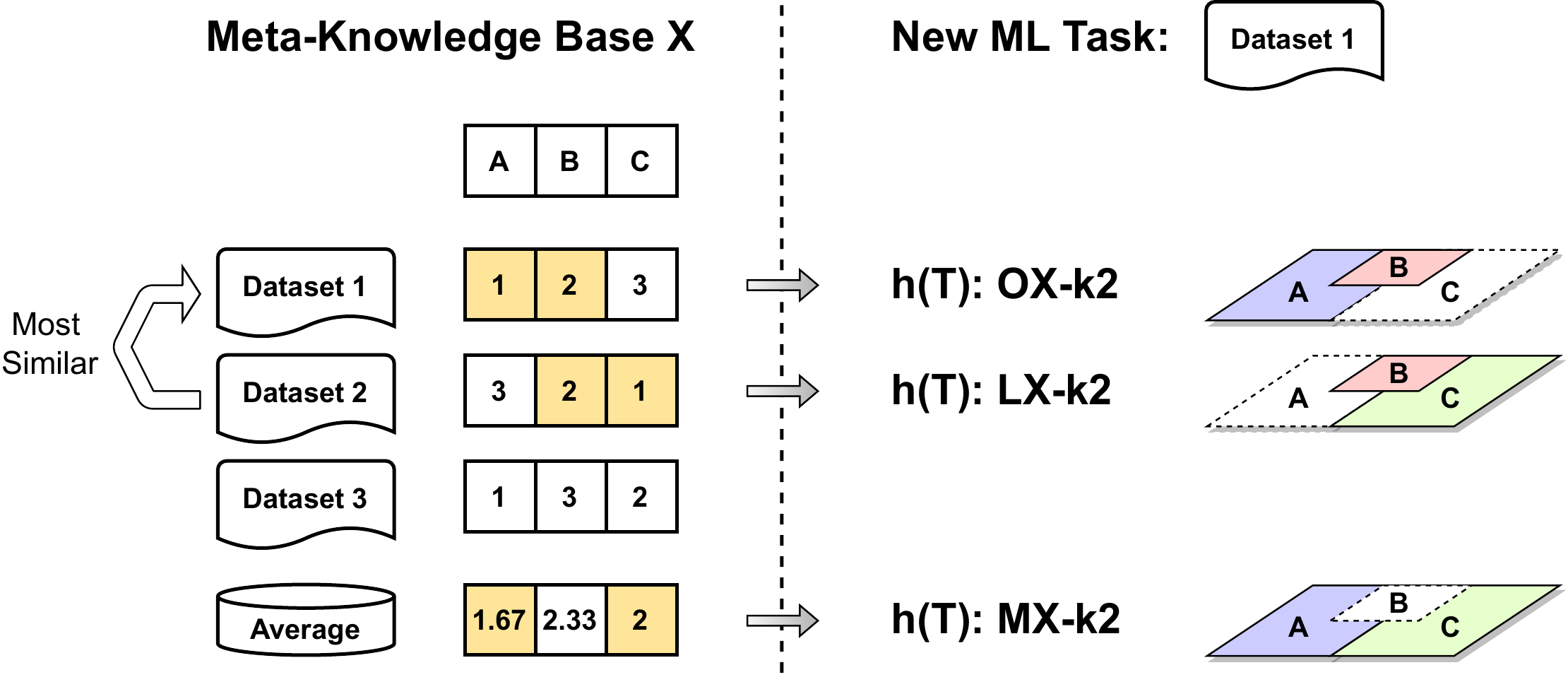}
    \vspace*{-1mm}
     \caption{Simple example for how meta-learning drives search-space reduction strategies. Meta-knowledge base X provides rankings for three predictors (A, B and C) per dataset and overall. A new AutoML run encounters a dataset and seeks a search space consisting of only two predictors, i.e.~$k=2$. Oracle strategy \textit{OX-k2} recommends the best two predictors previously evaluated on the same dataset. Landmarker strategy \textit{LX-k2} recommends the best two predictors previously evaluated on the most similar dataset. Global-leaderboard strategy \textit{MX-k2} recommends the best two predictors overall.}
    \label{Fig:Strategy}
\end{figure*}

The remaining 30 strategies are best understood with reference to Fig.~\ref{Fig:Strategy}.
First, in compiling a meta-knowledge base, as described in Section~\ref{sec:chap4_meta_construction}, one can calculate a mean-error metric for every predictor, per dataset, by averaging across all relevant single-fold evaluations.
Error rates are, of course, a poor way of comparing predictor performance between datasets, as the context of each ML problem may be radically different.
In contrast, rankings can capture the relative performance across predictors.
They can also be averaged across all datasets to indicate generally strong performers.
The left side of Fig.~\ref{Fig:Strategy} exemplifies the per-dataset and overall rankings that a meta-knowledge base, labelled $X$, may produce.
In this work, we examine two sources of rankings: \textit{automl-meta}, indexed by $X=1$, and \textit{default-meta}, indexed by $X=2$.

Now, when a strategy leverages a meta-knowledge base to reduce a configuration space, another choice must be made: how severe should the culling be?
Settings parameter $k$ tunes this, with $k=2$ exemplified in Fig.~\ref{Fig:Strategy}; each displayed strategy reduces the entire configuration space of three predictors to a subspace of two.
In this work, $\mathcal{T}$ contains 30 predictors, so, to capture a broader spread of choices and also for legacy reasons~\citep{ngke21}, we investigate five subspace sizes of $k$ in \{1,4,8,10,19\}.
Whatever the case, the idea behind this settings parameter is that a configuration space should consist of only the `best' $k$ predictors, as determined by a reduction strategy.

So, consider AutoML is encountering an ML problem to solve, i.e.~a dataset.
Which predictor rankings should be used when deciding on a configuration subspace?
Three possible approaches used in this work, all exemplified simply within Fig.~\ref{Fig:Strategy}, are listed as follows:
\begin{itemize}
    \item \textbf{oracle (\textit{O1-k1}, \textit{O1-k4}, \textit{O1-k8}, \textit{O1-k10}, \textit{O1-k19}, \textit{O2-k1}, \textit{O2-k4}, \textit{O2-k8}, \textit{O2-k10} and \textit{O2-k19})}:
    Given perfect meta-knowledge, the best predictors for a dataset should, as a truism, be the ones already evaluated as the best on that same dataset.
    Thus, this strategy recommends the best $k$ predictors found previously for the same dataset when defining a searchable subspace.
    This oracle is often unavailable in practice, with AutoML usually being applied to new datasets.
    Within this work, the oracle-based recommendations are treated as optimal in Section~\ref{Sec:EvalOracle}, against which other strategic recommendations are compared.
    The strategy is then actually applied within Section~\ref{sec:results}, identifying how reliable the `perfect meta-knowledge' assumption is in practice.

    \item \textbf{landmarked (\textit{L1-k1}, \textit{L1-k4}, \textit{L1-k8}, \textit{L1-k10}, \textit{L1-k19},\textit{L2-k1}, \textit{L2-k4}, \textit{L2-k8}, \textit{L2-k10}, \textit{L2-k19})}:
    It is often hypothesised that `similar' datasets should possess similar attributes, e.g.~how ML predictors behave on them.
    Thus, this strategy recommends the best $k$ predictors found previously for the most similar dataset when defining a searchable subspace.
    In this work, we use the relative landmarking method to quantify similarity; see Section~\ref{sec:chap4_landmarkers}.
    Notably, the two-hour PCO runtime for this strategy includes evaluating ML landmarking predictors and determining the most similar dataset to a problem on hand.

    \item \textbf{global leaderboard (\textit{M1-k1}, \textit{M1-k4}, \textit{M1-k8}, \textit{M1-k10}, \textit{M1-k19}, \textit{M2-k1}, \textit{M2-k4}, \textit{M2-k8}, \textit{M2-k10} and \textit{M2-k19})}:
    Although one may interpret the no-free-lunch theorem~\citep{adal19} to suggest that no algorithm is better than another overall, there is evidence that some ML predictors have a more robust baseline performance than others for practical ML problems, e.g.~for Kaggle competitions.
    Thus, this strategy recommends the best $k$ predictors found previously, according to rankings averaged over all datasets, when defining a searchable subspace.
\end{itemize}

In summary, the 30 non-control strategies we investigate for function $h$ in Eq.~(\ref{Eq:Strategy}) are split by three strategic approaches, two meta-knowledge bases (e.g.~O1 and O2 for \textit{automl-meta} and \textit{default-meta}, respectively), and five values for subspace size $k$.
All 30 of these strategies employ AVATAR.
Naively, for $k=n$ and with $E(x)$ representing the CV error of the optimal ML pipeline found for approach $x$, we could expect the following for the full set of 33 strategies:
\begin{eqnarray}
\label{Eq:Hierarchy}
    \nonumber && E(\mathrm{baseline})>E(\mathrm{avatar}),\\
    \nonumber && E(\mathrm{baseline})>E(\mathrm{r30}),\\
    && E(\mathrm{avatar})>E(\operatorname{MX-kn})>E(\operatorname{LX-kn})>E(\operatorname{OX-kn}).
\end{eqnarray}
Essentially, the AVATAR upgrade should boost search efficiency and outperform \textit{baseline}, while \textit{r30} theoretically performs even better by having a previously optimised pipeline structure.
As for the order of strategies driven by meta-learning, one can assume that they improve based on the increasing relevance of the meta-knowledge to a problem at hand.
Previous results loosely support this ordering~\citep{ngke21}.
Of course, many effects can contribute to statistical variance and overturn the expected ordering in Eq.~(\ref{Eq:Hierarchy}).
The variability of AutoML PCO with SMAC is a significant contributor, as are the time constraints that remained fixed, regardless of whether a dataset is small or large.
Additionally, the landmarked strategies are especially susceptible to how well dataset similarity can be quantified.

However, a facet largely ignored in previous work is the danger of using rankings as a metric for predictor quality.
For instance, consider Fig.~\ref{Fig:Strategy}, where the example meta-knowledge base has been constructed so that each of the three depicted strategies recommends a different configuration subspace.
It is not difficult to imagine that simple modifications in the predictor rankings can cause drastic changes to strategic recommendations.
We hypothesise that the `challenge' of an individual dataset is essential in determining how robust these rankings are.
For instance, the difference between the best and second-best predictor can be dramatic for a `difficult' dataset, while the gap between best and twentieth-best predictor on an `easy' dataset could be so small that stochastic perturbation reorders the rankings.
In light of this, Section~\ref{sec:chap4_dataset_understand} dives deep into how the performance of all 30 predictors is distributed for individual datasets.
Section \ref{sec:results} then considers the implications of these easy/difficult datasets on both the recommendations of reduction strategies and their evaluated performance.

\subsection{Landmarkers}
\label{sec:chap4_landmarkers}

Typical reasoning in the field of meta-learning is that previous experience is most relevant to a problem at hand if past and present contexts are similar.
Accordingly, it is routine to approach this by defining and compiling a set of so-called meta-features to describe a dataset, which is then subsequently compared between datasets~\citep{albu15,lebu15}.
Naturally, identifying the most appropriate metrics to denote this similarity is a topic of active research, but landmarking has proved to be a popular option~\citep{va19}; we employ this procedure in relevant experiments.

A set of landmarkers, $\Theta=\{\theta_i\}$, is generally a collection of ML predictors that are simple and efficient to execute.
Ideally, they represent a diversity of problem types.
The theory is that if a landmarker is well-suited for a problem type $A$, and it produces an ML model with solid performance, e.g.~good classification accuracy, on dataset $B$, then dataset $B$ belongs to the class of problems designated by $A$.
Any ML pipeline that works well for one dataset in class $A$ is then presumed to work well for any other of that same problem type.

However, in practice, it is challenging to pick a perfect set of landmarkers, especially as the choice of meta-features to describe complex problems has an impact on the effectiveness of similarity-based meta-learning \citep{va19,albu15}.
Given that we include the evaluation of landmarkers as part of the overall AutoML optimisation time within relevant experiments, we have decided in this study to prioritise fast execution time.
Therefore, noting the average evaluation time of all predictor-containing pipelines in \textit{automl-meta}, we select the following five fastest predictors (amongst the 30 listed in Section~\ref{sec:chap4_settings}) for our set of landmarkers: RandomTree, ZeroR, IBk, NaiveBayes, and OneR. 
We acknowledge that this choice is relatively crude, but it adheres to the opportunistic principles behind this study; are rough metrics for dataset similarity still helpful in providing additional intelligence when reducing the input search space for AutoML pipeline selection?

\begin{algorithm}[!htbp]
\small
\caption{\small Designing Configuration Spaces with Relative Landmarking}
\begin{algorithmic}[1]
\Require
      \Statex $\Theta$: The set of landmarkers
      \Statex $t_{new}$: The new dataset
      \Statex \{$t_{prior,j}$\}: The set of prior datasets
      \Statex $k$: The number of selected ML components
\Statex      
\For{$\theta_i$ \textbf{in}  $\Theta$}
        \State $E_{new,i}$ = \textit{evaluate}($\theta_i$, $t_{new}$)
\EndFor
\For{\textbf{each}  $t_{prior,j}$}
        \State $c_j$ = 
        \textit{calculateCorrelation}($E_{new}$, $E_{prior,j}$)
\EndFor
\State $t^{*}$ = \textit{getMostSimilarTask}($c$) 
\State $\mathcal{T}_{new}$ = \textit{selectKBestMLComponents}($t^{*}$, $k$)
\State \textbf{return} $\mathcal{T}_{new}$
\end{algorithmic}
\label{algorithm:construct_configuration_space}
\end{algorithm} 

Algorithm \ref{algorithm:construct_configuration_space} formalises how configuration space is constrained via the relative landmarking method, to then be used as input for AutoML pipeline composition and optimisation methods.
Essentially, it is a recipe for \textit{LX-kn} in Section~\ref{Sec:Strategies}.
First, the algorithm evaluates a new dataset $t_{new}$ with each landmarker $\theta_i$, resulting in a 10-fold CV error rate, $E_{new,i}$, per landmarker (lines 1-3).
Second, the algorithm calculates a Pearson correlation coefficient between the full performance vector of the new dataset, $E_{new}$, and a similarly landmarked vector of mean error rates, $E_{prior,j}$, for each prior dataset $t_{prior,j}$ (lines 4-6).
Third, the algorithm ranks the correlation coefficients and selects the dataset, $t^{*}$, that has the highest correlation coefficient (line 7).
Finally, the resulting configuration space to explore is constructed from all preprocessing components and the top $k$ best-performing predictors (line 8) for the most similar dataset, $t^{*}$.
Again, we emphasise that, for \textit{LX-kn} applied to a newly encountered dataset, the net evaluation time of landmarkers is deducted from the total time budget assigned to ML PCO processes.

\section{Experiments}
\label{sec:chap4_experiment}

Here we present the results from a suite of experiments -- associated code and data are available online\footnote{https://github.com/UTS-CASLab/autoweka} -- investigating the utility of meta-knowledge for AutoML search-space reduction.
The general experimental context, including the details of datasets and predictors, is presented first in Section~\ref{sec:chap4_settings}.
Phase one of the actual investigation begins by seeking a greater understanding of the meta-knowledge available.
Section \ref{sec:chap4_exp_metaknowledge} provides a general sense of how opportunistic AutoML-driven evaluations of ML predictors compare with default hyperparameters, i.e.~\textit{automl-meta} versus \textit{default-meta}.
Section \ref{sec:chap4_dataset_understand} dives deeper into individual datasets, assessing whether it is possible to draw distinctions between them according to `challenge', i.e.~the way that the aggregated performance of each applied ML predictor is distributed.
Phase two of the investigation then turns to the performance of search-space reduction strategies that leverage this meta-knowledge.
Section \ref{Sec:EvalOracle} considers, in the absence of AutoML PCO, how the recommendations of different strategies compare, assuming perfect meta-knowledge and, consequently, that oracle-based rankings reflect the ground truth for ML predictor performance.
Finally, Section~\ref{sec:results} evaluates every strategy via a PCO run and then discusses their actual performance concerning strategic approach, subspace size, the opportunistic/systematic quality of meta-knowledge, and the easy/difficult nature of datasets.

\subsection{Experimental Context}
\label{sec:chap4_settings}

We run all experiments on AWS EC2 \textit{t3.medium} virtual machines.
Each virtual machine has two vCPUs and 4 GB of memory. 
The experiments take the form of running AutoWeka4MCPS, an AutoML tool, which applies ML PCO with an implementation of SMAC~\citep{sabu18}.
During the compilation of meta-knowledge and final evaluation of reduction strategies, the AutoWeka4MCPS package runs for all 20 datasets listed in Table~\ref{tab:datasets}.  
These datasets are chosen as they were used in previous studies \citep{fekl15,olmo16,sabu18}, although their suitability as meta-knowledge for search-space reduction is assessed in both Section~\ref{sec:chap4_dataset_understand} and Section~\ref{sec:results}.

\begin{table}[htb!]
\centering
\caption {Dataset characteristics: the number of numeric attributes, nominal attributes, the number of distinct classes, instances in training and testing sets.}
\label{tab:datasets}
\fontsize{7.2}{8.5}\selectfont
\begin{tabular}{|l|l|l|l|l|l|}
\hline
\textbf{Dataset}  & \textbf{Numeric} & \textbf{Nominal} & \textbf{Distinct classes} & \textbf{Train} & \textbf{Test} \\ \hline
abalone           & 7                & 1                & 28                              & 2,924          & 1,253         \\ \hline
adult             & 6                & 8                & 2                               & 32,561         & 16,281        \\ \hline
amazon            & 10,000           & 0                & 50                              & 1,050          & 450           \\ \hline
car               & 0                & 6                & 4                               & 1,210          & 518           \\ \hline
cifar10small      & 3,072            & 0                & 10                              & 10,000         & 10,000        \\ \hline
convex            & 784              & 0                & 2                               & 8,000          & 50,000        \\ \hline
dexter            & 20,000           & 0                & 2                               & 420            & 180           \\ \hline
dorothea          & 100,000          & 0                & 2                               & 805            & 345           \\ \hline
gcredit           & 7                & 13               & 2                               & 700            & 300           \\ \hline
gisette           & 5,000            & 0                & 2                               & 4,900          & 2,100         \\ \hline
kddcup & 192              & 38               & 2                               & 35,000         & 15,000        \\ \hline
krvskp            & 0                & 36               & 2                               & 2,238          & 958           \\ \hline
madelon           & 500              & 0                & 2                               & 1,820          & 780           \\ \hline
mnist             & 784              & 0                & 10                              & 12,000         & 50,000        \\ \hline
secom             & 590              & 0                & 2                               & 1,097          & 470           \\ \hline
semeion           & 256              & 0                & 10                              & 1,116          & 477           \\ \hline
shuttle           & 9                & 0                & 7                               & 43,500         & 14,500        \\ \hline
waveform          & 40               & 0                & 3                               & 3,500          & 1,500         \\ \hline
winequality       & 11               & 0                & 11                              & 3,429          & 1,469         \\ \hline
yeast             & 8                & 0                & 10                              & 1,039          & 445           \\ \hline
\end{tabular}

\end{table}

\begin{table}[htb!]
\centering
\caption {List of predictors/meta-predictors, their symbols and their number of hyperparameters.}
\label{tab:list_predictor}
\fontsize{7.2}{8.5}\selectfont

\begin{tabular}{|l|l|l|}
\hline
\textbf{Predictor}                & \textbf{Symbol} & \textbf{Number of hyperparameters} \\ \hline
bayes.NaiveBayes                  & P0              & 0                                  \\ \hline
bayes.NaiveBayesMultinomial       & P1              & 0                                  \\ \hline
functions.Logistic                & P2              & 2                                  \\ \hline
functions.MultilayerPerceptron    & P3              & 7                                  \\ \hline
functions.SGD                     & P4              & 4                                  \\ \hline
functions.SimpleLogistic          & P5              & 4                                  \\ \hline
functions.SMO                     & P6              & 9                                  \\ \hline
functions.supportVector.RBFKernel & P7              & 9                                  \\ \hline
functions.VotedPerceptron         & P8              & 4                                  \\ \hline
lazy.IBk                          & P9              & 5                                  \\ \hline
lazy.KStar                        & P10             & 2                                  \\ \hline
rules.DecisionTable               & P11             & 4                                  \\ \hline
rules.JRip                        & P12             & 4                                  \\ \hline
rules.OneR                        & P13             & 1                                  \\ \hline
rules.PART                        & P14             & 3                                  \\ \hline
rules.ZeroR                       & P15             & 0                                  \\ \hline
trees.DecisionStump               & P16             & 0                                  \\ \hline
trees.J48                         & P17             & 2                                  \\ \hline
trees.LMT                         & P18             & 3                                  \\ \hline
trees.RandomForest                & P19             & 4                                  \\ \hline
trees.RandomTree                  & P20             & 3                                  \\ \hline
trees.REPTree                     & P21             & 6                                  \\ \hline
meta.AdaBoostM1                   & P22             & 4                                  \\ \hline
meta.AttributeSelectedClassifier  & P23             & 7                                  \\ \hline
meta.Bagging                      & P24             & 11                                 \\ \hline
meta.ClassificationViaRegression  & P25             & 2                                  \\ \hline
meta.LogitBoost                   & P26             & 8                                  \\ \hline
meta.MultiClassClassifier         & P27             & 6                                  \\ \hline
meta.RandomSubSpace               & P28             & 11                                 \\ \hline
meta.Vote                         & P29             & 3                                  \\ \hline
\end{tabular}

\end{table}

The configuration spaces that we manipulate consist of 30 predictors and all the preprocessing components available in Weka implemented in AutoWeka4MCPS.
The predictors are listed in Table~\ref{tab:list_predictor}, while all preprocessors are detailed in~\citep{sabu18}.
Among the listed predictors, eight meta-predictors, e.g.~ensemble methods, apply to at least one base learner. 
Each base learner is itself drawn from the 30 predictors, meta-predictors included.
We emphasise once more that, while it is possible to grow pipelines of infinite length recursively, the package applies a default limit of length seven.
Additionally, we note that components can vary significantly in the number of hyperparameters they involve, with \textit{RandomSubSpace} and \textit{Bagging} having the largest number at 11.

As a practical note, the implementation of AutoWeka4MCPS does not allow meta-predictors (P22--P29) or an SVM kernel (P7) to be run on their own, i.e. without a base learner or the Sequential Minimal Optimisation (SMO) function (P6), respectively.
Thus, should any of the reduction strategies in Section~\ref{Sec:Strategies} solely recommend meta-predictors and/or the SVM kernel, appropriate dependencies with the best rankings outside of the recommendation pool are also pulled in.
This can occasionally result in an AutoML process in Section~\ref{sec:results} presenting a base learner or SMO with alternate kernel as an optimal ML solution.
This is a non-ideal event, but the unavoidable quirk of implementation remains relatively rare.

\subsection{The Nature of the Meta-knowledge}
\label{sec:chap4_exp_metaknowledge}

In Section~\ref{sec:chap4_meta_construction}, we discussed the methodology underlying the creation of two meta-knowledge bases in opportunistic and systematic fashion, i.e.~\textit{automl-meta} and \textit{default-meta}, respectively.
Here, we examine how these meta-knowledge bases appear in practice, aiming to draw insight for later discussions around search-space reduction strategies.

\afterpage{
\begin{landscape}

\begin{figure*}[ht!] 
\centering 

\includegraphics[width=0.8\linewidth]{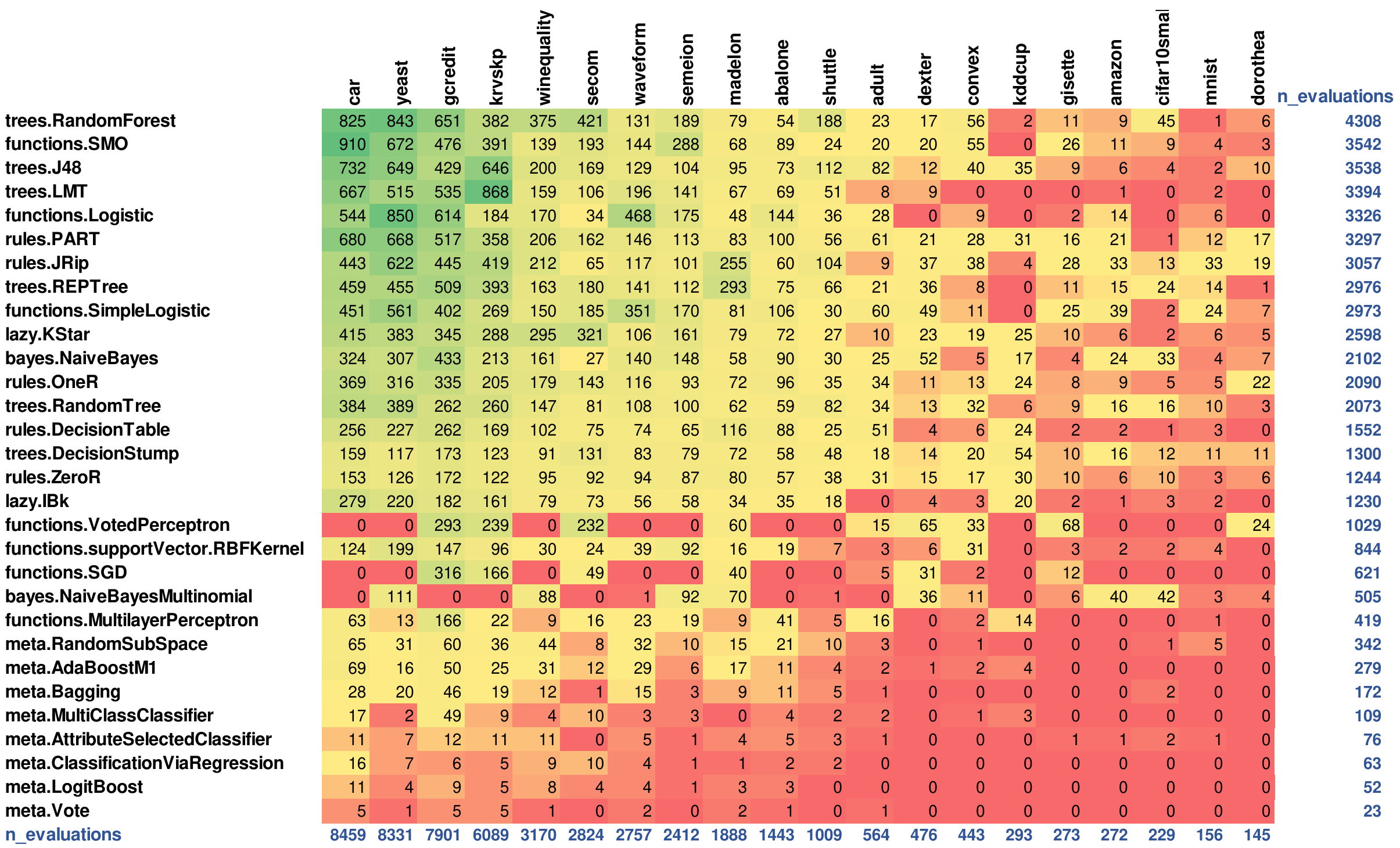}
\caption{The number of single-fold evaluations for each predictor on each dataset within the \textit{automl-meta} knowledge base. Predictors and datasets are both sorted from most to least evaluated.}

\label{fig:f1_automl_number_of_evaluations_heatmap}
\end{figure*}

\end{landscape}
}

\afterpage{
\begin{landscape}

\begin{figure*}[ht!] 
\centering 

\includegraphics[width=0.99\linewidth]{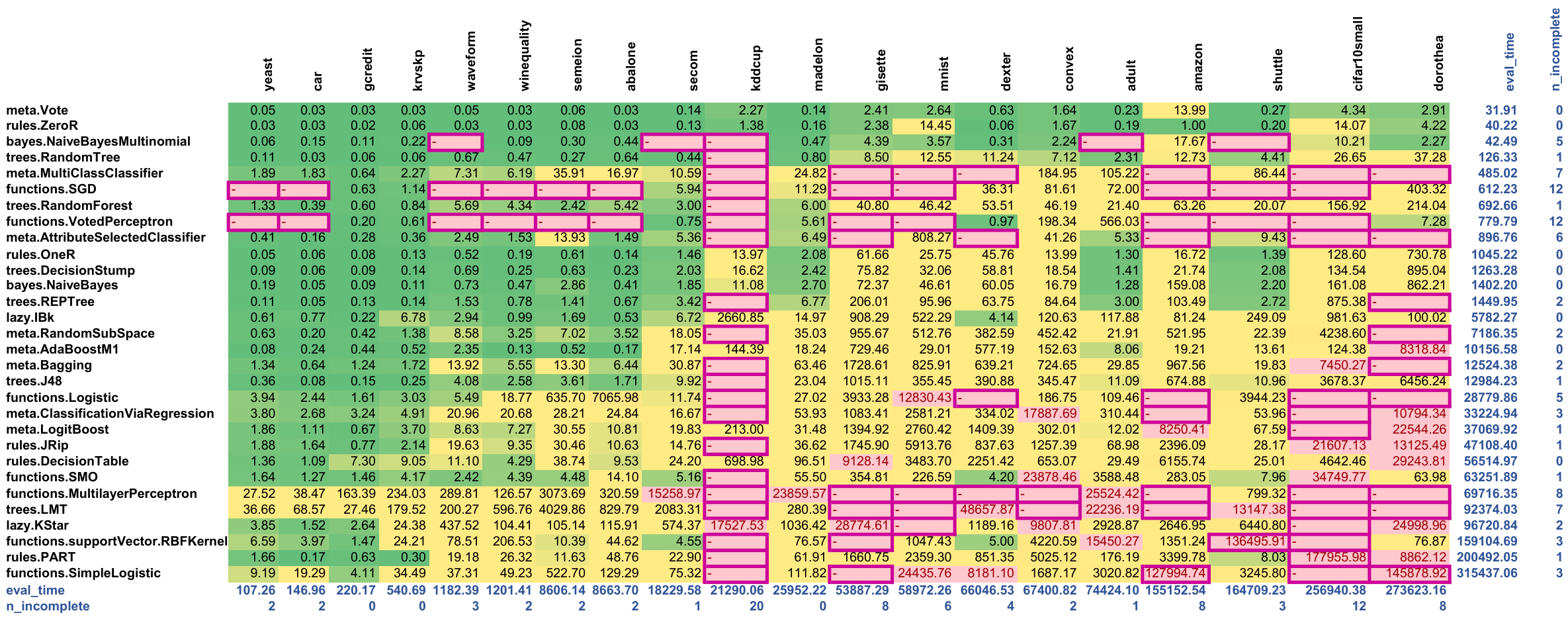}

\caption{The time in seconds to evaluate each predictor on each dataset, given default hyperparameters, via the 10-fold CV scheme. Describes time to generate the \textit{default-meta} knowledge base. Green cells denote low evaluation time ($\lesssim$ 10 s). Yellow cells denote moderate evaluation time ($\gtrsim$ 10 s and $\leq$ 7200 s). Pink cells denote high evaluation time ($>$ 7200 s), where 2 hours would be insufficient to evaluate all 10 CV folds. Pink borders denote extreme evaluation time ($>$ 72000 s) and incomplete runs, where 2 hours would be insufficient to evaluate a single CV fold. Predictors and datasets are both sorted from low to high total evaluation time. Incomplete runs have no associated time (0 s), so the total number per predictor/dataset is also noted.}

\label{fig:f2_default_time_of_evaluation_heatmap}
\end{figure*}

\end{landscape}
}

To begin with, Fig.~\ref{fig:f1_automl_number_of_evaluations_heatmap} depicts which predictor/dataset evaluations constitute the opportunistic \textit{automl-meta} knowledge base, where a single evaluation corresponds to one fold of a 10-fold CV calculation.
Each dataset receives five two-hour AutoML-based samplings, so ordering the datasets from most to least evaluated gives a fair indication of which ones require more computational resources for ML.
Although nonlinear ground-truth functions and other complexities hidden in the data are likely to contribute to lengthy training/validation times, processing speed is largely correlated with the characteristics in Table~\ref{tab:datasets}.
Predictors generally fit data quicker when there are fewer features, classes, and instances.
So, \textit{car}, \textit{yeast}, \textit{gcredit} and \textit{krvskp} are confirmed as lightweight datasets, as expected, with over 6000 hyperparameter configuration evaluations to leverage.
In contrast, \textit{dorothea}, \textit{mnist}, \textit{cifar10small}, \textit{amazon}, \textit{gisette} and \textit{kddcup} are examples of heavyweight datasets, each with under 300 evaluations for relevant reduction strategies to exploit.

This divide is reaffirmed within Fig.~\ref{fig:f2_default_time_of_evaluation_heatmap}, which shows how long it takes to run a 10-fold CV evaluation for each predictor on each dataset for default hyperparameters.
Specifically, for each of the four lightweight examples above, i.e.~\textit{car} et al., this form of systematic meta-knowledge, i.e.~\textit{default-meta}, was curated in under 10 minutes (600 s).
In contrast, for five of the six heavyweight examples above, i.e.~\textit{dorothea} et al., over 800 minutes (48000 s) -- sometimes well over -- were required per dataset.
The exception may seem to be \textit{kddcup}, but only because we cannot associate times with incomplete runs; these evaluations typically fail due to hardware limitations, e.g.~memory.
While the \textit{kddcup} dataset is particularly problematic, with 20 out of 30 predictors failing to run with default configurations, eight other datasets also posed problems for systematic curation.
Of course, the restrictions of computational resources manifest in similar ways between \textit{automl-meta} and \textit{default-meta}, so the impact of unreliable meta-knowledge due to evaluation sparsity is expected to be correlated.
Simply put, whether by insufficient AutoML-based sampling or incomplete default-hyperparameter evaluations, we are less confident in understanding the solution-performance space of \textit{dorothea} over \textit{car}.

Both Fig.~\ref{fig:f1_automl_number_of_evaluations_heatmap} and Fig.~\ref{fig:f2_default_time_of_evaluation_heatmap} give an indication of which ML algorithms are lightweight/heavyweight as well, but this is not as straightforward as for datasets.
The ordering of predictors in Fig.~\ref{fig:f2_default_time_of_evaluation_heatmap} is probably the most accurate estimation of computational complexity, given that there is no SMAC-based bias in evaluating any predictor.
However, incomplete runs distort the results somewhat.
Moreover, each predictor is evaluated for only one default configuration of hyperparameters.
Fortunately, there is still some degree of consistency.
In \textit{default-meta}, ZeroR, RandomTree, OneR, NaiveBayes and IBk are ranked 2, 4, 10, 12 and 14, respectively, for total benchmark evaluation time.
These five landmarkers, selected for having the fastest average evaluation times in \textit{automl-meta}, remain in the top half of predictors in \textit{default-meta} as assessed for efficiency.

Now, it is clear that forcing a poor choice of hyperparameters can hobble the efficiency of a predictor, so much so that a possibly decent accuracy may not be recorded in time.
After all, any missing values in \textit{automl-meta} and \textit{default-meta} earn predictors a loss of 100\%.
However, the systematic curation has a couple of seemingly positive consequences.
Firstly, with no restriction on process time, provided that predictor evaluation completes, \textit{default-meta} can potentially include good solutions that are inaccessible to \textit{automl-meta}.
This outcome is theoretically a strength of \textit{default-meta}, in that, despite its lack of solution-space coverage, it is not restricted from any part of that solution-space by computational resources.
Secondly, if the configuration space for a predictor is vast, as is the case for meta-predictors that wrap around base learners, expert knowledge can identify intelligent configurations to run.
Indeed, the two meta-predictors of Vote and MultiClassClassifier come first and fifth in terms of efficiency, respectively.
Furthermore, the average \textit{default-meta} efficiency rank of the eight meta-predictors is 13, which beats the average of all 30 predictors, i.e.~15.5.
This contrasts with Fig.~\ref{fig:f1_automl_number_of_evaluations_heatmap}, where all eight meta-predictors bottom the number-of-evaluations rankings.
They are presumably too unwieldy under random exploration to hold any appeal for SMAC.
So, one conclusion may be that \textit{default-meta} is a better meta-knowledge base than \textit{automl-meta} due to its `broader' set of evaluations.
However, a rebuttal may be that the application of search-space reduction strategies in Section~\ref{sec:results} is subject to the same constraints by which \textit{automl-meta} was generated, e.g.~five two-hour runs and biasing by SMAC.
Accordingly, does it help SMAC to point out predictors that perform well for configurations that SMAC could not even initially sample?

For the time being, the takeaway from Fig.~\ref{fig:f1_automl_number_of_evaluations_heatmap} and Fig.~\ref{fig:f2_default_time_of_evaluation_heatmap} is that there is a relatively consistent gradient of computational difficulty for datasets used in this work, which has substantial impact on evaluating predictor performance.
From an efficiency perspective, the appeal of predictors is harder to rank, being very sensitive to hyperparameter choices.
The number-of-evaluations rankings in Fig.~\ref{fig:f1_automl_number_of_evaluations_heatmap} lean towards quick predictors, but they are also biased towards what SMAC considers optimally accurate.
Sure enough, Section~\ref{sec:results} affirms that SMO, the second-most evaluated ML algorithm, is generally a robust predictor for this selection of datasets, even though Fig.~\ref{fig:f2_default_time_of_evaluation_heatmap} does not score SMO as particularly efficient.

\afterpage{
\begin{landscape}

\begin{figure*}[ht!] 
\centering 

\includegraphics[width=0.99\linewidth]{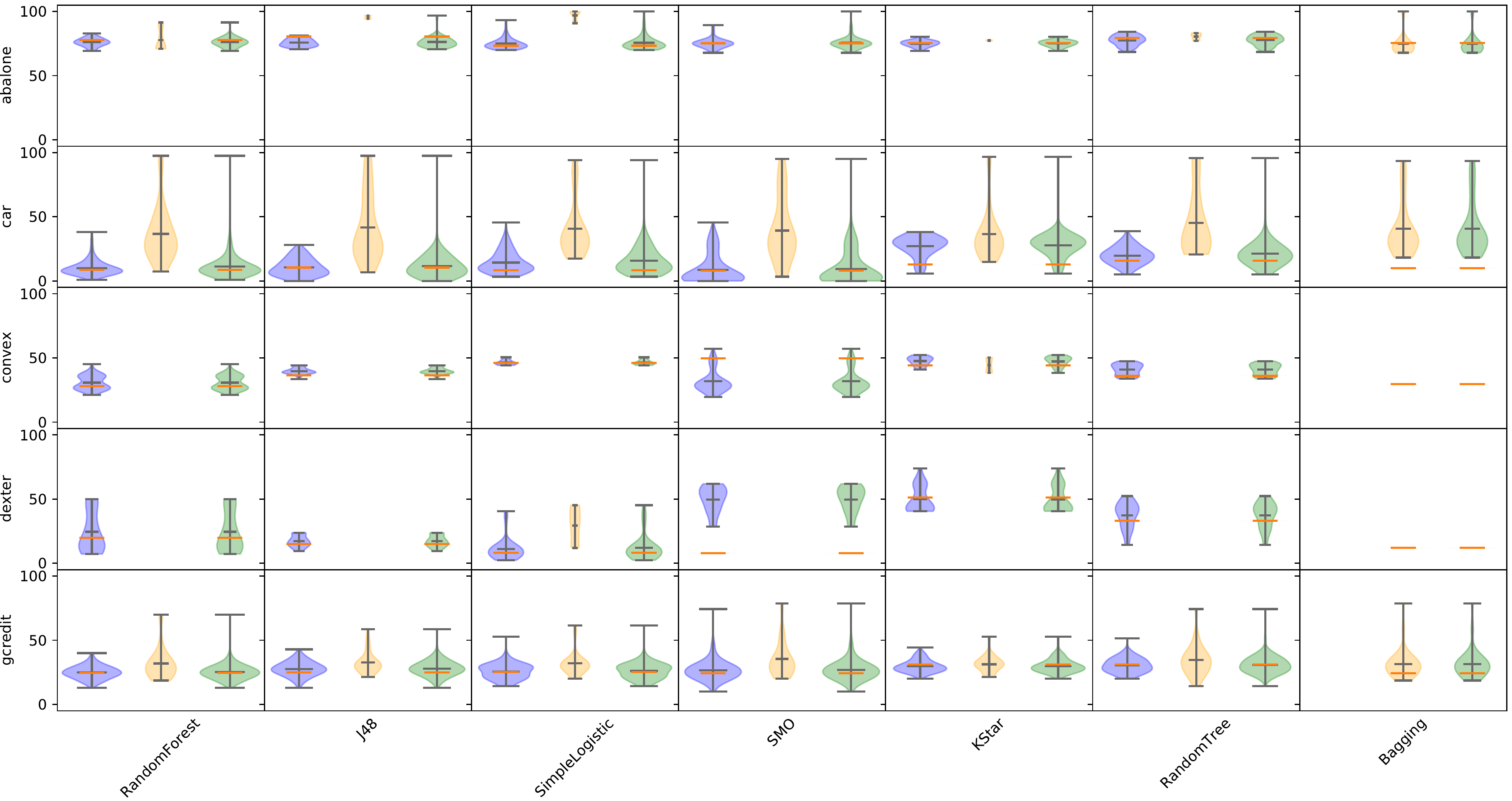}

\caption{Violin plots exemplifying the contents of the opportunistic and systematic meta-knowledge bases. Displayed data is partitioned by dataset and predictor. Violin shapes represent the distribution of single-fold predictor evaluations for \textit{automl-meta}, with their width denoting the density of evaluations around a particular score. Orange bars mark the 10-fold CV mean of a predictor evaluation, with default hyperparameters, for \textit{default-meta}. All evaluations are in terms of error rates. For each pair of dataset and predictor, the blue, yellow and green violin shapes represent the distribution of \textit{automl-meta} evaluations for single-component pipelines, multi-component pipelines and their combination, respectively.}

\label{fig:violin_chart_3types_apart}
\end{figure*}

\end{landscape}
}

Turning now from computational resource expenditures to evaluated accuracies, the violin plots of Fig.~\ref{fig:violin_chart_3types_apart} provide a representative snapshot of the meta-knowledge bases.
There are 600 combinations of datasets and predictors, i.e.~$20\times 30$, and the chart depicts 35 of them, i.e.~$5\times 7$.
The green distributions -- positioned third per table cell -- are key, depicting the performance spread of all evaluated ML pipelines in \textit{automl-meta} for each dataset-predictor pair.
Again, each evaluation within a distribution is for a single fold within a 10-fold CV scheme.
With appropriate scaling, each of these distributions decomposes into a blue distribution -- positioned first per table cell -- of single-component evaluations, plus a yellow distribution -- positioned second per table cell -- of multi-component evaluations.
Each displayed orange line additionally marks the mean of a 10-fold CV evaluation for a predictor with default hyperparameters, i.e.~\textit{default-meta}.
In the case of meta-predictors, e.g.~Bagging, every ML pipeline consists of two or more components due to the presence of a base learner, so there are no blue distributions.
The default evaluation is likewise overlaid over the yellow distribution for these cases.

Immediately, a few characteristics are noticeable from Fig.~\ref{fig:violin_chart_3types_apart}.
Each dataset has a unique `intrinsic' difficulty, discernible in that predictor performances tend to cluster around specific values.
Indeed, theoretically, each problem will be associated with its own maximal accuracy, beyond which classes cannot be separated further without overfitting.
The size of a dataset does not necessarily determine this.
So, while \textit{car} and \textit{gcredit} are both lightweight, near-perfect classification is possible for the former, while predictors struggle to lower loss beyond $10\%$ for the latter.
Similarly, despite all being considered more heavyweight, \textit{abalone}, \textit{convex} and \textit{dexter} have distributions clustered in differing regions along the error-rate axis.
The computational resources demanded by \textit{dexter} do not prevent predictors reaching near-perfect levels of accuracy, while, in contrast, \textit{abalone} classifiers tend to return ${\sim}75\%$ error rates.
These outcomes also indicate why relative rankings for assessing predictors have an appeal; there is no sensible way to translate absolute error rates between datasets.

One differentiation that the lightweight/heavyweight divide does seem to result in, as expected, is in the number of evaluations and associated coverage.
For \textit{car} and \textit{gcredit}, every predictor displayed in Fig.~\ref{fig:violin_chart_3types_apart} has had SMAC sample numerous multi-component pipelines.
In contrast, the other three datasets have smaller distributions of opportunistic evaluations to exploit, often failing to assess preprocessors and even meta-predictors.
Additionally, while SMAC has consistently found hyperparameter configurations that perform better than the defaults for all displayed predictors applied to \textit{car} and \textit{gcredit}, the defaults are still superior for several predictors applied to the other three datasets, e.g.~J48 and SMO.
Whether by luck or robust expert knowledge, the accuracy boosts of default hyperparameters are particularly pronounced for \textit{abalone}, with reductions in loss of up to ${\sim}40\%$.

In general, the expert tuning of default hyperparameters holds up reasonably well, supporting remarks in the literature~\citep{wemu20}.
Of 600 dataset-predictor pairs, 464 have both a default configuration and a SMAC-based distribution of evaluations to compare.
Given these comparison-ready pairs, default hyperparameters do better than the mean performance of ML pipelines evaluated by SMAC in 406/464 cases.
Granted, SMAC is expected to explore poorly performing ML pipelines as part of the AutoML process, but the exploitation of the strongest performers should counterbalance this effect.
Additionally, in 94/464 cases, default configurations do better than any SMAC-based exploratory evaluation, at least within five runs of two hours.
However, for completeness, it is worth noting that default hyperparameters do worse than the SMAC-based distribution mean in 58/464 cases, and 41 of these involve the default configuration being worse than anything SMAC evaluates.
In the off-chance that a default choice is bad for an ML problem, it tends to be really bad.

Returning to a vital facet of this research, we consider the opportunistic incorporation of multi-component pipelines in \textit{automl-meta}.
Evidently, as Fig.~\ref{fig:violin_chart_3types_apart} shows, five two-hour runs of SMAC per dataset are insufficient to generate exhaustive coverage of preprocessors, which is why this work focusses only on predictors when reducing configuration spaces.
The question then is whether leveraging multi-component pipelines provides any utility for reduction strategies.
Hypothetically, bumping a predictor up in the rankings due to strong synergy with a preprocessor may give SMAC a better chance to explore the pipeline within a reduced search space.
On the other hand, if SMAC is unlikely to explore that ML pipeline again, given that multi-component subspace is larger than single-component subspace, then perhaps exploiting the information is counterproductive.
The first step in analysing whether this is true is simply observing how the distributions of evaluations compare.

Of 600 dataset-predictor pairs, 383 combinations have single-component evaluations, which is not particularly low when considering that 160 pairings involving meta-predictors, i.e.~$8\times 20$, do not have single-component forms.
For multi-component evaluations, 337/600 pairs have them.
In the overlap, there are 231/600 pairs possessing both single-component and multi-component ML pipeline evaluations, e.g.~table cells in Fig.~\ref{fig:violin_chart_3types_apart} that have both blue and yellow distributions.
Within \textit{automl-meta}, we find that the mean of multi-component pipeline evaluations beats the mean of single-component evaluations only 31/231 times.
A comparison of minima provides a similar result of 28/231.
Thus, while there are clear cases where preprocessors improve the performance of ML solutions, the inclusion of multi-component pipelines may arguably, in general, distort performance assessments of predictors for little gain.
Admittedly, this distortion may be minor for the exact same reason that excellent multi-component evaluations are rare: five two-hour runs of AutoML are just too short for SMAC to substantially explore/exploit extended pipelines.

\afterpage{
\begin{landscape}
\begin{figure*}[ht!] 
\centering 
\includegraphics[width=0.8\linewidth]{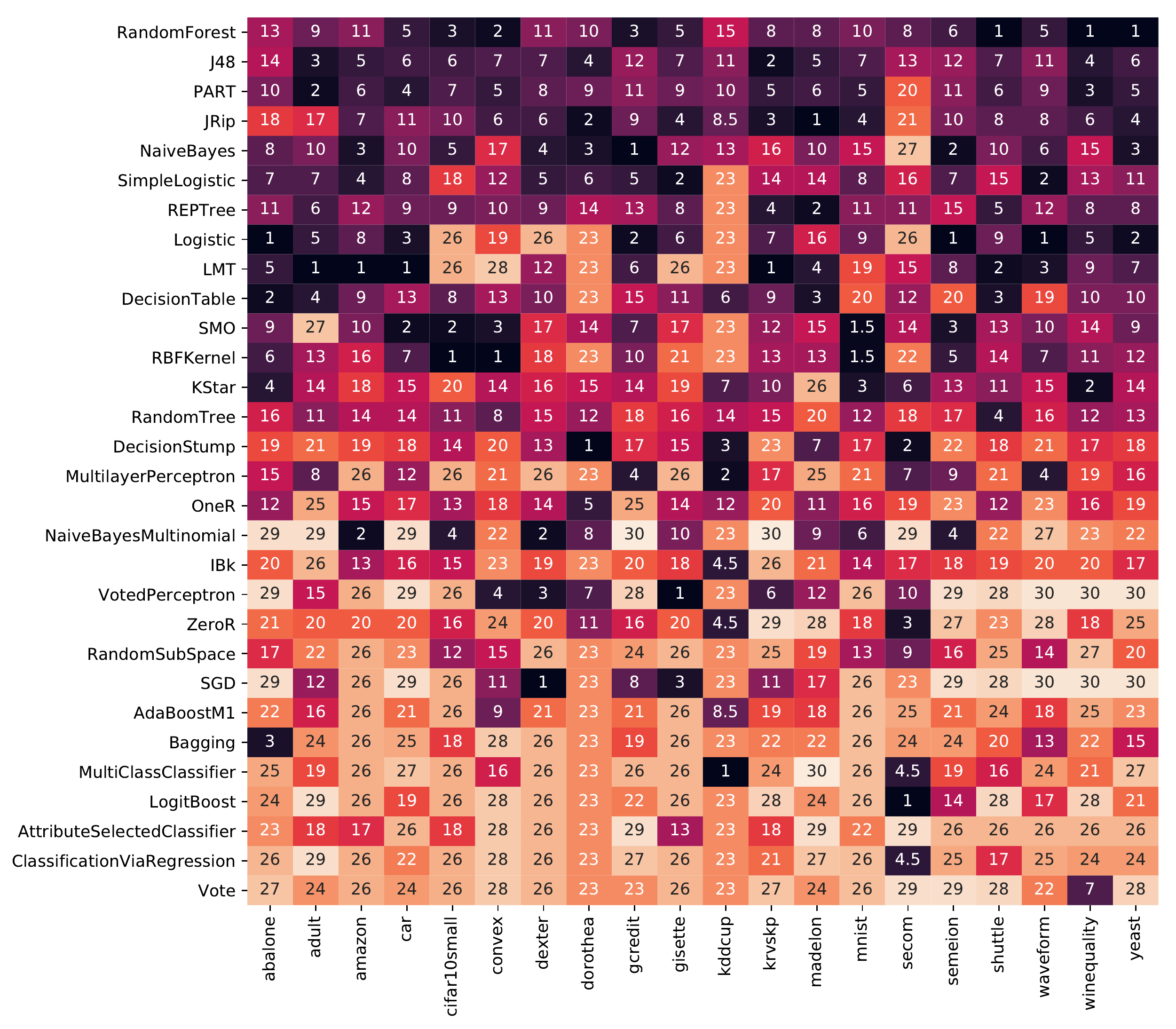}
\caption{Predictor rankings within \textit{automl-meta}. They are ranked per dataset according to the mean accuracy of evaluated ML pipelines that contain them. Equal accuracies for a dataset earn predictors an averaged rank, e.g.~4.5 and 4.5 instead of 4 and 5. Predictor labels are sorted by their average rank across all datasets, providing a `global leaderboard' for all \textit{M1-kn} strategies.}
\label{fig:f4_automl_predictor_rankings_per_dataset_heatmap}
\end{figure*}
\end{landscape}
}

\afterpage{
\begin{landscape}
\begin{figure*}[ht!] 
\centering 
\includegraphics[width=0.8\linewidth]{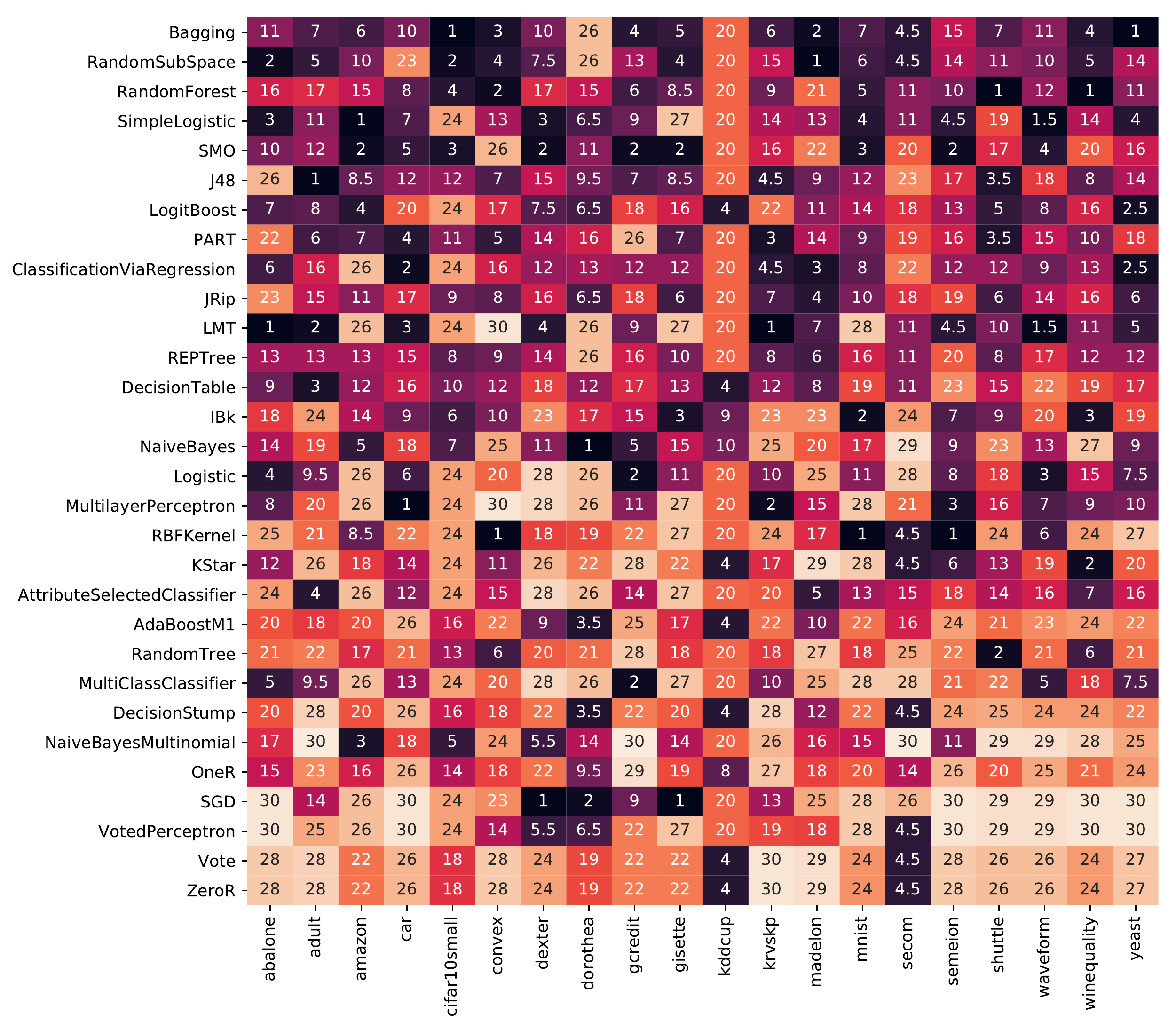}
\caption{Predictor rankings within \textit{default-meta}. They are ranked per dataset according to their 10-fold CV accuracy, given default hyperparameters. Equal accuracies for a dataset earn predictors an averaged rank, e.g.~4.5 and 4.5 instead of 4 and 5. Predictor labels are sorted by their average rank across all datasets, providing a `global leaderboard' for all \textit{M2-kn} strategies.}
\label{fig:f5_default_predictor_rankings_per_dataset_heatmap}
\end{figure*}
\end{landscape}
}

\begin{table}[htb!]

\vspace{-0.35cm}

\caption {Comparison between SMAC-based predictor rankings generated when multi-component evaluations are included, i.e.~\textit{automl-meta} in Fig.~\ref{fig:f4_automl_predictor_rankings_per_dataset_heatmap}, and excluded. Comparison metric is the Pearson correlation coefficient.}
\label{tab:correlation_single_vs_total}

\centering
\fontsize{6.5}{6.5}\selectfont
\setlength\tabcolsep{1.5pt}

\begin{tabular}{|c|c|}
\hline
\textbf{dataset} & \textbf{\begin{tabular}{@{}c@{}}correlation \\ (with vs.~without multi-component)\end{tabular}} \\ \hline
abalone          & 0.845                                        \\ \hline
adult            & 0.902                                        \\ \hline
amazon           & 0.916                                        \\ \hline
car              & 0.946                                        \\ \hline
cifar10small     & 0.932                                        \\ \hline
convex           & 0.863                                        \\ \hline
dexter           & 0.994                                        \\ \hline
dorothea         & 0.997                                        \\ \hline
gcredit          & 0.945                                        \\ \hline
gisette          & 0.96                                         \\ \hline
kddcup           & 0.765                                        \\ \hline
krvskp           & 0.898                                        \\ \hline
madelon          & 0.936                                        \\ \hline
mnist            & 0.953                                        \\ \hline
secom            & 0.49                                         \\ \hline
semeion          & 0.896                                        \\ \hline
shuttle          & 0.883                                        \\ \hline
waveform         & 0.864                                        \\ \hline
winequality      & 0.874                                        \\ \hline
yeast            & 0.941                                        \\ \hline
\end{tabular}

\end{table}

Ultimately, all configuration-space reduction strategies based on meta-learning, as listed in Section~\ref{Sec:Strategies}, care only about how predictors are ranked.
In the case of the single-fold evaluation distributions of \textit{automl-meta}, there are several options for how to construct these rankings, each uniquely justifiable.
We settle for comparing mean performances in this work to better reflect that optimisation is a process; there is little point highlighting a predictor with the possibility of exceptional performance if subsequent applications of SMAC are unable/unwilling to journey there from a poor mean.
Thus, given this choice, predictor rankings for \textit{automl-meta} and \textit{default-meta} are presented in Fig.~\ref{fig:f4_automl_predictor_rankings_per_dataset_heatmap} and Fig.~\ref{fig:f5_default_predictor_rankings_per_dataset_heatmap}, respectively.

An immediate question in the case of \textit{automl-meta}, given the preceding discussion on ML pipelines, is whether evaluating multi-component solutions impacted these rankings.
To examine this, Table~\ref{tab:correlation_single_vs_total} compares the rankings in Fig.~\ref{fig:f4_automl_predictor_rankings_per_dataset_heatmap} with what they would be if multi-component evaluations were not included.
In general, with 11 datasets showing a correlation of above 0.9 and another seven datasets also above 0.84, evaluating extended pipelines does not severely upend rankings.
It is important to note that, for this correlation analysis, excluding multi-component pipeline evaluations does place all meta-predictors at the back of the rankings, as unevaluated predictors score a loss of $100\%$.
This manner of handling missing values is rarely a major issue, as SMAC does not find meta-predictors generally appealing in two-hour runs anyway.
Nonetheless, major deviation from strong correlation does occasionally occur for datasets that, in the presence of multi-component evaluations, rank meta-predictors well, e.g.~\textit{kddcup} and \textit{secom}.
Consequently, the impact of multi-component evaluations related to preprocessors is even more subtle once the influence of meta-predictors is ablated; their advantage/disadvantage remains somewhat inconclusive for the reduction strategies employed in this work.

\begin{figure}[h] 
    \centering
    \includegraphics[width=0.99\linewidth]{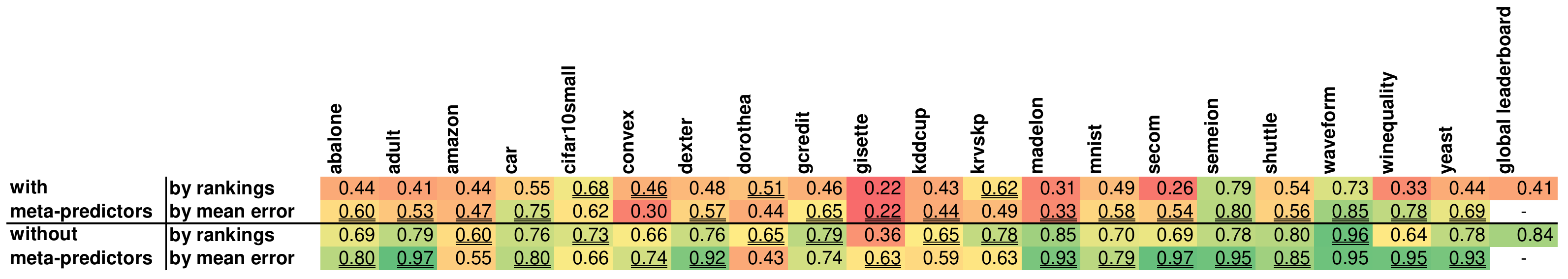}
    \vspace*{-1mm}
     \caption{Correlation coefficients measuring the similarity of predictor performance between \textit{automl-meta} and \textit{default-meta}; green to red denotes high to low similarity. Correlations are calculated per dataset in terms of both rankings and the fold-averaged error rates that inform those rankings. The correlation between global leaderboards is solely in terms of rankings. Correlations are additionally presented for both the 30 and 22 predictors that include and exclude meta-predictors, respectively. Underlined values indicate whether the rankings or error rates are most similar.}
    \label{fig:default_and_automl_correlations_with_without_meta_predictors}
\end{figure}

A core research question for this monograph is whether the choice of opportunistic or systematic meta-knowledge matters when constraining AutoML search space.
At a first glance, the rankings in Fig.~\ref{fig:f4_automl_predictor_rankings_per_dataset_heatmap} and Fig.~\ref{fig:f5_default_predictor_rankings_per_dataset_heatmap} seem to differ significantly, suggesting that the spaces recommended by \textit{O1-kn} and \textit{O2-kn}, or \textit{M1-kn} and \textit{M2-kn}, may be substantially inconsistent.
Indeed, ranking-specific correlations in the first row of Fig.~\ref{fig:default_and_automl_correlations_with_without_meta_predictors} show that only four datasets have a coefficient of above 0.6, with none over 0.8.
The global leaderboards, i.e.~rankings averaged over all 20 datasets, are likewise only weakly similar with a value of 0.41.
One plausible factor that could cause this discrepancy is the ambiguity of rankings; it is not usually apparent if the difference between a best and second-best predictor is large or small.
Thus, even if the mean performance of two predictors differs only slightly between \textit{automl-meta} and \textit{default-meta}, their rankings could be flipped.
Sure enough, the second row in Fig.~\ref{fig:default_and_automl_correlations_with_without_meta_predictors} supports this hypothesis, with predictor performances being more similar in terms of error rates than rankings for 15 datasets.
This result underscores the danger of reading too much into relative performance, especially for datasets where the accuracies of ML algorithms are tightly bunched up; see Section~\ref{sec:chap4_dataset_understand} for attempts to identify such cases.

Nonetheless, despite the seemingly vast discrepancies between \textit{automl-meta} and \textit{default-meta}, it is difficult not to catch certain trends.
For instance, the top four SMAC-identified global performers in Fig.~\ref{fig:f4_automl_predictor_rankings_per_dataset_heatmap} show up in the same order within the top ten default-hyperparameter spots of Fig.~\ref{fig:f5_default_predictor_rankings_per_dataset_heatmap}; they are the tree-based RandomForest and J48, as well as the rule-based PART and JRip.
At the same time, it is several meta-predictors that flesh out these top ten \textit{default-meta} rankings, i.e.~Bagging, RandomSubSpace, LogitBoost, and ClassificationViaRegression.
In truth, \textit{default-meta} may be a better reflection of conventional ML knowledge, as many meta-predictors represent powerful learning principles, e.g.~ensemble-based approaches, and these are given their due here.
It is also convincing that ZeroR -- the default configuration of meta-predictor Vote also collapses to ZeroR -- is at the end of the \textit{default-meta} rankings; the ML algorithm is effectively a simple-heuristic majority-class classifier and not expected to be particularly smart.
However, powerful approaches can be complex and slow, and \textit{automl-meta} demonstrates this by ranking every meta-predictor as globally less accurate than ZeroR.

Driven by this insight, Fig.~\ref{fig:default_and_automl_correlations_with_without_meta_predictors} repeats the correlation analysis for 22 ML algorithms, ablating away the impact of meta-predictors.
Sure enough, correlation strengthens for virtually every dataset in both rankings and mean error rates.
Where it does not, the decrease is no more than 0.01.
Moreover, barring \textit{gisette}, every ranking-specific correlation coefficient in the third row is now 0.6 or greater, with three datasets at 0.8 or above.
In fact, both \textit{automl-meta} and \textit{default-meta} provide a fairly consistent opinion on which ML algorithms are typically better than others, with a global-leaderboard correlation of 0.84.
Accordingly, if a reduced configuration space does not include meta-predictors or is otherwise large enough for SMAC to glide over these options, should they indeed be counterproductive recommendations, then perhaps the choice of meta-knowledge base is largely irrelevant.
We examine if this is the case in Section~\ref{sec:results}.
Nonetheless, even without the meta-predictor question, the fourth row of Fig.~\ref{fig:default_and_automl_correlations_with_without_meta_predictors} maintains that, for 13 datasets, predictor error rates are more consistent between meta-knowledge bases than their rankings.
This warrants closer investigation into how the performance of all predictors is distributed per dataset.

\subsection{The Challenge of the Datasets}
\label{sec:chap4_dataset_understand}

As the previous subsections show, the 20 datasets adopted for investigation in this work display various characteristics and ML outcomes.
Conventional meta-learning research would consider such a diversity of meta-knowledge to be a strength due to its scope of representation, even if it remains unclear how best to quantify and assess that diversity in general~\citep{muma18}.
However, what is sometimes neglected in the field is that diversity is only useful if it is exploited well.
So, when preliminary results for this investigation into configuration-space reduction~\citep{ngke21} appeared only weakly conclusive, we hypothesised that averaging out the nuances of 20 datasets was partially responsible.
Specifically, aggregation might prove limiting in a couple of ways: (1) the strength of recommendations based on meta-knowledge may be diluted by predictor rankings that are less robust for some datasets than others, and (2) conclusions drawn across an entire benchmark may be weakened if specificity in strategic recommendations only matters for certain datasets.
The question is then as follows: for which datasets is the choice of predictor or search-space reduction strategy likely to matter?

\begin{figure}[ht!] 
  
    \subfloat[abalone\label{fig:violin_plots_n_predictor_abalone}]{%
        \includegraphics[width=0.24\linewidth]{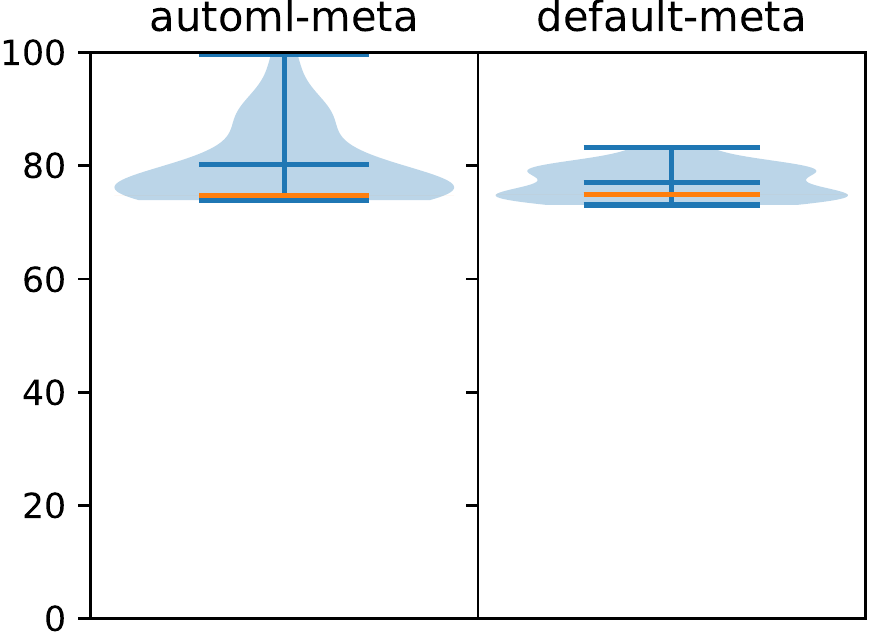}
        }%
    \hfill
    \subfloat[adult\label{fig:violin_plots_n_predictor_adult}]{%
        \includegraphics[width=0.24\linewidth]{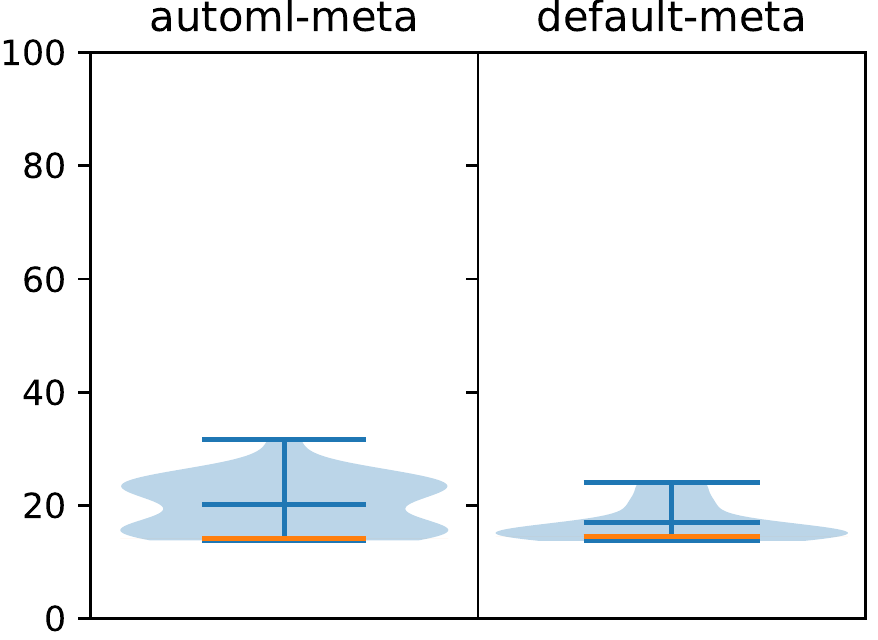}
        }%
    \hfill
    \subfloat[amazon\label{fig:violin_plots_n_predictor_amazon}]{%
        \includegraphics[width=0.24\linewidth]{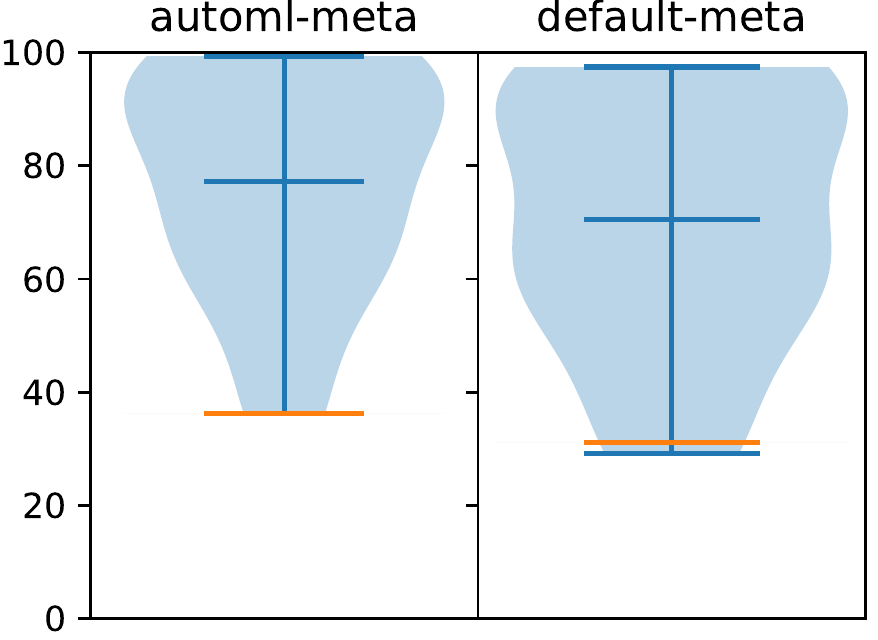}
        }%
    \hfill
    \subfloat[car\label{fig:violin_plots_n_predictor_car}]{%
        \includegraphics[width=0.24\linewidth]{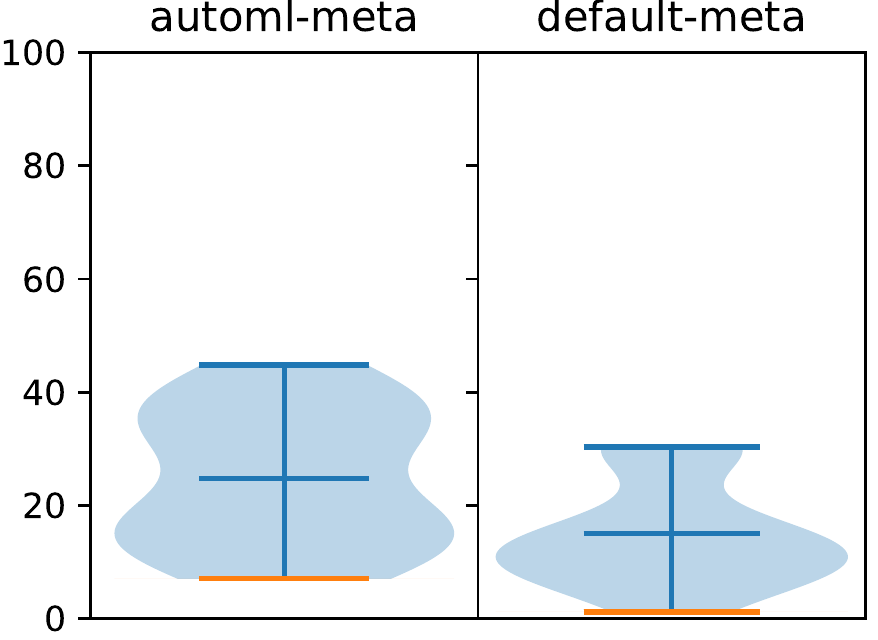}
        }%

    \subfloat[cifar10small\label{fig:violin_plots_n_predictor_cifar10small}]{%
        \includegraphics[width=0.24\linewidth]{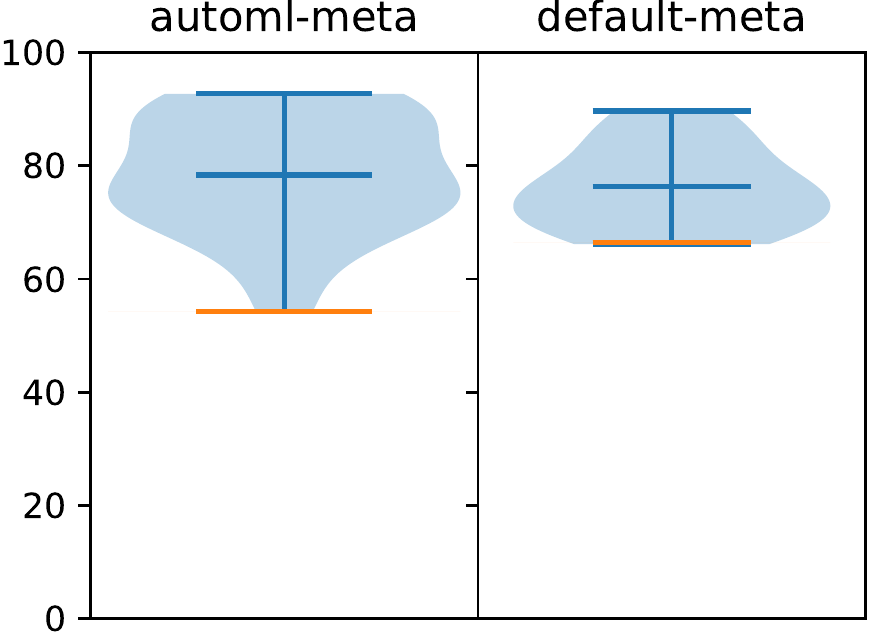}
        }%
    \hfill
    \subfloat[convex\label{fig:violin_plots_n_predictor_convex}]{%
        \includegraphics[width=0.24\linewidth]{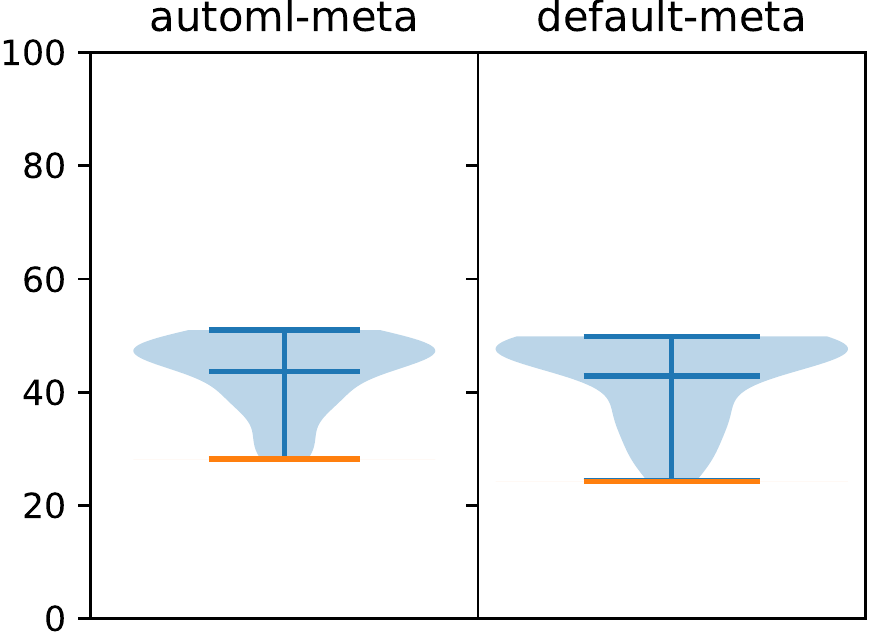}
        }%
    \hfill
    \subfloat[dexter\label{fig:violin_plots_n_predictor_dexter}]{%
        \includegraphics[width=0.24\linewidth]{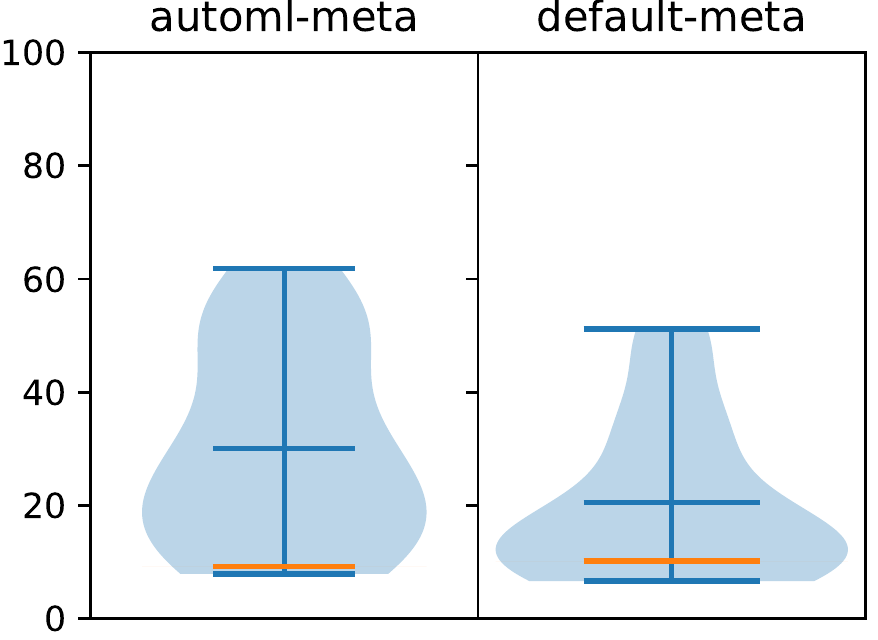}
        }%
    \hfill
    \subfloat[dorothea\label{fig:violin_plots_n_predictor_dorothea}]{%
        \includegraphics[width=0.24\linewidth]{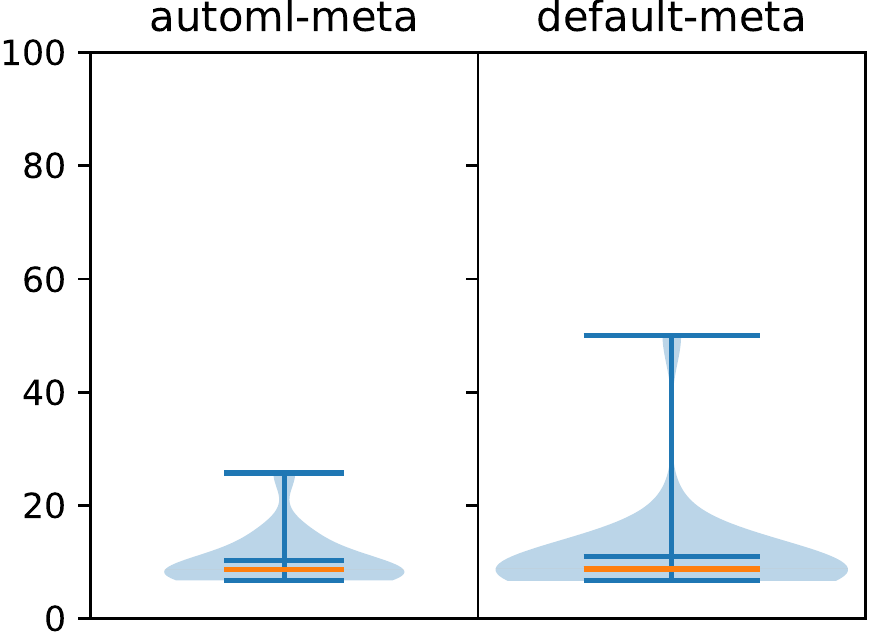}
        }%

    \subfloat[gcredit\label{fig:violin_plots_n_predictor_gcredit}]{%
        \includegraphics[width=0.24\linewidth]{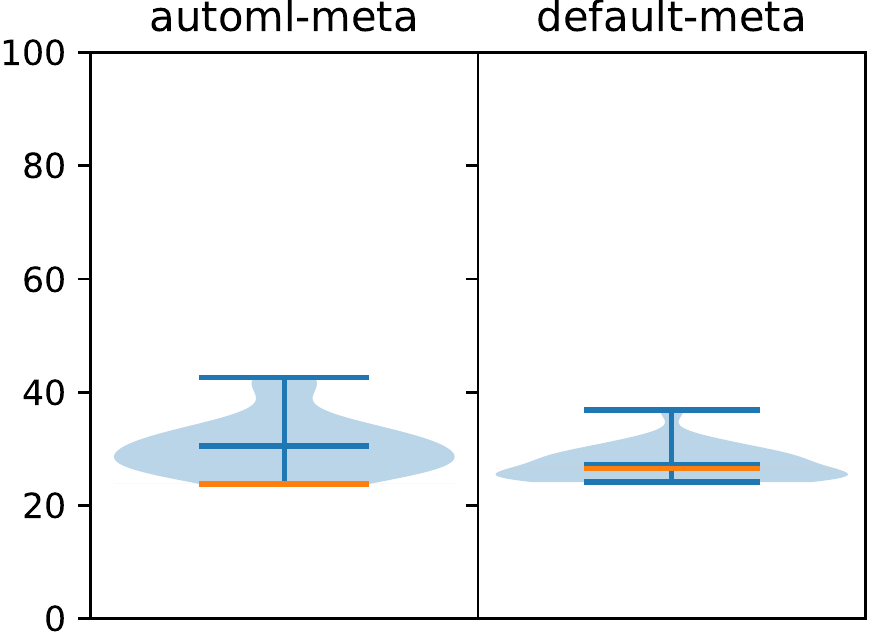}
        }%
    \hfill
    \subfloat[gisette\label{fig:violin_plots_n_predictor_gisette}]{%
        \includegraphics[width=0.24\linewidth]{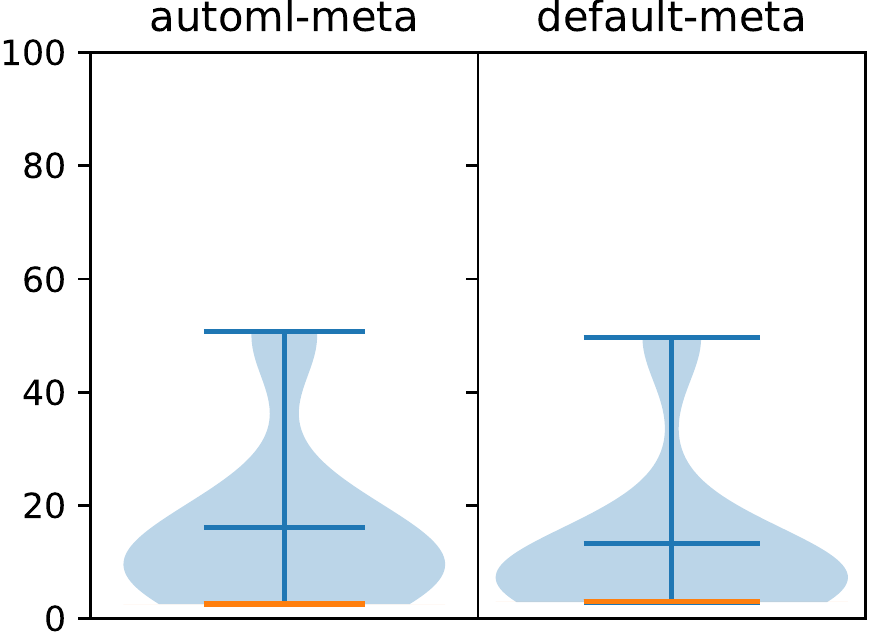}
        }%
    \hfill
    \subfloat[kddcup\label{fig:violin_plots_n_predictor_kddcup}]{%
        \includegraphics[width=0.24\linewidth]{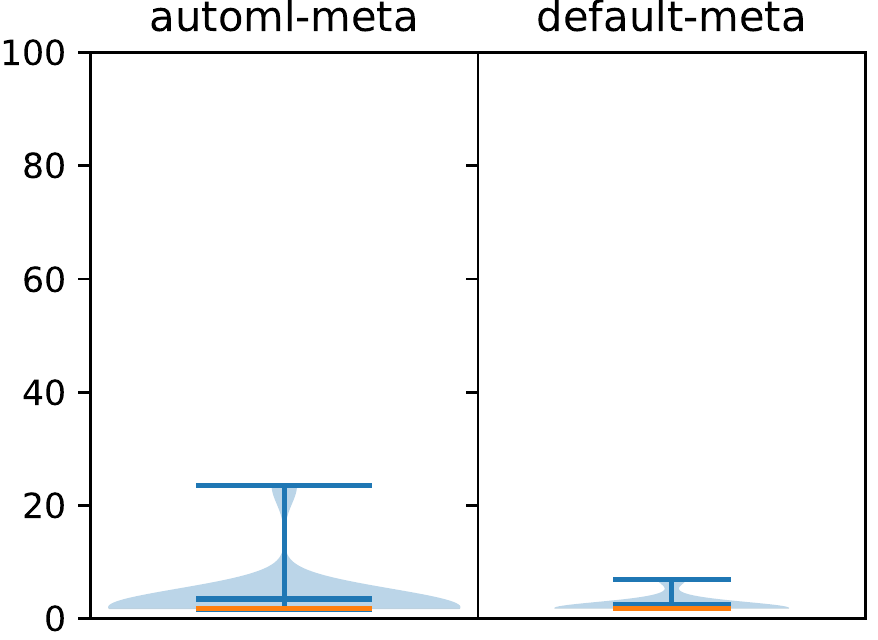}
        }%
    \hfill
    \subfloat[krvskp\label{fig:violin_plots_n_predictor_krvskp}]{%
        \includegraphics[width=0.24\linewidth]{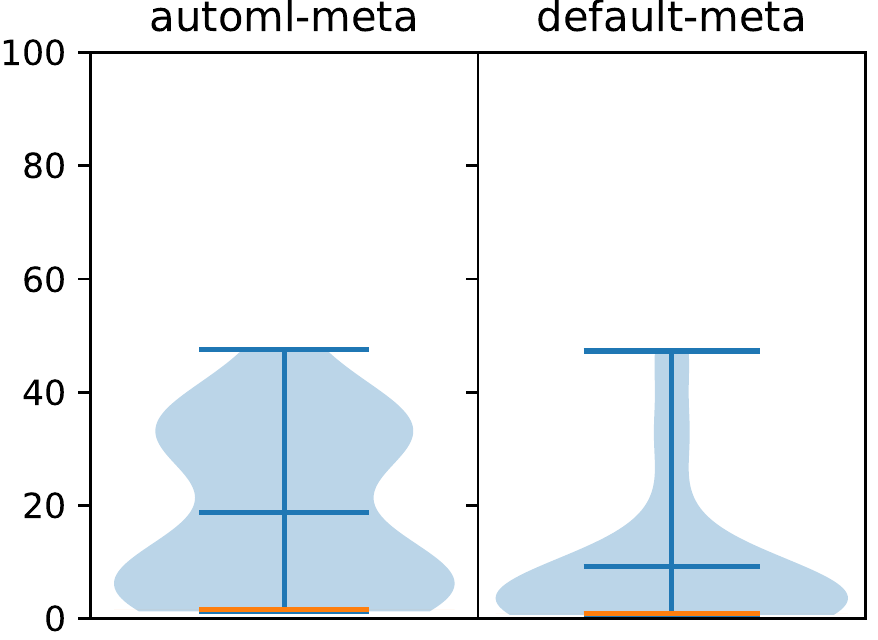}
        }%

    \subfloat[madelon\label{fig:violin_plots_n_predictor_madelon}]{%
        \includegraphics[width=0.24\linewidth]{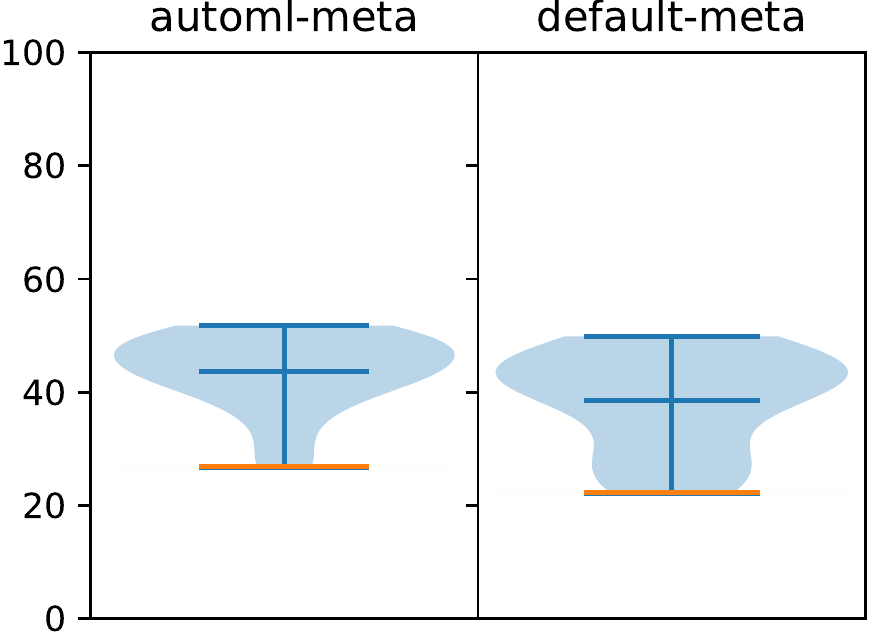}
        }%
    \hfill
    \subfloat[mnist\label{fig:violin_plots_n_predictor_mnist}]{%
        \includegraphics[width=0.24\linewidth]{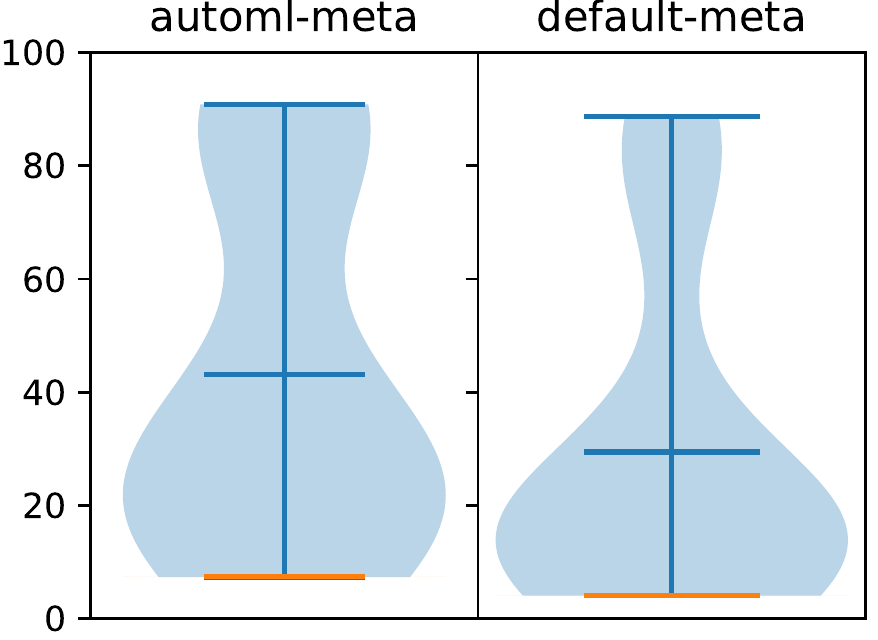}
        }%
    \hfill
    \subfloat[secom\label{fig:violin_plots_n_predictor_secom}]{%
        \includegraphics[width=0.24\linewidth]{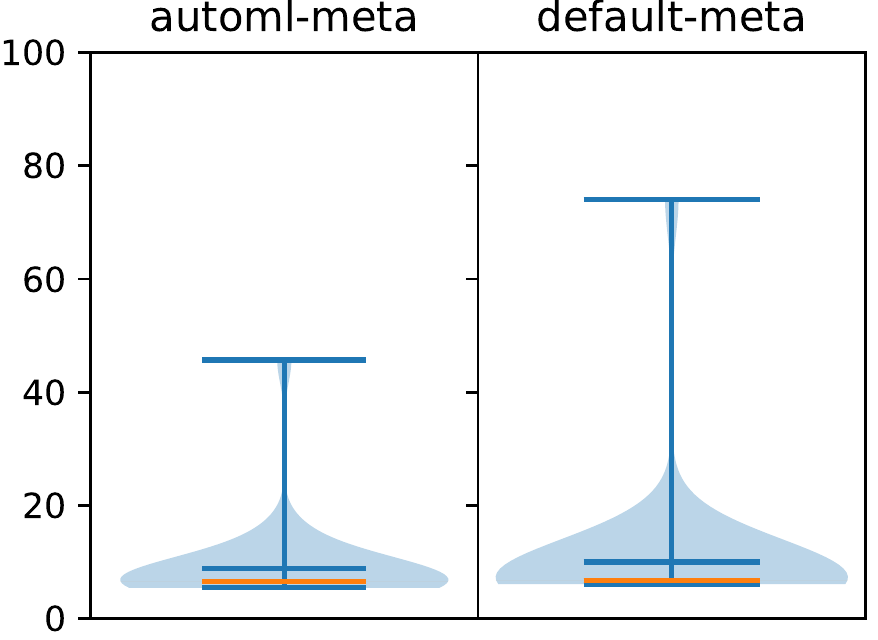}
        }%
    \hfill
    \subfloat[semeion\label{fig:violin_plots_n_predictor_semeion}]{%
        \includegraphics[width=0.24\linewidth]{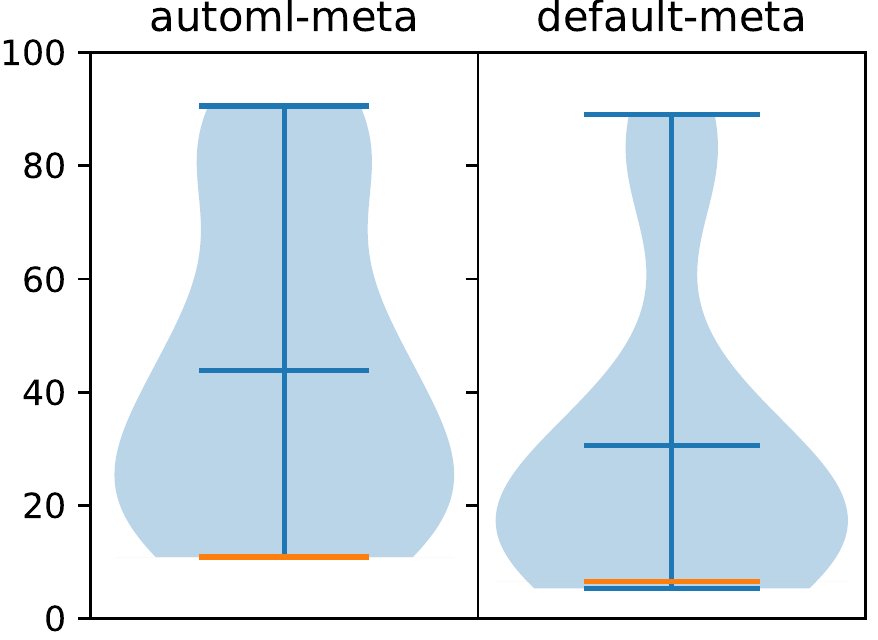}
        }%

    \subfloat[shuttle\label{fig:violin_plots_n_predictor_shuttle}]{%
        \includegraphics[width=0.24\linewidth]{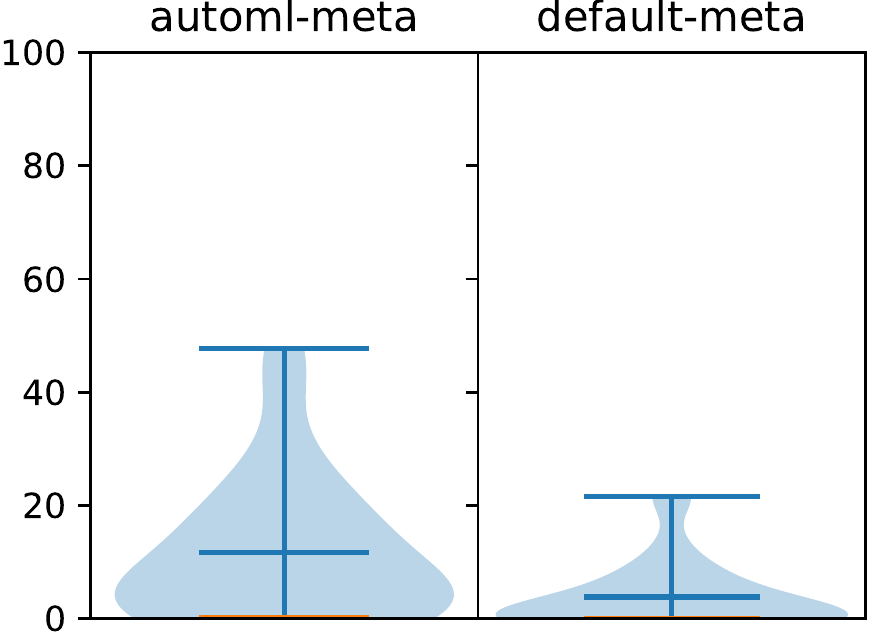}
        }%
    \hfill
    \subfloat[waveform\label{fig:violin_plots_n_predictor_waveform}]{%
        \includegraphics[width=0.24\linewidth]{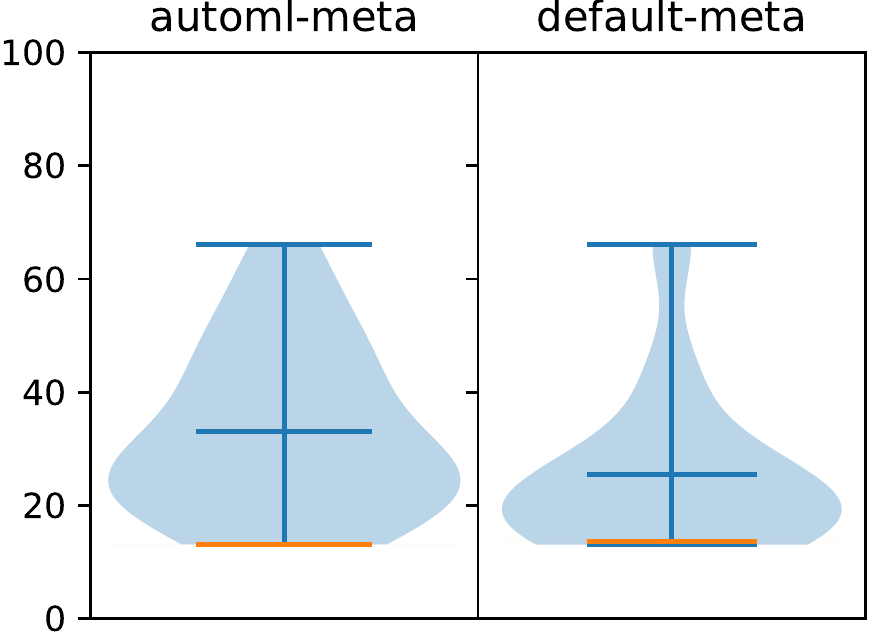}
        }%
    \hfill
    \subfloat[winequality\label{fig:violin_plots_n_predictor_winequality}]{%
        \includegraphics[width=0.24\linewidth]{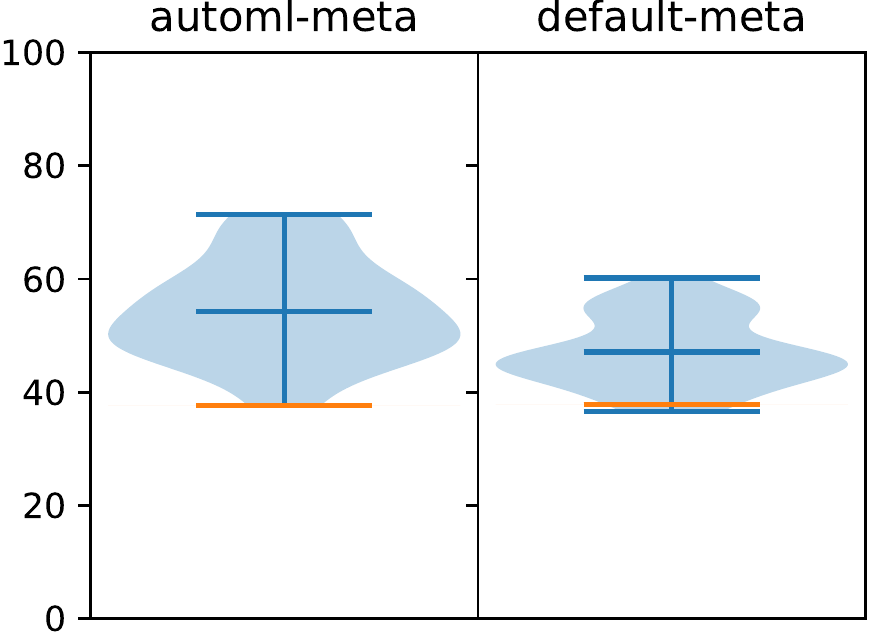}
        }%
    \hfill
    \subfloat[yeast\label{fig:violin_plots_n_predictor_yeast}]{%
        \includegraphics[width=0.24\linewidth]{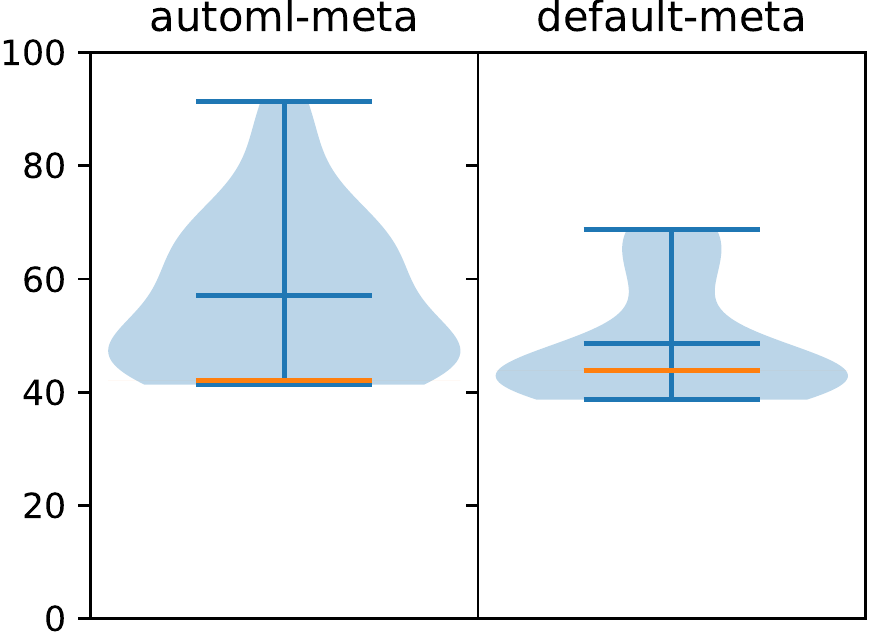}
        }%

   \caption{Violin plots depicting the distributions of mean error rates for 30 predictors. Each sub-figure, associated with a dataset, displays the distribution for \textit{automl-meta} and \textit{default-meta} on its left and right, respectively. The top, middle and bottom blue horizontal bars denote the maximum, mean and minimum of each distribution, respectively. The orange horizontal bars mark the highest mean error rate of predictors in the best performing group; see Table~\ref{tab:top_predictors_default}.}
    \label{fig:violin_plots_n_predictors}
\end{figure}

The obvious approach is to inspect how the mean performances of predictors, determined opportunistically and systematically within \textit{automl-meta} and \textit{default-meta}, respectively, are distributed.
These distributions are depicted per dataset in Fig.~\ref{fig:violin_plots_n_predictors}.
As the widths of the violin plots are normalised, predictors without evaluations result in distribution densities of reduced size, e.g.~\textit{kddcup} for \textit{default-meta} or \textit{dorothea} for \textit{automl-meta}.
However, what is remarkable is that the shape of the distribution is similar between meta-knowledge bases for so many datasets, even when Fig.~\ref{fig:default_and_automl_correlations_with_without_meta_predictors} would otherwise indicate the error-rate correlations are low, e.g.~for \textit{gisette}.
This outcome suggests that, while the nature of guessing hyperparameters induces variability within mean-performance evaluations, maybe generating substantial noise, these individual effects seem to wash out on the collective level.
Essentially, some quality intrinsic to each dataset impels predictors to perform statistically at a reasonably consistent spread of levels.
We call this intrinsic quality the `challenge' of a dataset.

Quantifying and associating challenge, even loosely, is another matter.
What is apparent in Fig.~\ref{fig:violin_plots_n_predictors} is that mean-performance distributions typically come in two types: bottom-heavy and top-heavy.
For bottom-heavy distributions, any random sampling of predictors is likely to do well; these datasets can be considered `easy', where the choice of ML algorithm employed barely matters.
We stress again the nuance of terminology here; \textit{abalone} and \textit{kddcup} are both considered `easy' by this definition, even though no predictor does better than 70\% loss for the former, while many predictors struggle to run until completion for the latter.
The point here is that most ML algorithms that solve an `easy' problem perform relatively well.
In contrast, predictor selection matters for top-heavy distributions, suggesting that specialist techniques are required to push the limits of accuracy for associated `hard' datasets.

As a start, we define a simple skewness factor as one measure of dataset challenge, 
\begin{equation}
\label{eq:skewness}
    (d_{\textrm{mean}}-d_{\textrm{min}})/(d_{\textrm{max}}-d_{\textrm{min}}),
\end{equation}
where the $d$ variables represent the minimum, mean and maximum of each distribution in Fig.~\ref{fig:violin_plots_n_predictors}.
Any value less than 0.5 is bottom-heavy, while any value greater than 0.5 is top-heavy.
Immediately, it becomes evident that the seemingly representative sampling of 20 datasets may not be the best benchmark for configuration-space reduction approaches.
The only consistently `hard' datasets are \textit{amazon}, \textit{convex} and \textit{madelon}, with skewness values for \{\textit{automl-meta}, \textit{default-meta}\} of \{0.65, 0.61\}, \{0.68, 0.73\} and \{0.67, 0.59\}, respectively.
This list is expanded by \textit{cifar10small} for \textit{automl-meta}, with a skewness of 0.63, although the scope of this work makes it unclear whether additional optimisation time would have nudged its predictor error rates lower to better resemble the bottom-heavy \textit{default-meta} distribution.
Conversely, perhaps SMAC had enough time to achieve unusual breakthroughs, especially given that \textit{cifar10small} is not only rare for the minimum of \textit{automl-meta} improving upon the minimum of \textit{default-meta}, it is also unique for the extent of that improvement.

\begin{figure}[!ht] 
  \centering 
        \includegraphics[width=0.99\linewidth]{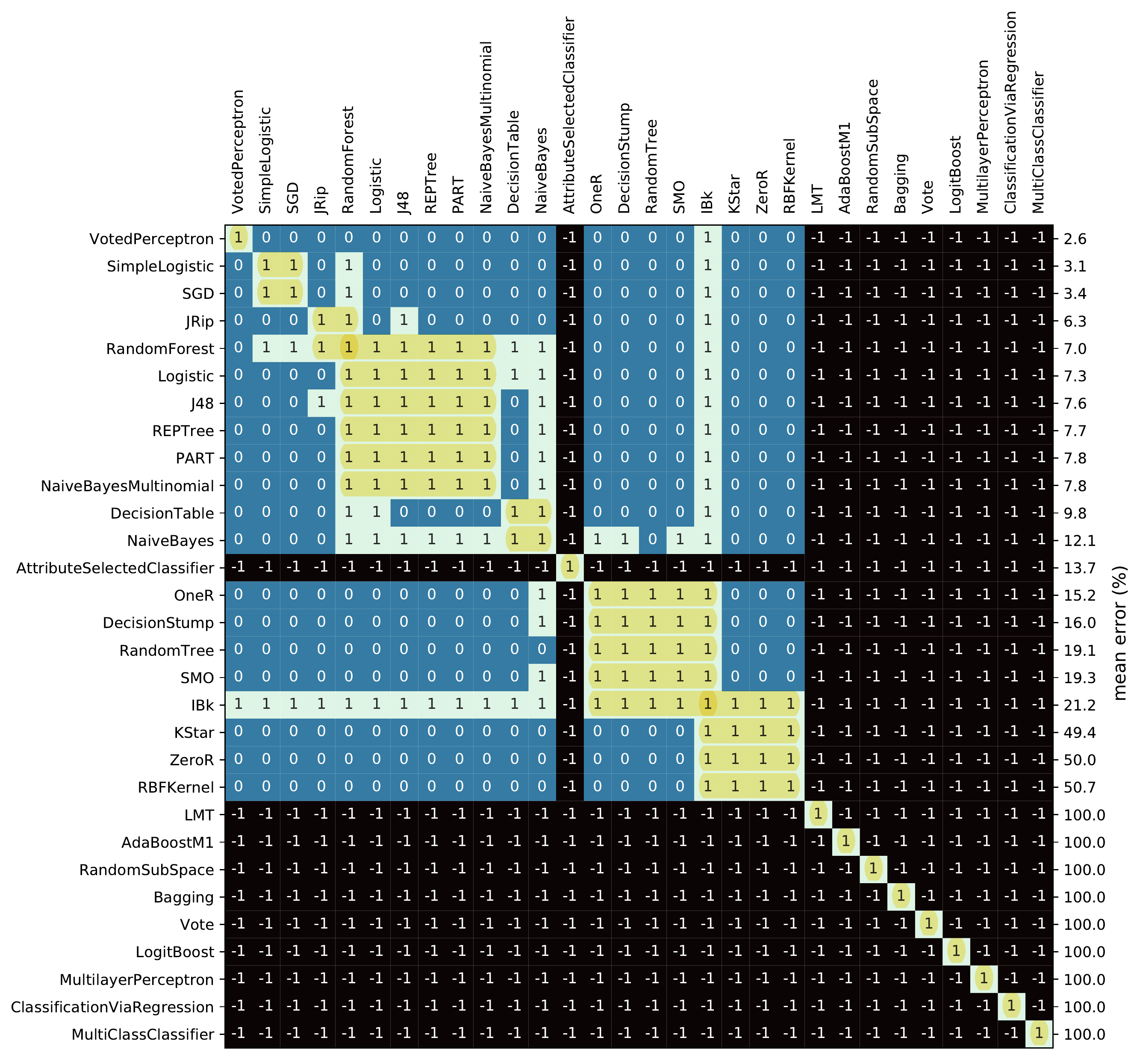}
    \caption{Example matrix for \textit{gisette} denoting which pairs of predictors have no significant differences between their single-fold evaluation distributions in \textit{automl-meta}. Matrix values are based on Welch's t-test, where a lack of significant difference, i.e.~a p-value above 0.95, is marked with 1. Significant differences and failures to test them, due to sample sizes, are marked 0 and -1, respectively. Predictors are ordered by mean error-rate, while yellow squares cluster groups of statistically inseparable performers.}
    \label{fig:heatmap_automl_predictor_per_dataset}
\end{figure}

\begin{figure}[!ht]
  \centering 
        \includegraphics[width=0.99\linewidth]{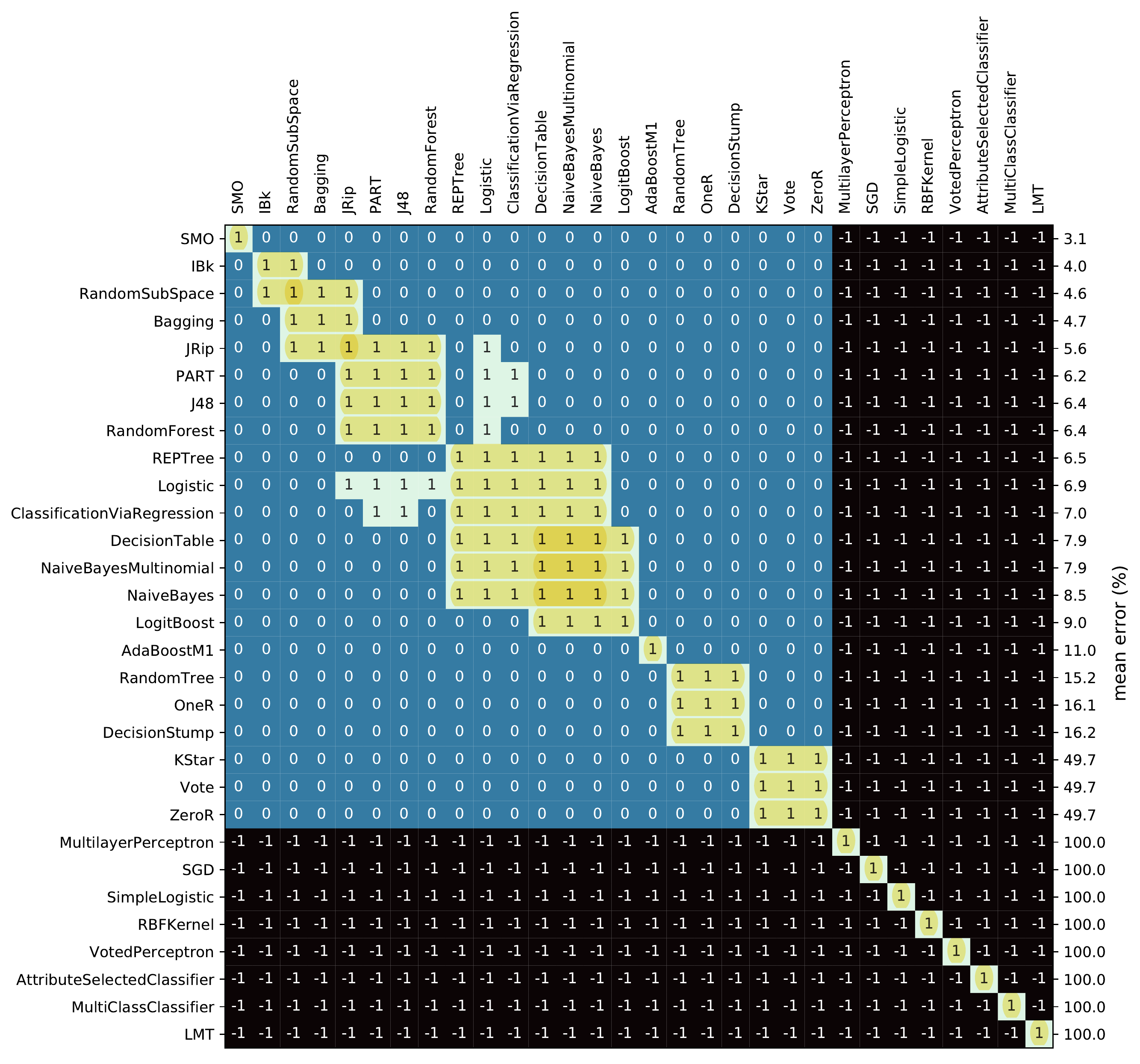}
    \caption{Example matrix for \textit{gisette} denoting which pairs of predictors have no significant differences between their single-fold evaluation distributions in \textit{default-meta}. Matrix values are based on Welch's t-test, where a lack of significant difference, i.e.~a p-value above 0.95, is marked with 1. Significant differences and failures to test them, due to sample sizes, are marked 0 and -1, respectively. Predictors are ordered by mean error-rate, while yellow squares cluster groups of statistically inseparable performers.}
    \label{fig:heatmap_default_predictor_per_dataset}
\end{figure}

Admittedly, skewness is only a high-level metric for assessing the challenge of a dataset.
It is entirely possible that, despite many ML algorithms doing comparably well on an easy problem, there may still be uniquely best-performing predictors.
Whether they are worth pursuing over the other almost-best performers is another question, dependent on what performance ML means to stakeholders running an ML application.
Nonetheless, within the bounds of this research, we merely identify the group of most-accurate predictors per dataset and examine its size.

Establishing best-performing groups of predictors is non-trivial, relying on significance testing.
In essence, for every pair of predictors, the distributions of their single-fold evaluations, exemplified for \textit{automl-meta} in Fig.~\ref{fig:violin_chart_3types_apart}, are assessed as to whether they are likely to be indistinguishable, i.e.~neither predictor is better than the other.
In the case of \textit{automl-meta}, different predictors will have different numbers of evaluations, as well as different variances, so we choose to use Welch's t-test.
Granted, Welch's t-test is not perfect for the task, especially given its assumption of normality for the distributions involved.
Whether predictor performance is normal over variations in CV fold is already one question, i.e.~in \textit{default-meta}, so allowing variations in hyperparameters, i.e.~in \textit{automl-meta}, further weakens the assumption.
Nonetheless, it is the best available option for significance testing in the scope of this work, and arguments in the literature suggest that the t-test is relatively robust to deviations from normality anyway, especially for large sample sizes.

Once an indistinguishability matrix is generated, such as Fig.~\ref{fig:heatmap_automl_predictor_per_dataset} exemplifies for \textit{gisette} in \textit{automl-meta}, it is possible to cluster statistically identical performers together.
However, there are a few complications.
For instance, as the \textit{gisette} matrix shows in rankings three to five, the performance of RandomForest is indistinguishable from both SGD and JRip, but SGD is significantly better -- it is both different and has a lower average loss -- than JRip.
With the mean performance of RandomForest being worse than either SGD or JRip, we decide to group it with the latter.
Given such conventions, the yellow squares in Fig.~\ref{fig:heatmap_automl_predictor_per_dataset} denote the actual groups of statistically indistinguishable performers, sometimes overlapping, with the yellow square in the top left corner representing the best-performing predictors, i.e.~only VotedPerceptron in this case.
That stated, it is unclear how robust this methodology is for \textit{automl-meta}.
Beyond the question of normality, sample sizes and evaluation spreads based on hyperparameter ranges result in awkward behaviour.
For instance, while unevaluated predictors with 100\% error cannot be compared with any other distribution, thus being marked by a row and column of -1 in an indistinguishability matrix, \textit{gisette} exemplifies a predictor, AttributeSelectedClassifier, that is incomparable due to a sample size of one.
Meanwhile, IBk, which Fig.~\ref{fig:f1_automl_number_of_evaluations_heatmap} shows has been evaluated twice for \textit{gisette}, possesses substantial spread and insubstantial sampling; the matrix indicates no significant differences against any other predictor.

\begin{landscape}

\begin{table*}[htb!]
\centering
\caption {The size (n\_preds) of both the best and second-best performing groups of predictors, per dataset, for \textit{default-meta}. The minimum, maximum and mean of 10-fold CV error rate for each group of predictors is displayed, as well as the number of predictors that sit in both groups (n\_overlap). The median and mean performance of all 30 predictors is also shown. Skewness factors provide comparisons with an alternate `challenge' metric based on the distribution of averaged predictor performances, per dataset, for both \textit{automl-meta} and \textit{default-meta}; $<$ 0.5 is bottom-heavy and $>$ 0.5 is top-heavy. The final columns detail the probability of selecting a best-performing predictor if $k$ out of 30 predictors are randomly selected.}
\label{tab:top_predictors_default}
\fontsize{7.2}{8.5}\selectfont

\begin{tabular}{|l|ll|ll|l|l|l|ll|lllll|}
\hline
\multirow{2}{*}{\textbf{Dataset}} & \multicolumn{2}{l|}{\textbf{\begin{tabular}[c]{@{}l@{}}The best\\ performing\\ group\end{tabular}}} & \multicolumn{2}{l|}{\textbf{\begin{tabular}[c]{@{}l@{}}The second best\\ performing\\ group\end{tabular}}} & \multirow{2}{*}{\textbf{\begin{tabular}[c]{@{}l@{}}median\\ error\end{tabular}}} & \multirow{2}{*}{\textbf{\begin{tabular}[c]{@{}l@{}}mean\\ error\end{tabular}}} & \multirow{2}{*}{\textbf{n\_overlap}} & \multicolumn{2}{l|}{\textbf{skewness}}                 & \multicolumn{5}{l|}{\textbf{\begin{tabular}[c]{@{}l@{}}The probability of at least\\ one best predictor\\ in the search space\end{tabular}}}              \\ \cline{2-5} \cline{9-15} 
                                  & \multicolumn{1}{l|}{\textbf{n\_preds}}                 & \textbf{mean {[}min-max{]}}                & \multicolumn{1}{l|}{\textbf{n\_preds}}                    & \textbf{mean {[}min-max{]}}                    &                                                                                  &                                                                                &                                      & \multicolumn{1}{l|}{\textbf{automl}} & \textbf{default} & \multicolumn{1}{l|}{\textbf{k1}} & \multicolumn{1}{l|}{\textbf{k4}} & \multicolumn{1}{l|}{\textbf{k8}} & \multicolumn{1}{l|}{\textbf{k10}} & \textbf{k19} \\ \hline
\textbf{car}                      & \multicolumn{1}{l|}{1}                                 & 1.24 {[}1.24-1.24{]}                       & \multicolumn{1}{l|}{3}                                    & 5.01 {[}4.79-5.21{]}                           & 13.18                                                                            & 15.07                                                                          & 0                                    & \multicolumn{1}{l|}{0.47}            & 0.48             & \multicolumn{1}{l|}{3\%}         & \multicolumn{1}{l|}{13\%}        & \multicolumn{1}{l|}{27\%}        & \multicolumn{1}{l|}{33\%}         & 63\%         \\ \hline
\textbf{convex}                   & \multicolumn{1}{l|}{1}                                 & 24.33 {[}24.33-24.33{]}                    & \multicolumn{1}{l|}{1}                                    & 27.93 {[}27.93-27.93{]}                        & 46.89                                                                            & 42.89                                                                          & 0                                    & \multicolumn{1}{l|}{0.68}            & 0.73             & \multicolumn{1}{l|}{3\%}         & \multicolumn{1}{l|}{13\%}        & \multicolumn{1}{l|}{27\%}        & \multicolumn{1}{l|}{33\%}         & 63\%         \\ \hline
\textbf{gisette}                  & \multicolumn{1}{l|}{1}                                 & 3.08 {[}3.08-3.08{]}                       & \multicolumn{1}{l|}{2}                                    & 4.27 {[}3.96-4.57{]}                           & 7.42                                                                             & 13.73                                                                          & 0                                    & \multicolumn{1}{l|}{0.36}            & 0.25             & \multicolumn{1}{l|}{3\%}         & \multicolumn{1}{l|}{13\%}        & \multicolumn{1}{l|}{27\%}        & \multicolumn{1}{l|}{33\%}         & 63\%         \\ \hline
\textbf{mnist}                    & \multicolumn{1}{l|}{1}                                 & 4.07 {[}4.07-4.07{]}                       & \multicolumn{1}{l|}{1}                                    & 4.66 {[}4.66-4.66{]}                           & 16.46                                                                            & 29.47                                                                          & 0                                    & \multicolumn{1}{l|}{0.43}            & 0.30             & \multicolumn{1}{l|}{3\%}         & \multicolumn{1}{l|}{13\%}        & \multicolumn{1}{l|}{27\%}        & \multicolumn{1}{l|}{33\%}         & 63\%         \\ \hline
\textbf{amazon}                   & \multicolumn{1}{l|}{2}                                 & 30.14 {[}29.14-31.14{]}                    & \multicolumn{1}{l|}{2}                                    & 32.33 {[}31.14-33.52{]}                        & 70.19                                                                            & 70.49                                                                          & 1                                    & \multicolumn{1}{l|}{0.65}            & 0.61             & \multicolumn{1}{l|}{7\%}         & \multicolumn{1}{l|}{25\%}        & \multicolumn{1}{l|}{47\%}        & \multicolumn{1}{l|}{56\%}         & 87\%         \\ \hline
\textbf{cifar10small}             & \multicolumn{1}{l|}{2}                                 & 66.28 {[}66.14-66.42{]}                    & \multicolumn{1}{l|}{1}                                    & 68.21 {[}68.21-68.21{]}                        & 76.03                                                                            & 76.29                                                                          & 0                                    & \multicolumn{1}{l|}{0.63}            & 0.43             & \multicolumn{1}{l|}{7\%}         & \multicolumn{1}{l|}{25\%}        & \multicolumn{1}{l|}{47\%}        & \multicolumn{1}{l|}{56\%}         & 87\%         \\ \hline
\textbf{madelon}                  & \multicolumn{1}{l|}{2}                                 & 22.17 {[}22.09-22.25{]}                    & \multicolumn{1}{l|}{3}                                    & 23.55 {[}22.25-24.29{]}                        & 41.26                                                                            & 38.55                                                                          & 1                                    & \multicolumn{1}{l|}{0.67}            & 0.59             & \multicolumn{1}{l|}{7\%}         & \multicolumn{1}{l|}{25\%}        & \multicolumn{1}{l|}{47\%}        & \multicolumn{1}{l|}{56\%}         & 87\%         \\ \hline
\textbf{winequality}              & \multicolumn{1}{l|}{2}                                 & 37.26 {[}36.57-37.94{]}                    & \multicolumn{1}{l|}{4}                                    & 39.42 {[}37.94-41.38{]}                        & 45.99                                                                            & 47.10                                                                          & 1                                    & \multicolumn{1}{l|}{0.49}            & 0.45             & \multicolumn{1}{l|}{7\%}         & \multicolumn{1}{l|}{25\%}        & \multicolumn{1}{l|}{47\%}        & \multicolumn{1}{l|}{56\%}         & 87\%         \\ \hline
\textbf{semeion}                  & \multicolumn{1}{l|}{3}                                 & 6.12 {[}5.38-6.63{]}                       & \multicolumn{1}{l|}{5}                                    & 10.3 {[}9.77-12.1{]}                           & 21.42                                                                            & 30.66                                                                          & 0                                    & \multicolumn{1}{l|}{0.41}            & 0.30             & \multicolumn{1}{l|}{10\%}        & \multicolumn{1}{l|}{36\%}        & \multicolumn{1}{l|}{62\%}        & \multicolumn{1}{l|}{72\%}         & 96\%         \\ \hline
\textbf{adult}                    & \multicolumn{1}{l|}{5}                                 & 14.19 {[}13.77-14.55{]}                    & \multicolumn{1}{l|}{5}                                    & 14.51 {[}14.28-14.71{]}                        & 15.46                                                                            & 16.95                                                                          & 3                                    & \multicolumn{1}{l|}{0.35}            & 0.31             & \multicolumn{1}{l|}{17\%}        & \multicolumn{1}{l|}{54\%}        & \multicolumn{1}{l|}{82\%}        & \multicolumn{1}{l|}{89\%}         & 100\%        \\ \hline
\textbf{krvskp}                   & \multicolumn{1}{l|}{5}                                 & 0.81 {[}0.67-0.89{]}                       & \multicolumn{1}{l|}{6}                                    & 0.94 {[}0.76-1.21{]}                           & 4.20                                                                             & 9.24                                                                           & 4                                    & \multicolumn{1}{l|}{0.38}            & 0.18             & \multicolumn{1}{l|}{17\%}        & \multicolumn{1}{l|}{54\%}        & \multicolumn{1}{l|}{82\%}        & \multicolumn{1}{l|}{89\%}         & 100\%        \\ \hline
\textbf{waveform}                 & \multicolumn{1}{l|}{5}                                 & 13.31 {[}13.09-13.57{]}                    & \multicolumn{1}{l|}{3}                                    & 16.03 {[}15.06-16.8{]}                         & 20.60                                                                            & 25.49                                                                          & 0                                    & \multicolumn{1}{l|}{0.38}            & 0.23             & \multicolumn{1}{l|}{17\%}        & \multicolumn{1}{l|}{54\%}        & \multicolumn{1}{l|}{82\%}        & \multicolumn{1}{l|}{89\%}         & 100\%        \\ \hline
\textbf{shuttle}                  & \multicolumn{1}{l|}{6}                                 & 0.03 {[}0.02-0.05{]}                       & \multicolumn{1}{l|}{6}                                    & 0.04 {[}0.02-0.06{]}                           & 0.20                                                                             & 3.85                                                                           & 5                                    & \multicolumn{1}{l|}{0.24}            & 0.18             & \multicolumn{1}{l|}{20\%}        & \multicolumn{1}{l|}{61\%}        & \multicolumn{1}{l|}{87\%}        & \multicolumn{1}{l|}{93\%}         & 100\%        \\ \hline
\textbf{dexter}                   & \multicolumn{1}{l|}{7}                                 & 8.33 {[}6.67-10.24{]}                      & \multicolumn{1}{l|}{8}                                    & 9.64 {[}8.1-12.14{]}                           & 14.52                                                                            & 20.49                                                                          & 5                                    & \multicolumn{1}{l|}{0.41}            & 0.31             & \multicolumn{1}{l|}{27\%}        & \multicolumn{1}{l|}{73\%}        & \multicolumn{1}{l|}{95\%}        & \multicolumn{1}{l|}{98\%}         & 100\%        \\ \hline
\textbf{kddcup}                   & \multicolumn{1}{l|}{7}                                 & 1.8 {[}1.8-1.8{]}                          & \multicolumn{1}{l|}{1}                                    & 2 {[}2-2{]}                                    & 1.80                                                                             & 2.50                                                                           & 0                                    & \multicolumn{1}{l|}{0.08}            & 0.14             & \multicolumn{1}{l|}{23\%}        & \multicolumn{1}{l|}{68\%}        & \multicolumn{1}{l|}{92\%}        & \multicolumn{1}{l|}{96\%}         & 100\%        \\ \hline
\textbf{abalone}                  & \multicolumn{1}{l|}{9}                                 & 73.93 {[}73.08-74.9{]}                     & \multicolumn{1}{l|}{5}                                    & 75.27 {[}74.9-75.44{]}                         & 76.50                                                                            & 77.10                                                                          & 1                                    & \multicolumn{1}{l|}{0.24}            & 0.39             & \multicolumn{1}{l|}{30\%}        & \multicolumn{1}{l|}{78\%}        & \multicolumn{1}{l|}{97\%}        & \multicolumn{1}{l|}{99\%}         & 100\%        \\ \hline
\textbf{yeast}                    & \multicolumn{1}{l|}{13}                                & 41.39 {[}38.69-43.79{]}                    & \multicolumn{1}{l|}{13}                                   & 41.79 {[}40.42-43.79{]}                        & 44.23                                                                            & 48.68                                                                          & 12                                   & \multicolumn{1}{l|}{0.32}            & 0.33             & \multicolumn{1}{l|}{43\%}        & \multicolumn{1}{l|}{91\%}        & \multicolumn{1}{l|}{100\%}       & \multicolumn{1}{l|}{100\%}        & 100\%        \\ \hline
\textbf{dorothea}                 & \multicolumn{1}{l|}{13}                                & 7.49 {[}6.71-8.82{]}                       & \multicolumn{1}{l|}{6}                                    & 8.47 {[}7.60-9.70{]}                           & 8.07                                                                             & 11.00                                                                          & 4                                    & \multicolumn{1}{l|}{0.19}            & 0.10             & \multicolumn{1}{l|}{43\%}        & \multicolumn{1}{l|}{91\%}        & \multicolumn{1}{l|}{100\%}       & \multicolumn{1}{l|}{100\%}        & 100\%        \\ \hline
\textbf{gcredit}                  & \multicolumn{1}{l|}{15}                                & 25.06 {[}24.14-26.57{]}                    & \multicolumn{1}{l|}{9}                                    & 26.51 {[}25.29-27.43{]}                        & 26.79                                                                            & 27.30                                                                          & 5                                    & \multicolumn{1}{l|}{0.28}            & 0.22             & \multicolumn{1}{l|}{50\%}        & \multicolumn{1}{l|}{95\%}        & \multicolumn{1}{l|}{100\%}       & \multicolumn{1}{l|}{100\%}        & 100\%        \\ \hline
\textbf{secom}                    & \multicolumn{1}{l|}{16}                                & 6.23 {[}6.11-6.75{]}                       & \multicolumn{1}{l|}{6}                                    & 6.75 {[}6.47-7.02{]}                           & 6.56                                                                             & 10.02                                                                          & 3                                    & \multicolumn{1}{l|}{0.09}            & 0.06             & \multicolumn{1}{l|}{53\%}        & \multicolumn{1}{l|}{96\%}        & \multicolumn{1}{l|}{100\%}       & \multicolumn{1}{l|}{100\%}        & 100\%        \\ \hline
\end{tabular}

\end{table*}

\end{landscape}

The sampling of predictor performance in \textit{default-meta} tends to be much more uniform, which provides better-behaved significance tests, as exemplified by Fig.~\ref{fig:heatmap_default_predictor_per_dataset} for \textit{gisette}.
The drawback is, of course, clear; \textit{default-meta} matrices determine which predictors are significantly better than others only for default hyperparameters.
Nonetheless, given that mean-performance distributions have similar shape between \textit{automl-meta} and \textit{default-meta}, this may provide a sufficient gauge of dataset challenge.
Accordingly, Table~\ref{tab:top_predictors_default} describes the best and second-best groups of indistinguishable predictors per dataset in \textit{default-meta}.
The datasets are ordered by the number of best predictors, such that \textit{secom} at the bottom is considered `easiest', with 16 ML algorithms statistically capable of reaching a loss of ${\sim}6.23\%$, while \textit{car}, \textit{convex}, \textit{gisette} and \textit{mnist} at the top are considered `hardest', with only one front-runner for each being capable of achieving statistically better results than the other evaluated candidate predictors.

There are several key points to take from Table~\ref{tab:top_predictors_default}.
First of all, with skewness defined for the distribution of mean predictor accuracies in Eq.~(\ref{eq:skewness}), every dataset with skewness below 0.25, for either \textit{automl-meta} or \textit{default-meta}, is in the `easiest' 11 datasets.
Every dataset with skewness above 0.41 is in the `hardest' nine.
These include the top-heavy datasets identified earlier: \textit{convex}, \textit{amazon}, \textit{cifar10small}, and \textit{madelon}.
Thus, there is some consistency between both metrics for dataset challenge.
Separately, we also examined the best-performing predictor groups for \textit{automl-meta} and found that their sizes are equal or smaller than for \textit{default-meta}, with the exception of \textit{mnist} ($1\rightarrow 2$) and \textit{kddcup} ($7\rightarrow 10$).
This result suggests that \textit{automl-meta} rankings may be more discriminative, but, at the very least, `hard' datasets are robustly identified as such, regardless of meta-knowledge base.

Importantly, Table~\ref{tab:top_predictors_default} also examines what happens if a predictor pool is randomly reduced to size $k$.
Evidently, the number of predictors in the best-performing group does not have to grow much before random culling has reasonable odds of nominating an ML algorithm from the top performing group within the reduced pool.
For $k=8$, all datasets beyond \textit{car}, \textit{convex}, \textit{gisette} and \textit{mnist} will have about a 1-in-2 chance, or better, of yielding a top predictor, as judged by \textit{default-meta}, in an AutoML search space.
For the `easiest' 11 datasets, the same odds are achievable by picking only four out of 30 predictors.
This outcome reinforces the message that benchmark collections of datasets must be chosen carefully if they are to showcase the power of configuration-space reduction strategies based on meta-learning.
After all, if the fast and uninformed method of randomly culling a search space has good odds of recommending an optimal ML solution, more complex strategies become less appealing.
Similarly, it also becomes difficult to identify whether a novel culling strategy is intrinsically powerful or simply lucky for `easy' datasets.
For these reasons, having identified the issues that datasets of differing challenge can pose to search-space reduction strategies, the next step is to identify the performances that one should actually expect from these strategies.

\subsection{Evaluating Reduction Strategies: Against the Oracle}
\label{Sec:EvalOracle}

Clearly, neither the opportunistic \textit{automl-meta} nor the systematic \textit{default-meta} compiled in this work are exhaustive curations of meta-knowledge.
They are practical approaches for balancing the construction/maintenance of historical knowledge bases with the day-to-day processing of ML applications.
As such, any associated predictor rankings are likely to be imperfect, biased by the sampling choices that SMAC and human experts make.
Nonetheless, for the sake of an ablative approach, this section adopts a current working assumption that the rankings per dataset reflect a ground truth.
Recalling the 33 strategies in Section~\ref{Sec:Strategies}, this means that, for any reduced search-space size of $k=n$, the oracle strategy of \textit{OX-kn} should be the ideal method of predictor pool construction.
Technically, \textit{O1-kn} from \textit{automl-meta} and \textit{O2-kn} from \textit{default-meta} result in differing search spaces, so one can further hypothesise that \textit{O1-kn} should truly be the best, on account of being compiled with the same SMAC procedure that is used to test the strategies in Section~\ref{sec:results}.
However, for the current ablation analysis, we do not deal with this nuance here; \textit{OX-kn} is perfect within its category of meta-knowledge base $X$ and reduced size $k=n$.
Of course, barring reruns, most real-world applications encountering a novel dataset have no idea which ML algorithms performed well on it previously.
An oracle strategy is an idealised approach, unlikely to be employed in practice.
So, if the only options are landmarked and global-leaderboard strategies, i.e.~\textit{LX-kn} and \textit{MX-kn}, how would their expected performances compare with the `perfect' oracle?

\begin{figure}[!ht]
  \centering 
    \subfloat[automl-meta]{
        \includegraphics[width=0.99\linewidth]{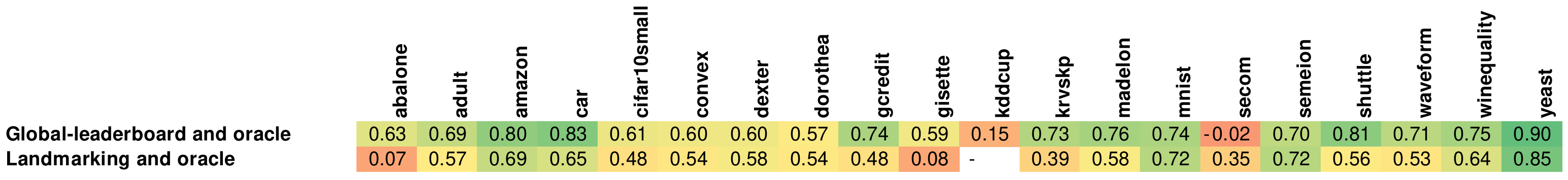}
      } \\
     \subfloat[default-meta]{
        \includegraphics[width=0.99\linewidth]{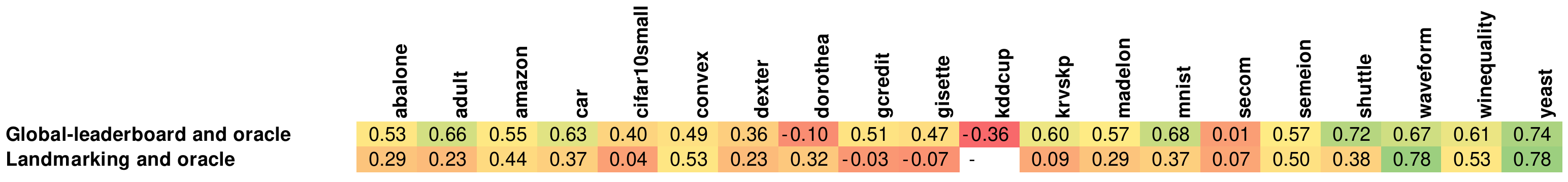}
      } 
      
    \caption{Correlations between the rankings of 30 predictors for a specific dataset, i.e.~oracle, and their rankings for (1) the most similar dataset, i.e.~landmarked, and (2) the dataset collection as a whole, i.e.~global leaderboard. Ranking correlations are calculated separately for \textit{automl-meta} and \textit{default-meta}.}
    \label{fig:correlation_expected_lmo_automl}
\end{figure}

\begin{table}[htb!]
\centering
\caption {Most similar dataset for every ML problem according to the method of landmarking used in this work. Landmarked similarity depends on the mean error rates within a meta-knowledge base, i.e.~\textit{automl-meta} (L1) or \textit{default-meta} (L2). Failure for the five relevant landmarkers to complete an evaluation within two hours results in missing values, i.e.~\textit{kddcup}.}
\label{tab:tab_dataset_landmarking_similarity}
\fontsize{7.2}{8.5}\selectfont

\begin{tabular}{|l|l|l|}
\hline
\textbf{Dataset}      & \textbf{L1}  & \textbf{L2} \\ \hline
\textbf{abalone}      & gisette      & madelon     \\ \hline
\textbf{adult}        & waveform     & dexter      \\ \hline
\textbf{amazon}       & semeion      & dexter      \\ \hline
\textbf{car}          & mnist        & mnist       \\ \hline
\textbf{cifar10small} & semeion      & yeast       \\ \hline
\textbf{convex}       & mnist        & winequality \\ \hline
\textbf{dexter}       & cifar10small & adult       \\ \hline
\textbf{dorothea}     & madelon      & kddcup      \\ \hline
\textbf{gcredit}      & amazon       & amazon      \\ \hline
\textbf{gisette}      & waveform     & waveform    \\ \hline
\textbf{kddcup}       & -            & -           \\ \hline
\textbf{krvskp}       & mnist        & mnist       \\ \hline
\textbf{madelon}      & gisette      & abalone     \\ \hline
\textbf{mnist}        & semeion      & car         \\ \hline
\textbf{secom}        & kddcup       & winequality \\ \hline
\textbf{semeion}      & mnist        & mnist       \\ \hline
\textbf{shuttle}      & mnist        & gisette     \\ \hline
\textbf{waveform}     & mnist        & yeast       \\ \hline
\textbf{winequality}  & mnist        & convex      \\ \hline
\textbf{yeast}        & semeion      & waveform    \\ \hline
\end{tabular}

\end{table}

Anticipating the real-world absence of oracular predictor rankings for a specific dataset, the first step is to examine how closely the collection-averaged rankings and the rankings for the most similar dataset, i.e.~the global leaderboard and landmarked rankings, respectively, reflect the ground truth.
Technically, the global leaderboard does include oracular rankings in its averaging process, but we found separately that their exclusion does not seem to matter; every leaderboard compiled for 19 datasets had a correlation value of over 0.99 with the leaderboard based on all 20 datasets.
Acknowledging this fact, comparisons between oracular rankings and those that inform both \textit{MX-kn} and \textit{LX-kn} are displayed in Fig.~\ref{fig:correlation_expected_lmo_automl}.
As a side note, the most similar dataset to any particular ML problem is determined as part of a landmarked reduction strategy.
For completeness, these most similar datasets are listed out in Table~\ref{tab:tab_dataset_landmarking_similarity}, calculated independently for each meta-knowledge base, i.e.~\textit{automl-meta} and \textit{default-meta}.
Given that the five chosen landmarkers in this work were unable to solve \textit{kddcup} within even two hours, all associated \textit{LX-kn} results are left blank both here and from now on.

Of the key points to take away from the correlations in Fig.~\ref{fig:correlation_expected_lmo_automl}, the most apparent one is that the global leaderboard has generally better fidelity to the ground truth than any landmarked rankings.
The exceptions in \textit{automl-meta} are for \textit{secom} and, barely, \textit{semeion}.
There are a few more in \textit{default-meta}, i.e.~\textit{convex}, \textit{dorothea}, \textit{secom}, \textit{waveform} and \textit{yeast}, although the discrepancies vary in magnitude and the point still stands.
These results are slightly surprising, as preliminary results based on a subset of the 20 datasets in this work suggested that, on average, landmarked strategies do better than those driven by the global leaderboard~\citep{ngke21}.
At the very least, a good landmarking approach should hew close to oracular rankings for any ML problems that deviate from the norm, and this is only really the case for \textit{secom}.
Perhaps, for most of these 20 datasets, they are all more similar to their `average', some pseudo-dataset corresponding to the global leaderboard, than any other ML problem in the benchmarking collection.
In other words, for any particular dataset, no other ML problem may be sufficiently similar to provide useful meta-knowledge, which is entirely possible if either the benchmarking collection is small or very diverse.
Alternatively, the landmarking method used in this work may simply be poor at identifying similarities.
Whatever the case, this outcome emphasises how similarity-based meta-learning, while theoretically ideal, is challenging to set up; its effectiveness is hard to guarantee, and this can make it less appealing compared to simpler alternatives.

Admittedly, there is a large spread in correlation values within Fig.~\ref{fig:correlation_expected_lmo_automl}, even for the global leaderboard.
At one extreme, any strategy is likely to match the oracle well for \textit{yeast}, while, at the other, \textit{secom} in \textit{automl-meta} and both \textit{dorothea} and \textit{kddcup} in \textit{default-meta} are all poorly represented by the global leaderboard.
Curiously, with the exception of landmarked-versus-oracular rankings for both \textit{abalone} and \textit{waveform}, correlations with the respective oracle are also found to be poorer across the board for \textit{default-meta} as compared to \textit{automl-meta}.
It is not clear why this is the case, but the consistency of poor rankings for meta-predictors in \textit{automl-meta} are likely the culprit.


\begin{table*}[htb!]
\centering
\caption {Expected average loss for oracle strategies, \textit{eOX-kn}, applied to each dataset. Each value is the average of the mean error rates for the best $k=n$ predictors specific to a dataset, as determined from a meta-knowledge base, where $X=1$ denotes \textit{automl-meta} and $X=2$ denotes \textit{default-meta}.}
\label{tab:tab_expected_oracle}
\fontsize{7.2}{8.5}\selectfont

\begin{tabular}{|l|llllll|llllll|}
\hline
\textbf{Strategies} & \multicolumn{6}{c|}{\textbf{eO1}}                                                                                                                                                             & \multicolumn{6}{c|}{\textbf{eO2}}                                                                                                                                                             \\ \hline
\textbf{k values}   & \multicolumn{1}{l|}{\textbf{k1}} & \multicolumn{1}{l|}{\textbf{k4}} & \multicolumn{1}{l|}{\textbf{k8}} & \multicolumn{1}{l|}{\textbf{k10}} & \multicolumn{1}{l|}{\textbf{k19}} & \textbf{k30} & \multicolumn{1}{l|}{\textbf{k1}} & \multicolumn{1}{l|}{\textbf{k4}} & \multicolumn{1}{l|}{\textbf{k8}} & \multicolumn{1}{l|}{\textbf{k10}} & \multicolumn{1}{l|}{\textbf{k19}} & \textbf{k30} \\ \hline
abalone             & \multicolumn{1}{l|}{73.92}       & \multicolumn{1}{l|}{74.43}       & \multicolumn{1}{l|}{74.92}       & \multicolumn{1}{l|}{75.09}        & \multicolumn{1}{l|}{76.37}        & 82.18        & \multicolumn{1}{l|}{73.08}       & \multicolumn{1}{l|}{73.34}       & \multicolumn{1}{l|}{73.81}       & \multicolumn{1}{l|}{74.06}        & \multicolumn{1}{l|}{75.24}        & 78.62        \\ \hline
adult               & \multicolumn{1}{l|}{13.86}       & \multicolumn{1}{l|}{14.36}       & \multicolumn{1}{l|}{14.91}       & \multicolumn{1}{l|}{15.13}        & \multicolumn{1}{l|}{17.82}        & 28.11        & \multicolumn{1}{l|}{13.77}       & \multicolumn{1}{l|}{14.11}       & \multicolumn{1}{l|}{14.40}       & \multicolumn{1}{l|}{14.50}        & \multicolumn{1}{l|}{15.00}        & 19.72        \\ \hline
amazon              & \multicolumn{1}{l|}{36.19}       & \multicolumn{1}{l|}{46.53}       & \multicolumn{1}{l|}{56.28}       & \multicolumn{1}{l|}{60.79}        & \multicolumn{1}{l|}{76.00}        & 84.78        & \multicolumn{1}{l|}{29.14}       & \multicolumn{1}{l|}{35.14}       & \multicolumn{1}{l|}{44.37}       & \multicolumn{1}{l|}{49.44}        & \multicolumn{1}{l|}{64.71}        & 78.36        \\ \hline
car                 & \multicolumn{1}{l|}{7.06}        & \multicolumn{1}{l|}{9.20}        & \multicolumn{1}{l|}{11.26}       & \multicolumn{1}{l|}{12.39}        & \multicolumn{1}{l|}{18.68}        & 32.30        & \multicolumn{1}{l|}{1.24}        & \multicolumn{1}{l|}{4.07}        & \multicolumn{1}{l|}{6.19}        & \multicolumn{1}{l|}{6.90}         & \multicolumn{1}{l|}{9.83}         & 20.73        \\ \hline
cifar10small        & \multicolumn{1}{l|}{54.20}       & \multicolumn{1}{l|}{63.83}       & \multicolumn{1}{l|}{68.35}       & \multicolumn{1}{l|}{69.77}        & \multicolumn{1}{l|}{77.62}        & 85.58        & \multicolumn{1}{l|}{66.14}       & \multicolumn{1}{l|}{67.63}       & \multicolumn{1}{l|}{69.70}       & \multicolumn{1}{l|}{70.96}        & \multicolumn{1}{l|}{76.29}        & 85.77        \\ \hline
convex              & \multicolumn{1}{l|}{28.18}       & \multicolumn{1}{l|}{32.08}       & \multicolumn{1}{l|}{35.83}       & \multicolumn{1}{l|}{37.69}        & \multicolumn{1}{l|}{42.20}        & 54.92        & \multicolumn{1}{l|}{24.33}       & \multicolumn{1}{l|}{28.05}       & \multicolumn{1}{l|}{32.13}       & \multicolumn{1}{l|}{33.94}        & \multicolumn{1}{l|}{40.00}        & 46.69        \\ \hline
dexter              & \multicolumn{1}{l|}{7.91}        & \multicolumn{1}{l|}{9.54}        & \multicolumn{1}{l|}{13.15}       & \multicolumn{1}{l|}{15.09}        & \multicolumn{1}{l|}{26.96}        & 51.05        & \multicolumn{1}{l|}{6.67}        & \multicolumn{1}{l|}{7.74}        & \multicolumn{1}{l|}{8.57}        & \multicolumn{1}{l|}{9.17}         & \multicolumn{1}{l|}{13.03}        & 31.10        \\ \hline
dorothea            & \multicolumn{1}{l|}{6.78}        & \multicolumn{1}{l|}{7.16}        & \multicolumn{1}{l|}{7.52}        & \multicolumn{1}{l|}{7.84}         & \multicolumn{1}{l|}{10.34}        & 55.17        & \multicolumn{1}{l|}{6.71}        & \multicolumn{1}{l|}{6.93}        & \multicolumn{1}{l|}{7.13}        & \multicolumn{1}{l|}{7.24}         & \multicolumn{1}{l|}{8.52}         & 34.73        \\ \hline
gcredit             & \multicolumn{1}{l|}{23.77}       & \multicolumn{1}{l|}{24.79}       & \multicolumn{1}{l|}{25.65}       & \multicolumn{1}{l|}{26.01}        & \multicolumn{1}{l|}{27.73}        & 32.83        & \multicolumn{1}{l|}{24.14}       & \multicolumn{1}{l|}{24.18}       & \multicolumn{1}{l|}{24.41}       & \multicolumn{1}{l|}{24.63}        & \multicolumn{1}{l|}{25.52}        & 27.30        \\ \hline
gisette             & \multicolumn{1}{l|}{2.61}        & \multicolumn{1}{l|}{3.84}        & \multicolumn{1}{l|}{5.62}        & \multicolumn{1}{l|}{6.05}         & \multicolumn{1}{l|}{12.44}        & 41.24        & \multicolumn{1}{l|}{2.96}        & \multicolumn{1}{l|}{3.64}        & \multicolumn{1}{l|}{4.44}        & \multicolumn{1}{l|}{5.03}         & \multicolumn{1}{l|}{7.35}         & 33.50        \\ \hline
kddcup              & \multicolumn{1}{l|}{1.70}        & \multicolumn{1}{l|}{1.76}        & \multicolumn{1}{l|}{1.79}        & \multicolumn{1}{l|}{1.80}         & \multicolumn{1}{l|}{3.49}         & 51.74        & \multicolumn{1}{l|}{1.80}        & \multicolumn{1}{l|}{1.80}        & \multicolumn{1}{l|}{1.82}        & \multicolumn{1}{l|}{2.50}         & \multicolumn{1}{l|}{2.50}         & 67.50        \\ \hline
krvskp              & \multicolumn{1}{l|}{1.29}        & \multicolumn{1}{l|}{2.10}        & \multicolumn{1}{l|}{3.09}        & \multicolumn{1}{l|}{3.70}         & \multicolumn{1}{l|}{9.64}         & 21.46        & \multicolumn{1}{l|}{0.67}        & \multicolumn{1}{l|}{0.76}        & \multicolumn{1}{l|}{0.96}        & \multicolumn{1}{l|}{1.05}         & \multicolumn{1}{l|}{2.62}         & 9.24         \\ \hline
madelon             & \multicolumn{1}{l|}{26.72}       & \multicolumn{1}{l|}{30.06}       & \multicolumn{1}{l|}{34.78}       & \multicolumn{1}{l|}{36.28}        & \multicolumn{1}{l|}{40.36}        & 45.51        & \multicolumn{1}{l|}{22.09}       & \multicolumn{1}{l|}{23.19}       & \multicolumn{1}{l|}{25.93}       & \multicolumn{1}{l|}{27.67}        & \multicolumn{1}{l|}{33.81}        & 38.55        \\ \hline
mnist               & \multicolumn{1}{l|}{7.35}        & \multicolumn{1}{l|}{10.39}       & \multicolumn{1}{l|}{14.04}       & \multicolumn{1}{l|}{16.28}        & \multicolumn{1}{l|}{35.63}        & 58.30        & \multicolumn{1}{l|}{4.07}        & \multicolumn{1}{l|}{6.35}        & \multicolumn{1}{l|}{8.06}        & \multicolumn{1}{l|}{9.36}         & \multicolumn{1}{l|}{15.34}        & 43.57        \\ \hline
secom               & \multicolumn{1}{l|}{5.45}        & \multicolumn{1}{l|}{5.86}        & \multicolumn{1}{l|}{6.03}        & \multicolumn{1}{l|}{6.06}         & \multicolumn{1}{l|}{6.42}         & 18.04        & \multicolumn{1}{l|}{6.11}        & \multicolumn{1}{l|}{6.11}        & \multicolumn{1}{l|}{6.11}        & \multicolumn{1}{l|}{6.11}         & \multicolumn{1}{l|}{6.33}         & 13.02        \\ \hline
semeion             & \multicolumn{1}{l|}{10.89}       & \multicolumn{1}{l|}{14.22}       & \multicolumn{1}{l|}{15.34}       & \multicolumn{1}{l|}{17.57}        & \multicolumn{1}{l|}{28.64}        & 49.45        & \multicolumn{1}{l|}{5.38}        & \multicolumn{1}{l|}{6.12}        & \multicolumn{1}{l|}{8.74}        & \multicolumn{1}{l|}{9.78}         & \multicolumn{1}{l|}{15.62}        & 35.28        \\ \hline
shuttle             & \multicolumn{1}{l|}{0.03}        & \multicolumn{1}{l|}{0.22}        & \multicolumn{1}{l|}{0.55}        & \multicolumn{1}{l|}{1.19}         & \multicolumn{1}{l|}{5.26}         & 23.46        & \multicolumn{1}{l|}{0.02}        & \multicolumn{1}{l|}{0.03}        & \multicolumn{1}{l|}{0.04}        & \multicolumn{1}{l|}{0.05}         & \multicolumn{1}{l|}{0.63}         & 13.46        \\ \hline
waveform            & \multicolumn{1}{l|}{13.12}       & \multicolumn{1}{l|}{14.69}       & \multicolumn{1}{l|}{17.86}       & \multicolumn{1}{l|}{18.97}        & \multicolumn{1}{l|}{24.36}        & 37.52        & \multicolumn{1}{l|}{13.09}       & \multicolumn{1}{l|}{13.24}       & \multicolumn{1}{l|}{14.33}       & \multicolumn{1}{l|}{15.04}        & \multicolumn{1}{l|}{18.37}        & 32.94        \\ \hline
winequality         & \multicolumn{1}{l|}{37.62}       & \multicolumn{1}{l|}{42.46}       & \multicolumn{1}{l|}{45.08}       & \multicolumn{1}{l|}{45.75}        & \multicolumn{1}{l|}{49.34}        & 57.29        & \multicolumn{1}{l|}{36.57}       & \multicolumn{1}{l|}{38.22}       & \multicolumn{1}{l|}{40.47}       & \multicolumn{1}{l|}{41.33}        & \multicolumn{1}{l|}{43.49}        & 50.62        \\ \hline
yeast               & \multicolumn{1}{l|}{41.33}       & \multicolumn{1}{l|}{42.50}       & \multicolumn{1}{l|}{43.93}       & \multicolumn{1}{l|}{44.49}        & \multicolumn{1}{l|}{49.22}        & 59.95        & \multicolumn{1}{l|}{38.69}       & \multicolumn{1}{l|}{40.06}       & \multicolumn{1}{l|}{40.60}       & \multicolumn{1}{l|}{40.90}        & \multicolumn{1}{l|}{42.63}        & 52.10        \\ \hline
\end{tabular}

\end{table*}


Now, correlations between rankings provide only limited insight; it is possible to analyse the expected performances of reduction strategies directly.
First, the ongoing assumption of oracular perfection means that, for any reduced search-space size of $k=n$, an oracle strategy is expected to pool together the $k$ best predictors for each dataset.
Each predictor has a mean error rate, catalogued within meta-knowledge base $X$, so, assuming ignorance on how black-box SMAC prioritises predictors within a pool, one could assume that the average performance overall will be the mean of these $k$ mean performances.
Accordingly, with prefix \textit{e} emphasising the expected nature of these results, Table~\ref{tab:tab_expected_oracle} lists the average loss to anticipate from an oracle strategy, \textit{eOX-kn}, for each dataset.
As usual, $X=1$ denotes \textit{automl-meta} and $X=2$ denotes \textit{default-meta}.

Unsurprisingly, values for average loss in Table~\ref{tab:tab_expected_oracle} increase monotonically with $k$.
The \textit{eOX-k1} strategies should always select the best predictor evaluated for a dataset, with an expanding pool size successively including worse ML algorithms.
When all predictors -- this includes unevaluated algorithms awarded a loss of $100\%$ -- are in the search space, i.e.~for \textit{eOX-k30}, this marks the worst average attainable by an oracle strategy.
As a side note, closer inspection of the table does reaffirm earlier observations in Section~\ref{sec:chap4_dataset_understand} concerning the challenge of datasets.
Many `easy' datasets associated with numerous strongly performing predictors display sharp jumps in expected average loss when expanding from $k=19$ to $k=30$, e.g.~\textit{dorothea}, \textit{kddcup}, \textit{secom}, \textit{shuttle}, etc.
In contrast, the decrease in performance is much more steady across $k$ values for `hard' ML problems, e.g.~the four top-heavy datasets of \textit{convex}, \textit{amazon}, \textit{cifar10small}, and \textit{madelon}.

Progressing onwards, if oracle strategy \textit{eOX-kn} is the `smartest' choice, leveraging perfect knowledge of predictor performance on a dataset, then it is also worth establishing another baseline for the `dumbest' choice.
Specifically, randomly selecting a pool of $k$ predictors can be considered the most uninformed strategy.
Indeed, sometimes the best $k$ predictors will be selected, resulting in an optimal mean of mean performances.
Sometimes the worst $k$ will be selected, producing an abysmal mean of mean performances.
Thus, the expected average loss of what we term \textit{RX-kn} must involve an additional averaging over every combination of predictors possible.
For the record, the number of possibilities we average over ranges from ${30\choose 1} = 30$ to ${30\choose 19} = 54627300$.
In any case, it is possible to show both theoretically and numerically that the expected average loss of \textit{RX-kn} for any $k=n$ is equivalent to \textit{eOX-k30}, which is displayed in Table~\ref{tab:tab_expected_oracle}.

With bounds marking both perfect information and a complete lack thereof, one can situate the landmarked and global-leaderboard strategies within this spectrum, at least in terms of expectation.
Specifically, for any metric of interest $V(x)$ that can be extracted from strategy $x$, this metric can be normalised with respect to the oracle and random selection via
\begin{equation}
\label{eq:relative}
    \frac{V(\operatorname{S})-V(\operatorname{eOX-kn})}{V(\operatorname{RX-kn})-V(\operatorname{eOX-kn})},
\end{equation}
where strategy $S$ can be either \textit{eLX-kn} or \textit{eMX-kn}.

\begin{figure*}[!ht]
  \centering 
        \includegraphics[width=0.99\linewidth]{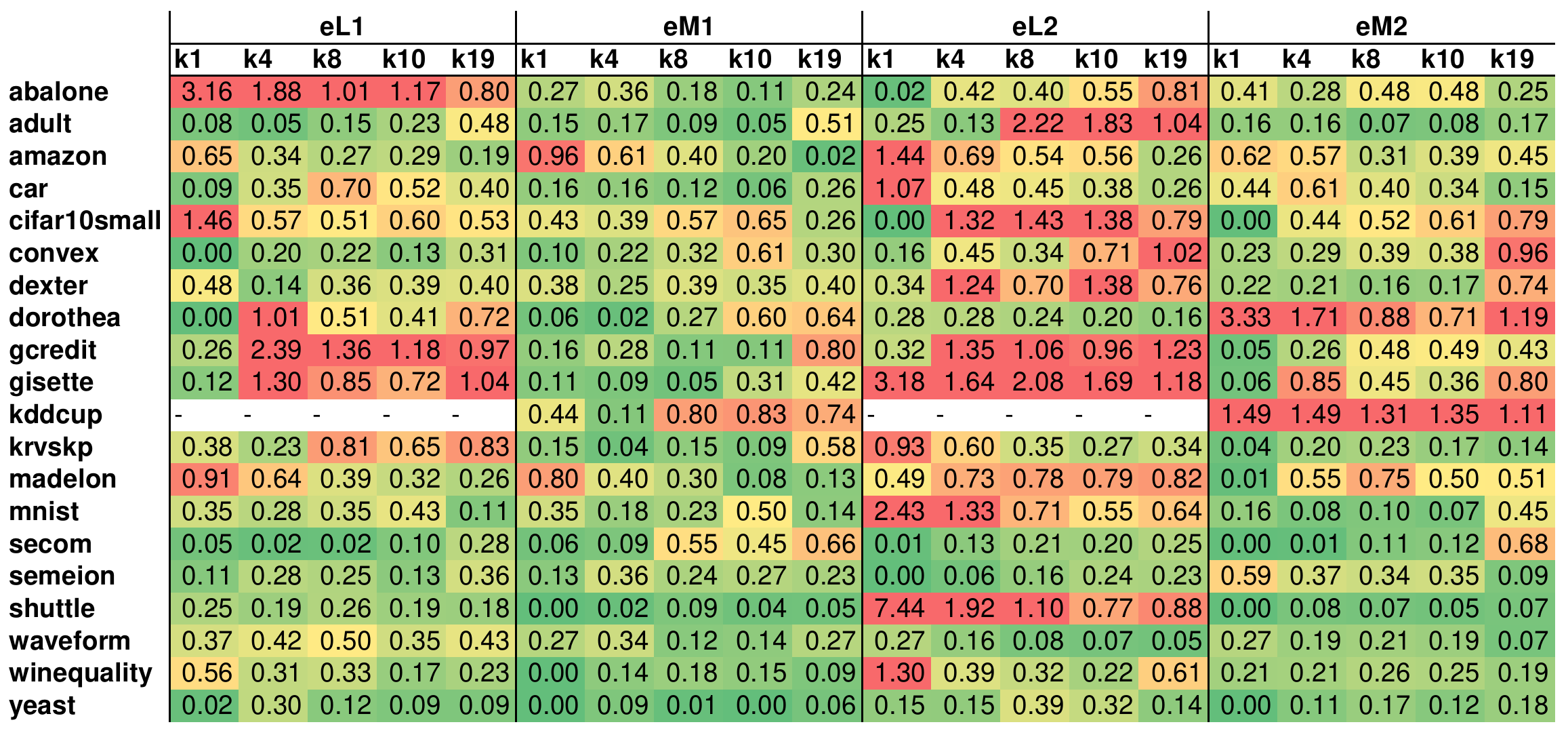}
      
    \caption{Normalised expected average loss for landmarked and global-leaderboard strategies, \textit{eLX-kn} and \textit{eMX-kn}, respectively, applied to each dataset. Prior to normalisation, each value is the average of the mean error rates for the strategy-selected best $k=n$ predictors specific to a dataset, as determined from a meta-knowledge base, where $X=1$ denotes \textit{automl-meta} and $X=2$ denotes \textit{default-meta}. Loss is then normalised linearly, where, for any $X$ and $k=n$, $0$ is the average loss for the ideal oracle strategy \textit{eOX-kn} and $1$ is the average loss for the uninformed random strategy \textit{RX-kn}. Notably, the average loss values for \textit{RX-kn} and \textit{eOX-k30} are identical; see text and Table~\ref{tab:tab_expected_oracle}.}
    \label{fig:normalised_distance_oracle_to_mean_eLM}
\end{figure*}

Currently, $V(x)$ represents the expected average of mean error rates for $k=n$ strategy-selected predictors, which handily reduces the denominator of Eq.~(\ref{eq:relative}) to $V(\operatorname{eOX-k30})-V(\operatorname{eOX-kn})$, and the normalisation of these expected averages are displayed in Fig.~\ref{fig:normalised_distance_oracle_to_mean_eLM}.
If a landmarked or global-leaderboard strategy scores close to $0$, then it is expected to perform substantially better than random selection, acting as a suitable replacement for the oracle strategy in a real-world ML application.
Alternatively, if they score close to $1$, meta-knowledge provides little competitive advantage over randomly pooling together a handful of $k$ predictors.
Additionally, given that there is no upper bound on the normalised value, \textit{eLX-kn} or \textit{eMX-kn} may cull configuration space so poorly that random search-space design would even be preferable; this is expected to occur for values greater than $1$.
Indeed, some of the more extreme $k=1$ values, e.g.~$7.44$ for \textit{shuttle}, arise due to the only predictor in the reduced pool of recommendations being one that has no successful evaluation, i.e.~an `as-bad-as-it-gets' loss of $100\%$.
In any case, it is essential to recall that the normalised scale shrinks for increasing values of $k$.
Given a fixed dataset and meta-knowledge base, the average loss of \textit{RX-kn} is static for any $k=n$, while the expected error rate for \textit{eOX-kn} moves towards it as $k$ grows; see Table~\ref{tab:tab_expected_oracle}.

Closer examination of the normalised average performances in Fig.~\ref{fig:normalised_distance_oracle_to_mean_eLM}, according to expectation, motivates several conclusions.
First of all, \textit{eM1} is the only pairing of strategy type and meta-knowledge base for which all normalised average loss values are below $1$.
This outcome suggests that opportunistic rankings averaged over all datasets are always expected to be usefully informative in culling AutoML search spaces.
Excluding the \textit{dorothea} and \textit{kddcup} datasets, which were complicated by run-to-completion issues, a global leaderboard built from default-hyperparameter evaluations, driving \textit{eM2}, is fairly robust as well.
In contrast, the shading in the figure indicates, overall, that opportunism-based landmarking lacks relative effectiveness, with \textit{eL1} expected to perform worse than random culling in some cases for \textit{abalone}, \textit{dorothea}, \textit{gcredit}, and \textit{gisette}.
As for landmarked strategies based on default hyperparameters, i.e.~\textit{eL2}, these results suggest they should rarely ever be recommended.
Again, it is possible that meta-predictors, while mostly ignored by \textit{automl-meta}, are better sampled in \textit{default-meta}, and the resulting mixed impact on rankings may weaken the correlations of such rankings between similar datasets, making landmarked recommendations less useful.

Nonetheless, it is risky to over-generalise based on Fig.~\ref{fig:normalised_distance_oracle_to_mean_eLM}.
For \textit{automl-meta}, landmarked strategies are expected to do better than the global leaderboard for \textit{secom} and, except for $k=19$, \textit{convex}.
The same is true in \textit{default-meta} for \textit{dorothea}, \textit{waveform}, and, except for $k=19$, \textit{semeion}.
Trends also vary from dataset to dataset.
For instance, for `hard' datasets \textit{amazon} and \textit{madelon}, an increase in $k$ provides a better chance of including their specialist best-performing ML algorithms within an AutoML search space.
Thus, leveraging \textit{automl-meta}, expected performances improve almost monotonically after \textit{eL1-k1} and \textit{eM1-k1} miss the mark.
Then, in contrast, there are `easy' datasets like \textit{shuttle} and \textit{yeast} that appear to have a small set of inferior performers at risk of popping up in randomly culled search spaces.
The slightest biasing based on a global leaderboard, from $k=1$ to $k=19$, is enough to dodge these bad choices; anything does well enough amongst the predictors that are left.
Nonetheless, despite varying trends, a weak generalisation can be made, based on figure shading, that the most extreme values for expected average loss, i.e.~those closest to and furthest from $0$, tend to occur for low $k$ values.
Essentially, if a reduction strategy hits or misses the mark with a small search space, the inclusion of more predictors will nullify these advantages/disadvantages; it is statistically unlikely for the sequence of the global leaderboard or landmarked rankings to identically match or entirely dispute the oracular rankings.

      


\begin{table*}[htb!]
\centering
\caption {Expected `optimal' loss for random culling strategies, \textit{RX-kn}, applied to each dataset. Random culling can suggest $30\choose n$ unique search spaces, each with $k=n$ predictors. Per search space, an `optimal' loss is the minimum of the dataset-specific mean error rates for the predictors involved, as determined from a meta-knowledge base, where $X=1$ denotes \textit{automl-meta} and $X=2$ denotes \textit{default-meta}. The `optimal' loss is then averaged across all $30\choose n$ combinations.}
\label{tab:tab_expected_random_strategy}
\fontsize{7.2}{8.5}\selectfont

\begin{tabular}{|l|llllll|llllll|}
\hline
\textbf{Strategies} & \multicolumn{6}{c|}{\textbf{R1}}                                                                                                                                                              & \multicolumn{6}{c|}{\textbf{R2}}                                                                                                                                                              \\ \hline
\textbf{k values}   & \multicolumn{1}{l|}{\textbf{k1}} & \multicolumn{1}{l|}{\textbf{k4}} & \multicolumn{1}{l|}{\textbf{k8}} & \multicolumn{1}{l|}{\textbf{k10}} & \multicolumn{1}{l|}{\textbf{k19}} & \textbf{k30} & \multicolumn{1}{l|}{\textbf{k1}} & \multicolumn{1}{l|}{\textbf{k4}} & \multicolumn{1}{l|}{\textbf{k8}} & \multicolumn{1}{l|}{\textbf{k10}} & \multicolumn{1}{l|}{\textbf{k19}} & \textbf{k30} \\ \hline
abalone             & \multicolumn{1}{l|}{82.18}       & \multicolumn{1}{l|}{75.32}       & \multicolumn{1}{l|}{74.60}       & \multicolumn{1}{l|}{74.44}        & \multicolumn{1}{l|}{74.06}        & 73.92        & \multicolumn{1}{l|}{78.62}       & \multicolumn{1}{l|}{74.29}       & \multicolumn{1}{l|}{73.58}       & \multicolumn{1}{l|}{73.43}        & \multicolumn{1}{l|}{73.16}        & 73.08        \\ \hline
adult               & \multicolumn{1}{l|}{28.11}       & \multicolumn{1}{l|}{15.77}       & \multicolumn{1}{l|}{14.62}       & \multicolumn{1}{l|}{14.42}        & \multicolumn{1}{l|}{14.04}        & 13.86        & \multicolumn{1}{l|}{19.72}       & \multicolumn{1}{l|}{14.57}       & \multicolumn{1}{l|}{14.22}       & \multicolumn{1}{l|}{14.13}        & \multicolumn{1}{l|}{13.90}        & 13.77        \\ \hline
amazon              & \multicolumn{1}{l|}{84.78}       & \multicolumn{1}{l|}{62.12}       & \multicolumn{1}{l|}{50.41}       & \multicolumn{1}{l|}{47.12}        & \multicolumn{1}{l|}{39.15}        & 36.19        & \multicolumn{1}{l|}{78.36}       & \multicolumn{1}{l|}{51.47}       & \multicolumn{1}{l|}{39.99}       & \multicolumn{1}{l|}{37.01}        & \multicolumn{1}{l|}{30.79}        & 29.14        \\ \hline
car                 & \multicolumn{1}{l|}{32.30}       & \multicolumn{1}{l|}{13.44}       & \multicolumn{1}{l|}{10.11}       & \multicolumn{1}{l|}{9.42}         & \multicolumn{1}{l|}{7.99}         & 7.06         & \multicolumn{1}{l|}{20.73}       & \multicolumn{1}{l|}{7.19}        & \multicolumn{1}{l|}{4.95}        & \multicolumn{1}{l|}{4.33}         & \multicolumn{1}{l|}{2.62}         & 1.24         \\ \hline
cifar10small        & \multicolumn{1}{l|}{85.58}       & \multicolumn{1}{l|}{70.73}       & \multicolumn{1}{l|}{64.94}       & \multicolumn{1}{l|}{63.22}        & \multicolumn{1}{l|}{57.98}        & 54.20        & \multicolumn{1}{l|}{85.77}       & \multicolumn{1}{l|}{72.10}       & \multicolumn{1}{l|}{68.64}       & \multicolumn{1}{l|}{67.94}        & \multicolumn{1}{l|}{66.55}        & 66.14        \\ \hline
convex              & \multicolumn{1}{l|}{54.92}       & \multicolumn{1}{l|}{37.79}       & \multicolumn{1}{l|}{33.62}       & \multicolumn{1}{l|}{32.40}        & \multicolumn{1}{l|}{29.51}        & 28.18        & \multicolumn{1}{l|}{46.69}       & \multicolumn{1}{l|}{34.53}       & \multicolumn{1}{l|}{29.94}       & \multicolumn{1}{l|}{28.72}        & \multicolumn{1}{l|}{25.94}        & 24.33        \\ \hline
dexter              & \multicolumn{1}{l|}{51.05}       & \multicolumn{1}{l|}{17.54}       & \multicolumn{1}{l|}{11.45}       & \multicolumn{1}{l|}{10.37}        & \multicolumn{1}{l|}{8.57}         & 7.91         & \multicolumn{1}{l|}{31.10}       & \multicolumn{1}{l|}{10.11}       & \multicolumn{1}{l|}{8.16}        & \multicolumn{1}{l|}{7.82}         & \multicolumn{1}{l|}{7.15}         & 6.67         \\ \hline
dorothea            & \multicolumn{1}{l|}{55.17}       & \multicolumn{1}{l|}{13.01}       & \multicolumn{1}{l|}{7.49}        & \multicolumn{1}{l|}{7.21}         & \multicolumn{1}{l|}{6.91}         & 6.78         & \multicolumn{1}{l|}{34.73}       & \multicolumn{1}{l|}{7.87}        & \multicolumn{1}{l|}{7.04}        & \multicolumn{1}{l|}{6.96}         & \multicolumn{1}{l|}{6.79}         & 6.71         \\ \hline
gcredit             & \multicolumn{1}{l|}{32.83}       & \multicolumn{1}{l|}{26.25}       & \multicolumn{1}{l|}{25.14}       & \multicolumn{1}{l|}{24.86}        & \multicolumn{1}{l|}{24.19}        & 23.77        & \multicolumn{1}{l|}{27.30}       & \multicolumn{1}{l|}{24.81}       & \multicolumn{1}{l|}{24.35}       & \multicolumn{1}{l|}{24.27}        & \multicolumn{1}{l|}{24.15}        & 24.14        \\ \hline
gisette             & \multicolumn{1}{l|}{41.24}       & \multicolumn{1}{l|}{7.81}        & \multicolumn{1}{l|}{4.61}        & \multicolumn{1}{l|}{4.07}         & \multicolumn{1}{l|}{2.94}         & 2.61         & \multicolumn{1}{l|}{33.50}       & \multicolumn{1}{l|}{5.68}        & \multicolumn{1}{l|}{4.06}        & \multicolumn{1}{l|}{3.77}         & \multicolumn{1}{l|}{3.14}         & 2.96         \\ \hline
kddcup              & \multicolumn{1}{l|}{51.74}       & \multicolumn{1}{l|}{7.13}        & \multicolumn{1}{l|}{1.92}        & \multicolumn{1}{l|}{1.78}         & \multicolumn{1}{l|}{1.73}         & 1.70         & \multicolumn{1}{l|}{67.50}       & \multicolumn{1}{l|}{19.47}       & \multicolumn{1}{l|}{4.02}        & \multicolumn{1}{l|}{2.45}         & \multicolumn{1}{l|}{1.80}         & 1.80         \\ \hline
krvskp              & \multicolumn{1}{l|}{21.46}       & \multicolumn{1}{l|}{5.04}        & \multicolumn{1}{l|}{2.61}        & \multicolumn{1}{l|}{2.23}         & \multicolumn{1}{l|}{1.51}         & 1.29         & \multicolumn{1}{l|}{9.24}        & \multicolumn{1}{l|}{1.55}        & \multicolumn{1}{l|}{0.93}        & \multicolumn{1}{l|}{0.84}         & \multicolumn{1}{l|}{0.72}         & 0.67         \\ \hline
madelon             & \multicolumn{1}{l|}{45.51}       & \multicolumn{1}{l|}{36.11}       & \multicolumn{1}{l|}{31.92}       & \multicolumn{1}{l|}{30.62}        & \multicolumn{1}{l|}{27.57}        & 26.72        & \multicolumn{1}{l|}{38.55}       & \multicolumn{1}{l|}{28.60}       & \multicolumn{1}{l|}{24.71}       & \multicolumn{1}{l|}{23.86}        & \multicolumn{1}{l|}{22.42}        & 22.09        \\ \hline
mnist               & \multicolumn{1}{l|}{58.30}       & \multicolumn{1}{l|}{20.28}       & \multicolumn{1}{l|}{12.18}       & \multicolumn{1}{l|}{10.78}        & \multicolumn{1}{l|}{8.04}         & 7.35         & \multicolumn{1}{l|}{43.57}       & \multicolumn{1}{l|}{10.85}       & \multicolumn{1}{l|}{7.07}        & \multicolumn{1}{l|}{6.36}         & \multicolumn{1}{l|}{4.73}         & 4.07         \\ \hline
secom               & \multicolumn{1}{l|}{18.04}       & \multicolumn{1}{l|}{6.14}        & \multicolumn{1}{l|}{5.93}        & \multicolumn{1}{l|}{5.88}         & \multicolumn{1}{l|}{5.67}         & 5.45         & \multicolumn{1}{l|}{13.02}       & \multicolumn{1}{l|}{6.19}        & \multicolumn{1}{l|}{6.11}        & \multicolumn{1}{l|}{6.11}         & \multicolumn{1}{l|}{6.11}         & 6.11         \\ \hline
semeion             & \multicolumn{1}{l|}{49.45}       & \multicolumn{1}{l|}{20.22}       & \multicolumn{1}{l|}{15.06}       & \multicolumn{1}{l|}{14.22}        & \multicolumn{1}{l|}{12.48}        & 10.89        & \multicolumn{1}{l|}{35.28}       & \multicolumn{1}{l|}{10.92}       & \multicolumn{1}{l|}{7.80}        & \multicolumn{1}{l|}{7.18}         & \multicolumn{1}{l|}{5.90}         & 5.38         \\ \hline
shuttle             & \multicolumn{1}{l|}{23.46}       & \multicolumn{1}{l|}{2.19}        & \multicolumn{1}{l|}{0.55}        & \multicolumn{1}{l|}{0.35}         & \multicolumn{1}{l|}{0.11}         & 0.03         & \multicolumn{1}{l|}{13.46}       & \multicolumn{1}{l|}{0.22}        & \multicolumn{1}{l|}{0.04}        & \multicolumn{1}{l|}{0.03}         & \multicolumn{1}{l|}{0.02}         & 0.02         \\ \hline
waveform            & \multicolumn{1}{l|}{37.52}       & \multicolumn{1}{l|}{19.70}       & \multicolumn{1}{l|}{16.19}       & \multicolumn{1}{l|}{15.35}        & \multicolumn{1}{l|}{13.65}        & 13.12        & \multicolumn{1}{l|}{32.94}       & \multicolumn{1}{l|}{15.64}       & \multicolumn{1}{l|}{13.85}       & \multicolumn{1}{l|}{13.54}        & \multicolumn{1}{l|}{13.13}        & 13.09        \\ \hline
winequality         & \multicolumn{1}{l|}{57.29}       & \multicolumn{1}{l|}{45.97}       & \multicolumn{1}{l|}{43.00}       & \multicolumn{1}{l|}{42.04}        & \multicolumn{1}{l|}{39.10}        & 37.62        & \multicolumn{1}{l|}{50.62}       & \multicolumn{1}{l|}{41.35}       & \multicolumn{1}{l|}{39.23}       & \multicolumn{1}{l|}{38.62}        & \multicolumn{1}{l|}{37.25}        & 36.57        \\ \hline
yeast               & \multicolumn{1}{l|}{59.95}       & \multicolumn{1}{l|}{45.42}       & \multicolumn{1}{l|}{43.10}       & \multicolumn{1}{l|}{42.63}        & \multicolumn{1}{l|}{41.63}        & 41.33        & \multicolumn{1}{l|}{52.10}       & \multicolumn{1}{l|}{41.28}       & \multicolumn{1}{l|}{40.24}       & \multicolumn{1}{l|}{40.01}        & \multicolumn{1}{l|}{39.34}        & 38.69        \\ \hline
\end{tabular}

\end{table*}


Crucially, optimisation ultimately seeks an optimum, by definition.
So, while the expected average loss of a strategy is illustrative of an optimisation \textit{journey}, an expected `optimal' loss of a strategy is more indicative of the optimisation \textit{destination}.
The two need not be correlated.
Therefore, we also examine the mean performances of the best ML algorithms within strategically reduced search spaces, i.e.~the minimum of the mean error rates for the best $k$ predictors according to various strategies.
We note that this definition of `optimal' does not refer to the performance of the absolute best ML pipeline per predictor, only the best mean performance of a predictor, hence the quotation marks.

Now, for an oracle strategy that returns a search space of any size, given a fixed dataset and meta-knowledge base, the `optimal' loss will always be the same; the `smartest' strategy should always include the best predictor for a dataset in its search space.
In contrast to before, it is now the `dumbest' strategy of random culling that varies in expectation across different values of $k$.
Sure enough, Table~\ref{tab:tab_expected_random_strategy} lists the `optimal' loss to expect when randomly culling a search space to size $k=n$ for any dataset and meta-knowledge base.
Here, we re-emphasise that random culling can produce $30\choose n$ possible search spaces, so the expected `optimal' loss is technically an average across all possibilities.
Thus, unsurprisingly, the `optimal' loss decreases for increasing values of $k$, as larger search spaces have a greater chance of including the best predictors.
At one extreme, for \textit{RX-k1}, there is only one predictor within each of its $30\choose 1$ possible search spaces, so the minimum and average of mean error rates per search space are trivially identical.
Subsequently averaging across these 30 search spaces is then equivalent to averaging the mean performance of all 30 predictors, so the expected `optimal' loss of \textit{RX-k1} should be equal to the expected average loss of \textit{eOX-k30}.
At the other extreme, \textit{RX-k30} is guaranteed to have the best predictor within its search space, so its minimum of mean error rates should be equivalent to the average of the lowest mean error rate, i.e.~the expected average loss of \textit{eOX-k1}.
Both forms of alignment are indeed confirmed when inspecting Table~\ref{tab:tab_expected_oracle} and Table~\ref{tab:tab_expected_random_strategy}.

\begin{figure*}[!ht]
  \centering 
        \includegraphics[width=0.99\linewidth]{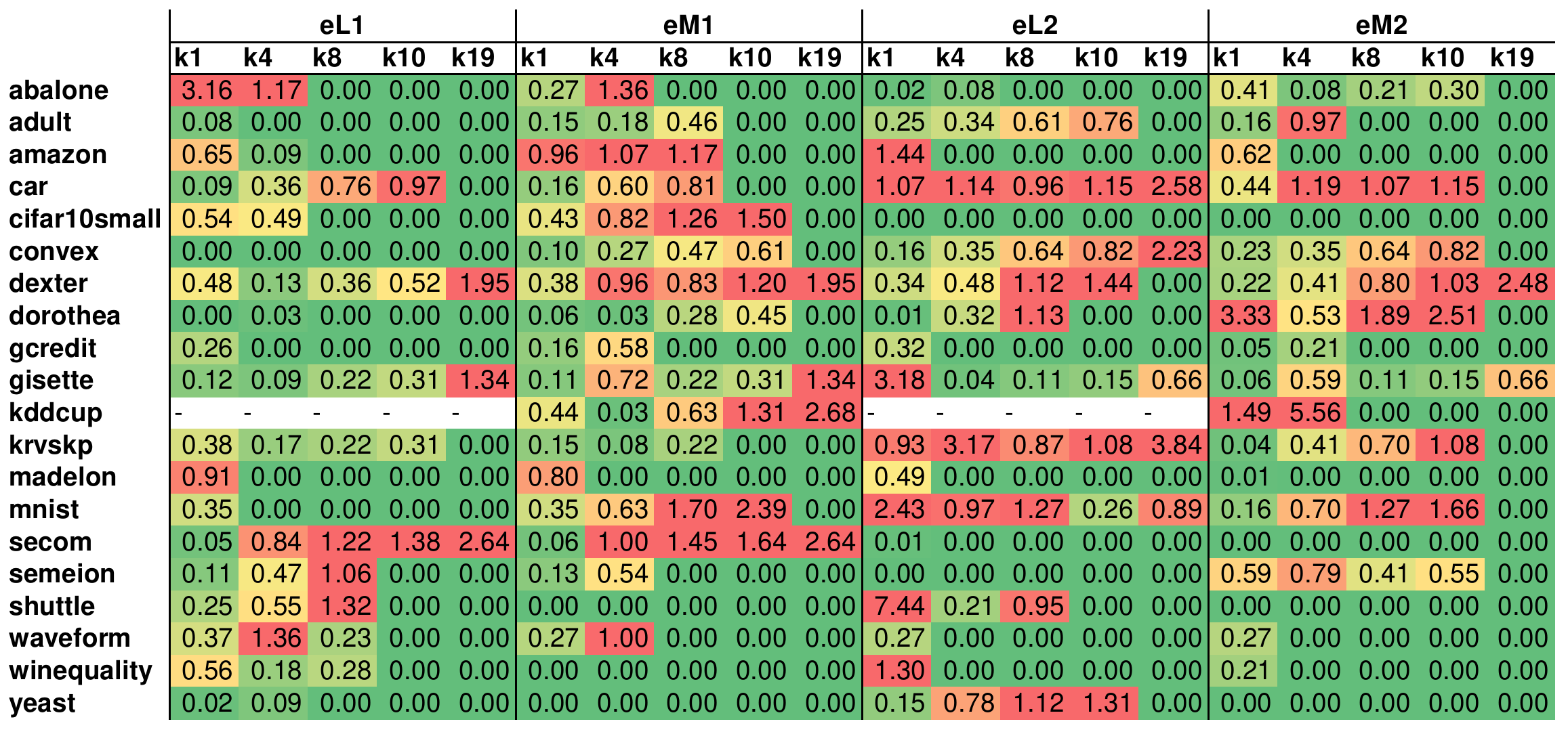}
    \caption{Normalised expected `optimal' loss for landmarked and global-leaderboard strategies, \textit{eLX-kn} and \textit{eMX-kn}, respectively, applied to each dataset. Prior to normalisation, each value is the minimum of the mean error rates for the strategy-selected best $k=n$ predictors specific to a dataset, as determined from a meta-knowledge base, where $X=1$ denotes \textit{automl-meta} and $X=2$ denotes \textit{default-meta}. Loss is then normalised linearly, where, for any $X$ and $k=n$, $0$ is the `optimal' loss for the ideal oracle strategy \textit{eOX-kn} and $1$ is the `optimal' loss for the uninformed random strategy \textit{RX-kn}. Notably, the `optimal' loss values for \textit{eOX-kn} and \textit{RX-k30} are identical; see text and Table~\ref{tab:tab_expected_random_strategy}.}
    \label{fig:normalised_distance_oracle_to_min_eLM}
\end{figure*}

Given that the best predictor for \textit{RX-k30} is identical to the best predictor for any oracle strategy, i.e.~\textit{eOX-kn}, we can now perform a normalisation of `optimal' performance for every reduction strategy.
Specifically, if metric $V(x)$ is now the expected minimum of mean error rates, the denominator of Eq.~(\ref{eq:relative}), i.e.~the normalisation range, can be rewritten as $V(\operatorname{RX-kn})-V(\operatorname{RX-k30})$.
Accordingly, the mean performance of the best predictor for every landmarked and global-leaderboard strategy is situated within this spectrum in Fig.~\ref{fig:normalised_distance_oracle_to_min_eLM}.
A value of $0$ means that the reduction method, like an oracle strategy, is expected to include the best predictor for a dataset, as asserted by a meta-knowledge base, within its culled search space.
In contrast, a value of $1$ suggests that, on average, there is no difference between a strategic reduction and a random reduction of a search space to size $k$, at least in terms of how good the best predictor in that space should be.
Furthermore, as before, non-zero values greater or less than $1$ indicate the extent to which random culling or an informed strategy is preferable over the other.

Overall, analysis of the expected `optimal' loss in Fig.~\ref{fig:normalised_distance_oracle_to_min_eLM} serves a different purpose to the expected average loss in Fig.~\ref{fig:normalised_distance_oracle_to_mean_eLM}.
For increasing $k$, every strategy will sooner or later include the best ML algorithm for a dataset, hitting a normalised value of $0$.
The question is how quickly they can do so.
This consideration is especially important as the normalisation scale is gradually shortening, with \textit{RX-kn} approaching \textit{eOX-kn} for increasing $k$.
So, if a search-space expansion does not find a better predictor than the current best, the normalised `optimal' loss increases in value, reflecting a worsening performance.
This effect can colour the narrative for how varying reduction approaches perform on different datasets.
For instance, \textit{eM1} strategies do not identify the best predictor for both \textit{cifar10small} and \textit{dorothea} until $k=19$, but, for the latter, there is at least some improvement on the initial guess when expanding from $k=1$ to $k=4$, i.e.~the normalised `optimal' loss decreases from $0.06$ to $0.03$.

Unfortunately, it is difficult to draw overarching conclusions from Fig.~\ref{fig:normalised_distance_oracle_to_min_eLM}.
One might hope to find distinctions based on dataset challenge, given that informed strategies should struggle to recommend the specialist ML algorithms capable of conquering top-heavy `hard' ML problems.
Yet only \textit{convex} proves somewhat problematic, with \textit{madelon} even demonstrating the effectiveness of \textit{any} informed strategy at locating its best predictor for $k\ge 4$.
Broadening the definition of challenge, in accordance with Table~\ref{tab:top_predictors_default}, \textit{gisette} is considered a `hard' problem and, sure enough, informed strategies fail to include its optimal predictor within an AutoML search space even for $k=19$, regardless of meta-knowledge base.
However, meta-knowledge can also frequently fail to recommend the optimum for `easier' datasets, such as \textit{dexter} and, for \textit{automl-meta}, \textit{secom}.

Nevertheless, despite the variety in performance trends, both Fig.~\ref{fig:normalised_distance_oracle_to_mean_eLM} and Fig.~\ref{fig:normalised_distance_oracle_to_min_eLM} combine to provide a useful perspective on search-space reduction strategies across different datasets.
Indeed, while the normalised `optimum' and average of mean predictor performances should be identical for $k=1$, they provide parallel threads of commentary on what to anticipate when $k>1$.
Sometimes, as is the case for \textit{eM1} and \textit{yeast}, informed strategies are expected to consistently dominate random culling, pooling together predictors that perform well, one of which is the best.
Other times, the recommended pool is broadly unimpressive, such as when \textit{eL2} is applied to \textit{mnist}, peaking at $k=10$ with a normalised `optimal' and average loss of 0.26 and 0.55, respectively.
This result may seem respectable, but every other combination of dataset, strategy type and meta-knowledge base can beat at least one of those values for some $k$; in contrast, \textit{eL2} can do no better on \textit{mnist}, often comparable to or worse than random culling.
However, beyond these two examples, there are also cases like \textit{eM1} applied to \textit{gisette}, where the best predictor eludes the global-leaderboard strategy, yet the average performance to expect from the associated pool of predictors is still much better than one could derive via random culling, i.e.~the normalised average loss is 0.42 or lower for $k\le 19$.
Alternatively, \textit{eL1} seems generally terrible for \textit{gcredit} when $k\ge 4$, with the normalised average loss never improving beyond 0.97, yet, counter-intuitively, an `optimal' loss of 0.00 indicates that the best predictor for \textit{gcredit} consistently features in the search space recommended by \textit{eL1}.

Ultimately, while illustrative, expected averages and `optima' do not account for how a black-box optimiser will explore a strategically recommended search space.
If the expected average is bad, but the expected `optimum' is good, will an optimiser get stuck exploring poor solutions, or will it successfully exploit a path towards high performance?
A priori, this is unclear.
Even if the same optimiser, i.e.~SMAC, is used to test the strategies that leverage the meta-knowledge it has compiled, i.e.~\textit{automl-meta}, stochastic variability is inevitable.
Of course, the ablative analysis that has informed expectation in this section has likewise disregarded stochastic variability in the compilation of meta-knowledge.
Thus, the final step of this research is to accept all these sources of statistical noise and examine the actual application of search-space reduction strategies, seeking to identify if any resulting performance trends are as expected.

\subsection{Evaluating Reduction Strategies: True Performance}
\label{sec:results}

By this point, we have comprehensively analysed two opportunistic/systematic meta-knowledge bases of ML pipeline evaluations for a diverse array of 20 classification problems.
As part of this, we also explored differences between datasets predominantly based on the notion of `challenge'.
A handful of ML problems were determined to be particularly `hard', where an intelligent selection of an ML algorithm could significantly impact accuracy.
Finally, leveraging predictor rankings derived from the meta-knowledge bases, we established expectations for various search-space reduction strategies, comparing them against well informed and uninformed approaches, i.e.~oracle-based strategies and random culling, respectively.
Here, we examine the actual outcomes of each strategy in Section~\ref{Sec:Strategies} being run.

Crucially, each of the 33 search-space reduction strategies in this work is applied five times per dataset, with each of the five runs allowed to take two hours on the hardware specified in Section~\ref{sec:chap4_settings}.
Per run, AutoWeka4MCPS returns the best 10-fold CV evaluation and its associated ML pipeline.
Thus, it is possible to assess the consistency of each strategy by examining, again with Welch's t-test, how many pairs of the five 10-fold evaluations denote statistical indistinguishability.
Given that, for $n=5$ solutions, there are $n(n-1)/2=10$ pairings, this consistency metric ends up quadratic, i.e.~\textit{nonlinear}.
Specifically, if only two out of $n=5$ solutions are indistinguishable, then this is equivalent to one pairing out of $n(n-1)/2=10$, i.e.~a consistency of $0.1$.
Three indistinguishable solutions possess three pairings, i.e.~a consistency of $0.3$, and four indistinguishable solutions have six associated pairings, i.e.~a consistency of $0.6$.
Of course, consistency values of $1$ and $0$ represent all solutions performing similarly and dissimilarly, respectively.
We emphasise, however, that consistency simply measures performance similarity; a strategy need not return the same ML pipeline with the same components each time.

Notably, not all runs complete successfully.
This eventuality could be due to a catastrophic failure based on hardware or simply because SMAC fails to grant a single ML pipeline a full 10-fold CV evaluation by the end of two hours.
Thus, beyond consistency, it is also possible to count the number of times, from $0$ to $5$, that SMAC assisted by a particular reduction strategy fails to complete.
As part of this definition, we consider any completion failure statistically distinct from any other run, which places a practical ceiling on the consistency metric.
For instance, if a strategy fails to complete three times, i.e.~$3$ failures, this leaves only a single pair of ML pipelines to possibly be statistically indistinguishable, thus capping consistency at $0.1$.

\begin{table*}[htb!]


\caption {Performance consistency (0 to 1) and number of completion failures (0 to 5) averaged across all search-space reduction strategies applied to each dataset. See text for metric definitions.
}
\label{tab:tab_solution_consistency_crashes_metrics}

\centering
\fontsize{6.0}{8.5}\selectfont
\setlength\tabcolsep{3.0pt}
\begin{tabular}{|l|l|l|l|l|l|l|l|l|l|l|l|l|l|l|l|l|l|l|l|l|}
\hline
  & \rotatebox{90}{\textbf{abalone}} & \rotatebox{90}{\textbf{adult}} & \rotatebox{90}{\textbf{amazon}} & \rotatebox{90}{\textbf{car}} & \rotatebox{90}{\textbf{cifar10small}} & \rotatebox{90}{\textbf{convex}} & \rotatebox{90}{\textbf{dexter}}& \rotatebox{90}{\textbf{dorothea}} & \rotatebox{90}{\textbf{gcredit}} & \rotatebox{90}{\textbf{gisette}} & \rotatebox{90}{\textbf{kddcup}} & \rotatebox{90}{\textbf{krvskp}} & \rotatebox{90}{\textbf{madelon}} & \rotatebox{90}{\textbf{mnist}} & \rotatebox{90}{\textbf{secom}} & \rotatebox{90}{\textbf{semeion}} & \rotatebox{90}{\textbf{shuttle}} & \rotatebox{90}{\textbf{waveform}} & \rotatebox{90}{\textbf{winequality}} & \rotatebox{90}{\textbf{yeast}} \\ \hline
\textbf{Average consistency}  & 0.43             & 0.31           & 0.17            & 0.60         & 0.07                  & 0.09            & 0.16            & 0.04              & 0.33             & 0.10             & 0.17            & 0.52            & 0.52             & 0.01           & 0.76           & 0.23             & 0.44             & 0.38              & 0.50                 & 0.41           \\ \hline
\textbf{Average failures}  & 0.39             & 1.39           & 2.00            & 0.00         & 3.06                  & 1.55            & 1.15            & 3.39              & 0.00             & 2.09             & 2.85            & 0.00            & 0.33             & 3.27           & 0.76           & 0.06             & 0.39             & 0.12              & 0.06                 & 0.03           \\ \hline
\end{tabular}

\end{table*}

With definitions established for performance consistency and completion failures, Table~\ref{tab:tab_solution_consistency_crashes_metrics} displays average values, per dataset, across all 33 strategies.
By and large, this table correlates closely with the number of evaluations present for each dataset in \textit{automl-meta}, as displayed by Fig.~\ref{fig:f1_automl_number_of_evaluations_heatmap}.
For instance, any dataset that was computationally manageable enough to receive over 2500 evaluations in \textit{automl-meta}, averaging over 500 per two-hour run, likewise enables strategy-assisted SMAC to reach relatively consistent outcomes, such that at least three out of five returned ML pipelines appear statistically indistinguishable, i.e.~consistency is above $0.3$.
These computationally tractable datasets include \textit{car}, \textit{yeast}, \textit{gcredit}, \textit{krvskp}, \textit{winequality}, \textit{secom}, and \textit{waveform}.
In fact, \textit{car} and \textit{secom} both have a consistency value of $0.6$ or above, meaning that, on average, associated strategies return no more than one statistically distinct solution.
Given that this fifth solution may often be a crash for \textit{secom}, with an average of $0.76$ completion failures, \textit{secom} nonetheless supports remarkably consistent outcomes.

In contrast, every dataset that received under $750$ evaluations in \textit{automl-meta}, averaging under $150$ per two-hour run, manifests an average failure count of over $1$, i.e.~\textit{adult}, \textit{dexter}, \textit{convex}, \textit{kddcup}, \textit{gisette}, \textit{amazon}, \textit{cifar10small}, \textit{mnist}, and \textit{dorothea}.
With the marginal exception of \textit{adult}, strategy-assisted SMAC never approaches or surpasses a consistency of $0.3$ for these ML problems, seemingly lacking time to settle into an optimum where more than two returned results have similar performance.
Moreover, the most computationally intractable datasets in \textit{automl-meta}, receiving under $250$ evaluations in total and averaging under $50$ per run, cause substantial problems for reduction strategies, with an average failure count of over $3$ and an average consistency of under $0.1$.
In essence, AutoML generally fails to converge to a stable solution for \textit{cifar10small}, \textit{mnist}, and \textit{dorothea}.
This outcome does not mean that search-space reduction based on meta-learning is useless for these datasets; in contrast, a narrowed focus may be more important than ever for finding any ML solution of value in a time-critical computationally intractable setting.
However, this result highlights just how vital hardware and runtime are for both generating useful meta-knowledge and ensuring the stable convergence of strategy-assisted ML model search.

\begin{table}[htb!]
\centering
\caption {Performance consistency (0 to 1) and number of completion failures (0 to 5) averaged across all search-space reduction strategies that are of a certain type, use a certain meta-knowledge base, or reduce the search space to a certain size $k$. See text for metric definitions.}
\label{tab:tab_solution_consistency_metrics_summary}
\fontsize{7.2}{8.5}\selectfont
\begin{tabular}{|ll|l|l|}
\hline
\multicolumn{2}{|l|}{\textbf{}}                                                              & \textbf{Consistency} & \textbf{Failures} \\ \hline
\multicolumn{1}{|l|}{\multirow{6}{*}{\textbf{Strategies}}}     & \textbf{Landmarking}        & 0.29                 & 1.41             \\ \cline{2-4} 
\multicolumn{1}{|l|}{}                                         & \textbf{Global leaderboard} & 0.36                 & 0.80             \\ \cline{2-4} 
\multicolumn{1}{|l|}{}                                         & \textbf{Oracle}             & 0.32                 & 1.22             \\ \cline{2-4} 
\multicolumn{1}{|l|}{}                                         & \textbf{Avatar}             & 0.24                 & 1.10             \\ \cline{2-4} 
\multicolumn{1}{|l|}{}                                         & \textbf{Baseline}           & 0.11                 & 2.00             \\ \cline{2-4} 
\multicolumn{1}{|l|}{}                                         & \textbf{r30}                & 0.29                 & 0.55             \\ \hline
\multicolumn{1}{|l|}{\multirow{2}{*}{\textbf{Meta-knowledge}}} & \textbf{automl-meta}        & 0.33                 & 1.14             \\ \cline{2-4} 
\multicolumn{1}{|l|}{}                                         & \textbf{default-meta}       & 0.32                 & 1.14             \\ \hline
\multicolumn{1}{|l|}{\multirow{5}{*}{\textbf{k values}}}       & \textbf{k1}                 & 0.38                 & 0.90             \\ \cline{2-4} 
\multicolumn{1}{|l|}{}                                         & \textbf{k4}                 & 0.37                 & 1.13             \\ \cline{2-4} 
\multicolumn{1}{|l|}{}                                         & \textbf{k8}                 & 0.31                 & 1.23             \\ \cline{2-4} 
\multicolumn{1}{|l|}{}                                         & \textbf{k10}                & 0.31                 & 1.17             \\ \cline{2-4} 
\multicolumn{1}{|l|}{}                                         & \textbf{k19}                & 0.24                 & 1.27             \\ \hline
\end{tabular}

\end{table}

In continuing with an analysis of performance consistency and completion failures, it is possible to average results across other groupings of strategies.
For instance, averaging can be done by strategy type, meta-knowledge base, or the size $k$ of the reduced search space.
These results are displayed in Table~\ref{tab:tab_solution_consistency_metrics_summary}, from which it is immediately clear that consistency and completion failures varied much more dramatically across datasets.
Nonetheless, certain conclusions can be derived.
Obviously, the \textit{baseline} strategy has the highest failure rate and the lowest consistency rating, wasting time on evaluating incompatible ML pipelines.
The AVATAR system is responsible for normalising the failure rate of the \textit{avatar} strategy, even though its consistency value remains second-lowest in the table; SMAC is still faced with exploring a configuration pool of 30 predictors.
Sure enough, the stresses of an expansive search space on an AutoML run are evident across the board.
Performance consistency and the number of successful completions generally improve as $k$ decreases.

Other noteworthy results include \textit{r30} having the lowest failure rate.
This outcome is unsurprising, as the continuation strategy is predicated on a fixed pipeline structure; this composition has already been proven to run for some configuration.
Additionally, hyperparameter space is much more constrained, thus allowing SMAC to produce a 10-fold CV evaluation quickly.
Beyond \textit{r30}, however, there is also a monotonic increase and decrease of consistency and failures, respectively, to take note of, specifically when moving from \textit{LX-kn} through \textit{OX-kn} to \textit{MX-kn}.
This result makes partial sense, given that landmarking was identified in Section~\ref{Sec:EvalOracle} as sub-optimal for the 20 datasets used in this work; there are likely many cases where poor recommendations are provided.
However, it is less clear why the global leaderboard bests the oracle.
One can speculate that the reasoning comes down to matters of stability.
The global leaderboard prioritises ML algorithms that do generally well, robustly determined across all 20 datasets, while the oracle highlights predictors that do exceptionally well for a given ML problem.
The problem here is that stochastic variability for datasets with a small number of evaluations in \textit{automl-meta} and multiple crashes in \textit{default-meta} is likely to distort oracular rankings much more than the global leaderboard.
As a side note, there appears to be no differentiation between \textit{automl-meta} and \textit{default-meta} in terms of promoting stable convergence, which is a notable outcome for two meta-knowledge bases constructed so differently.

\begin{figure*}[!ht]
  \centering 
        \includegraphics[width=0.99\linewidth]{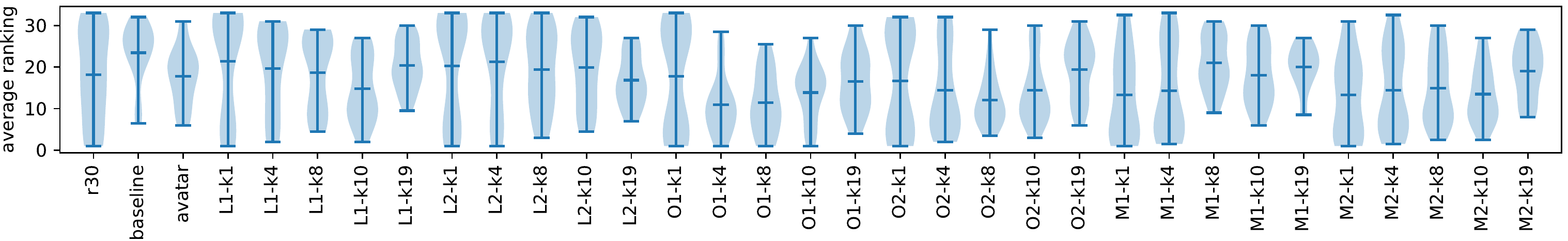}
    \caption{Violin plots depicting the distribution of average performance rankings for search-space reduction strategies across 20 datasets.}
    \label{fig:violin_plots_strategy}
\end{figure*}

We now turn to assessing how well search-space reduction strategies perform in practice.
This process involves examining their resulting error rates, as averaged over five runs per dataset; this averaging procedure is precisely why the prior examination of performance consistency was necessary.
To begin this final analysis, we first seek overarching conclusions across the entire benchmark set of ML problems.
We note that, per dataset, each strategy can be ranked from lowest average loss, i.e.~rank 1, to highest average loss, i.e.~rank 33.
As in previous work~\citep{ngke21}, the distribution of rankings across 20 datasets, per strategy, can then be displayed in violin plot format; the results are shown in Fig.~\ref{fig:violin_plots_strategy}.

Consequently, several key points are apparent.
The optimal-pipeline HPO strategy, \textit{r30}, seems almost uniformly distributed from best to worst across all 20 datasets.
Evidently, a prior 30 hours of AutoML can sometimes be very beneficial in uncovering a highly competitive ML pipeline.
Just as often, however, (1) those 30 hours prove unnecessary, (2) the `ideal' pipeline structure turns out to be a local optimum that HPO on its own cannot escape, or (3) a two-hour HPO from scratch fails to achieve, let alone beat, the original performance of the optimal ML pipeline with its now-forgotten hyperparameters.
As for the $k=30$ runs, \textit{baseline} generally performs poorly, while \textit{avatar} appears middle-of-the-road.
In fact, given that all reduction strategies based on meta-learning employ AVATAR, the \textit{avatar} strategy serves as a good delineation between gambling on meta-knowledge paying off and that gamble going very wrong.
Certainly, for a pool of 30 predictors, $k=19$ already represents forgoing that gamble, as all the performance distributions of the associated strategies start to mimic that of \textit{avatar}, neither excelling nor doing terribly.
In contrast, the $k=1$ strategies represent the most extreme version of that gamble, putting all faith in one predictor.
They are correspondingly bimodal, doing either very well or very poorly.

Closer inspection of Fig.~\ref{fig:violin_plots_strategy} suggests that, in terms of distributions, \textit{LX-k1} is top-heavy, \textit{MX-k1} is bottom-heavy, and \textit{OX-k1} is balanced.
The message seems to be that if an AutoML user wants to gamble on only one predictor, they should go for one that typically performs well, i.e.~trust in a global leaderboard.
However, as $k$ increases, oracle strategies appear to take the lead.
Perhaps the previously mentioned stochastic variability in oracular rankings, compared to the robustness of the global leaderboard, may be mitigated with a larger pool size, allowing for strong performers specific to a dataset to shine through.
Meanwhile, performance distributions for landmarked strategies remain generally underwhelming, reaffirming commentary in Section~\ref{Sec:EvalOracle} about the challenges of leveraging dataset similarity.

\begin{figure*}[!ht]
  \centering 
        \includegraphics[width=0.99\linewidth]{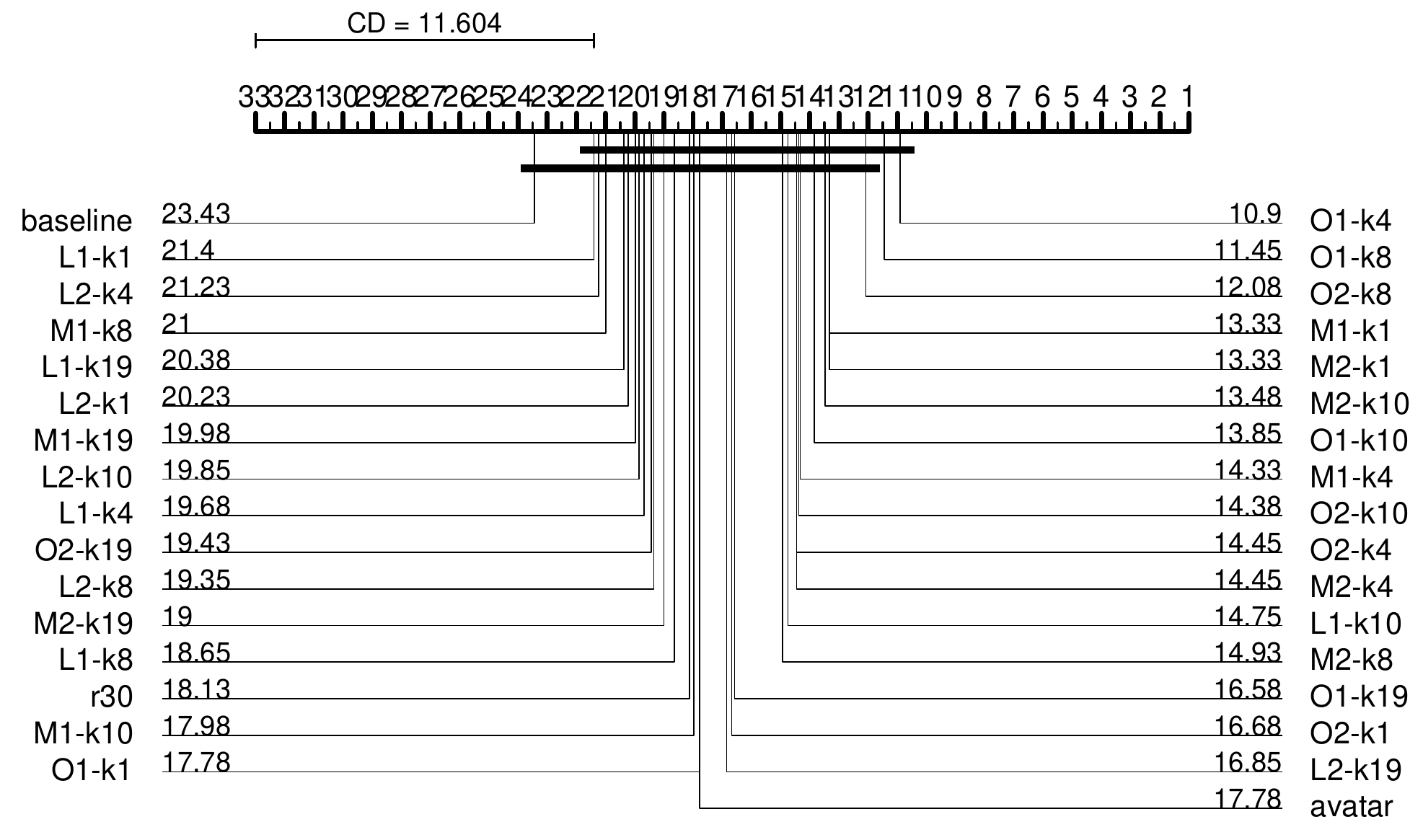}
    \caption{Critical difference diagram displaying the rankings of 33 reduction strategies, as averaged across 20 datasets. Black horizontal bars indicate groups of strategies that are statistically indistinguishable according to the Nemenyi test. Confidence value is $95\%$.}
    \label{fig:cd_diagram-strategy}
\end{figure*}

Each distribution of rankings in Fig.~\ref{fig:violin_plots_strategy} has a mean, and these mean values can be ordered on a diagram of their own, as shown in Fig.~\ref{fig:cd_diagram-strategy}.
Typically, a Nemenyi test~\citep{de06} is used to compare such rankings, defining a critical difference (CD) metric that determines how far apart two methodologies must be, in terms of rankings, to be considered significantly different.
This CD value is dependent on the number of strategies and datasets involved.
In the CD diagram displayed by Fig.~\ref{fig:cd_diagram-strategy}, which is for a confidence value of $95\%$, only the \textit{O1-k4}/\textit{O1-k8} pair of oracle strategies and the \textit{baseline} approach can be considered significantly different, on account of being spread further apart in rankings than the CD value.
On the surface, this result is reasonable, affirming several common-sense hypotheses.
Specifically, prior knowledge is most powerful for a problem when (1) it is drawn from that exact same problem, (2) solved with the exact same constraints, and (3) applied with a little fuzziness to account for uncertainty.

\begin{table}[htb!]
\centering
\caption{Mean performance rankings averaged across all search-space reduction strategies that are of a certain type, use a certain meta-knowledge base, or reduce the search space to a certain size $k$.}
\label{tab:tab_average_ranking_aggregated_by_strategies_metaknowledge_kvalues}
\fontsize{7.2}{8.5}\selectfont

\begin{tabular}{|l|l|l|}
\hline
\textbf{}                                & \textbf{}                   & \textbf{Average ranking} \\ \hline
\multirow{6}{*}{\textbf{Strategies}}     & \textbf{Landmarking}        & 19.24                    \\ \cline{2-3} 
                                         & \textbf{Global leaderboard} & 16.18                    \\ \cline{2-3} 
                                         & \textbf{Oracle}             & 14.76                    \\ \cline{2-3} 
                                         & \textbf{Avatar}             & 17.78                    \\ \cline{2-3} 
                                         & \textbf{Baseline}           & 23.43                    \\ \cline{2-3} 
                                         & \textbf{r30}                & 18.13                    \\ \hline
\multirow{2}{*}{\textbf{Meta-knowledge}} & \textbf{automl-meta}        & 16.80                    \\ \cline{2-3} 
                                         & \textbf{default-meta}       & 16.65                    \\ \hline
\multirow{5}{*}{\textbf{k values}}       & \textbf{k1}                 & 17.12                    \\ \cline{2-3} 
                                         & \textbf{k4}                 & 15.84                    \\ \cline{2-3} 
                                         & \textbf{k8}                 & 16.24                    \\ \cline{2-3} 
                                         & \textbf{k10}                & 15.71                    \\ \cline{2-3} 
                                         & \textbf{k19}                & 18.70                    \\ \hline
\end{tabular}

\end{table}

As with Table~\ref{tab:tab_solution_consistency_metrics_summary}, it is possible to further analyse the performance rankings of strategies via aggregate means, i.e.~grouping and then averaging them by strategy type, meta-knowledge base, and $k$ value.
The results are displayed in Table~\ref{tab:tab_average_ranking_aggregated_by_strategies_metaknowledge_kvalues}.
Recalling that a rank of 17 is the midway point for 33 strategies, the \textit{baseline} strategy is unsurprisingly poor.
Therefore, the adjacent control experiments, \textit{avatar} and \textit{r30}, show noticeable improvements, even though the inflexibility of the latter to go beyond HPO seems to limit its general power.
In the meantime, strategies based on oracular rankings and the global leaderboard do better than average.
Curiously, however, we cannot confirm the expectation of Eq.~(\ref{Eq:Hierarchy}) regarding the placement of landmarked approaches.
One difference between this monograph and prior work~\citep{ngke21} is that we previously chose to exclude any dataset for which at least one strategy experienced five failures, i.e.~no solution received a 10-fold CV evaluation.
Thus, our preliminary analysis was applied to 11 datasets, primarily those more computationally tractable, i.e.~positioned towards the left of Fig.~\ref{fig:f1_automl_number_of_evaluations_heatmap}.
Given that the landmarking process may be relatively fragile compared to using oracular rankings or a global leaderboard, the present inclusion of the other nine datasets with fewer evaluations may weaken the overall performance of \textit{LX-kn}.

Other conclusions suggested by Table~\ref{tab:tab_average_ranking_aggregated_by_strategies_metaknowledge_kvalues} include $k=19$ being too weak of a reduction to provide any performance benefits.
Associated strategies may genuinely be worse than the $k=30$ \textit{avatar} approach, i.e.~$18.70$ versus $17.78$; the optimisation speed may not have improved enough to counterbalance the potential loss of decent predictors.
However, it is also clear that $k=1$ reflects too severe a culling for so much uncertainty in the meta-knowledge bases.
This result supports the notion that AutoML search-space design must find a careful balance between efficient exploitation and cautious exploration.
As for assessments of opportunistic and systematic curation, there continues to be no clear demarcation between the performances they drive.
This outcome is again unintuitive, given that one would expect the predictor rankings of \textit{automl-meta} and \textit{default-meta} to differ substantially.
Further research may be required to determine why both seem equally viable as meta-knowledge for search-space reduction.

Now, one hypothesis for statistically weak aggregate conclusions is that performance trends may depend on problem characteristics, i.e.~only be significant for subsets of the dataset collection.
Indeed, Section~\ref{sec:chap4_dataset_understand} elaborated on the impact that dataset `challenge' could have on informing and assessing reduction strategies.
Unfortunately, no clear metric and threshold were determined to delineate between `hard' and `easy' ML problems.
Simply examining \textit{amazon}, \textit{convex}, \textit{madelon} and maybe \textit{cifar10small} due to their top-heavy predictor-performance distributions in the meta-knowledge bases provides too small a sample size for the Nemenyi test.
However, one can instead divide the benchmark collection between 11 datasets that are computationally lightweight and nine that are heavyweight; the threshold here is whether the dataset has over 1000 single-fold evaluations in \textit{automl-meta} according to Fig.~\ref{fig:f1_automl_number_of_evaluations_heatmap}.
This partitioning of tractability is also a reasonable reflection of which datasets require the shortest and longest predictor evaluation times in \textit{default-meta}, which is displayed in Fig.~\ref{fig:f2_default_time_of_evaluation_heatmap}.
Accordingly, generating a CD diagram like Fig.~\ref{fig:cd_diagram-strategy} for the 11 most tractable datasets suggests that it is now \textit{M1-k4} and \textit{M2-k10} that are significantly better than \textit{baseline}, whereas, for the nine most intractable datasets, insufficient information exists to be confident in any significant differences.

\begin{table}[htb!]
\centering
\caption {Fraction of lightweight and heavyweight datasets for which a search-space reduction strategy is ranked in the top 16 out of 33. A `lightweight' dataset and a `heavyweight' dataset have more than and less than 1000 single-fold evaluations, respectively, in \textit{automl-meta}.}
\label{tab:top_sixteen}
\begin{tabular}{|c|c|c|}
\hline
Strategy & \begin{tabular}[c]{@{}c@{}}Fraction of\\ Lightweight Datasets\\ With Strategy in Top 16\end{tabular} & \begin{tabular}[c]{@{}c@{}}Fraction of\\ Heavyweight Datasets\\ With Strategy in Top 16\end{tabular} \\ \hline
r30      & \cellcolor[HTML]{FCB87A}0.27                                                                         & \cellcolor[HTML]{63BE7B}0.67                                                                         \\ \hline
baseline & \cellcolor[HTML]{F8696B}0.00                                                                         & \cellcolor[HTML]{FDCA7D}0.33                                                                         \\ \hline
avatar   & \cellcolor[HTML]{FCB87A}0.27                                                                         & \cellcolor[HTML]{B1D580}0.56                                                                         \\ \hline
L1-k1    & \cellcolor[HTML]{FDD37F}0.36                                                                         & \cellcolor[HTML]{FDCA7D}0.33                                                                         \\ \hline
L2-k1    & \cellcolor[HTML]{FDD37F}0.36                                                                         & \cellcolor[HTML]{FFEB84}0.44                                                                         \\ \hline
O1-k1    & \cellcolor[HTML]{FDD37F}0.36                                                                         & \cellcolor[HTML]{B1D580}0.56                                                                         \\ \hline
O2-k1    & \cellcolor[HTML]{B9D780}0.55                                                                         & \cellcolor[HTML]{FFEB84}0.44                                                                         \\ \hline
M1-k1    & \cellcolor[HTML]{79C57D}0.64                                                                         & \cellcolor[HTML]{B1D580}0.56                                                                         \\ \hline
M2-k1    & \cellcolor[HTML]{79C57D}0.64                                                                         & \cellcolor[HTML]{FFEB84}0.44                                                                         \\ \hline
\end{tabular}
\end{table}

However, the conservative Nemenyi test is not without limitation; for instance, it assumes the performances of each strategy across all datasets are normally distributed and homogenous in variance.
This assumption is most significantly violated by the `all-or-nothing' $k=1$ strategies, which are in the top 16 strategies for approximately half of the ML problems and the bottom 16 for the other half.
Nemenyi tests applied to either half of these different partitions primarily reinforce the significant weaknesses of the landmarking approach used in this work, wherever significant differences are identifiable, so we examine another angle instead.
Partitioning datasets once more by the \textit{automl-meta} 1000-evaluation rule, Table~\ref{tab:top_sixteen} shows the fraction of datasets in both groups for which `control' and $k=1$ strategies are found to be in the top 16.
Immediately, it is clear that \textit{baseline} and \textit{avatar} are generally uncompetitive if substantial meta-knowledge exists to inform better strategies, whereas, when evaluations are sparse, extensive ML model searches become more reasonable.
Similarly, having 30 hours of prior ML pipeline search, i.e.~\textit{r30}, seems particularly useful for computational heavyweights.

In terms of the $k=1$ strategies, Table~\ref{tab:top_sixteen} suggests the \textit{automl-meta} oracle is counterintuitively both weak when evaluations are many and strong when they are sparse.
The weakness may be due to \textit{automl-meta} rankings being based on the average of single-fold evaluations, not the optimum.
These predictor-performance averages may shift substantially depending on the optimisation path of SMAC through hyperparameter space, which pipeline structures will only complicate, and thus the exactness of oracular rankings may be substantially unreliable.
In contrast, stochastic variability may not be an essential factor for heavyweight datasets, where having \textit{any} idea of what can/cannot run with decent performance within two hours may outweigh a more nuanced ordering of predictors.
For related reasons, this may be why \textit{O2-k1} performs worse than \textit{O1-k1} when \textit{automl-meta} evaluations are few; the \textit{default-meta} rankings do not care about time constraints, meaning that they are less reliable for two-hour SMAC runs applied to heavyweight datasets.
However, \textit{default-meta} rankings are not subject to any noisy averaging process, so this may be why \textit{O2-k1} outperforms \textit{O1-k1} for the computational lightweights, i.e.~default hyperparameters inspired by expert knowledge may hew closer to optimal error rates or, failing that, how they are proportionally distributed.
Of course, Table~\ref{tab:top_sixteen} indicates that it is perhaps best, if pursuing an all-or-nothing approach, to gamble on robust generalist ML algorithms, particularly if tackling a lightweight ML problem.
Notably, due to a quirk of implementation relating to meta-predictors, explained in Section~\ref{sec:chap4_settings}, \textit{M2-k1} is practically equivalent to \textit{M1-k1}.
Consequently, as Fig.~\ref{fig:f4_automl_predictor_rankings_per_dataset_heatmap} and Fig.~\ref{fig:f5_default_predictor_rankings_per_dataset_heatmap} indicate, both strategies essentially recommend RandomForest.
Admittedly, this leaderboard-based suggestion is found to be marginally less helpful for the computational heavyweights, again potentially due to these ML problems being much more sensitive to poor choices that do not compute well within two hours.

On another topic, given that oracular perfection underpinned the analysis in Section~\ref{Sec:EvalOracle}, the non-ideal performance of \textit{O1-k1} may seem surprising.
However, we re-emphasise that Table~\ref{tab:tab_average_ranking_aggregated_by_strategies_metaknowledge_kvalues} showed oracle-based strategies reign supreme.
Indeed, \textit{O1-k1} is simply more susceptible to stochastic variation than the global leaderboard, which, as an average across 20 datasets, suppresses uncorrelated noise.
Consider 11 datasets where \textit{O1-k1} is one of the worst 16 reduction strategies; for these problems, the average rank of an oracle-based strategy is 17.56 compared to the 15.79 of \textit{MX-kn}.
However, after solely excluding $k=1$ strategies from the average, \textit{OX-kn} moves to the front with a mean rank of 15.86, while strategies based on the global leaderboard now rank 16.09 on average.
This result suggests that $k\geq 4$ is usually enough to counter the effect of unreliable oracular rankings.
In contrast, consider eight datasets where \textit{M1-k1} is in the bottom 16; for these problems, the average rank of an oracle-based strategy is 14.46 compared to the 21.22 of \textit{MX-kn}.
Removing the $k=1$ strategies, the average ranks for \textit{OX-kn} and \textit{MX-kn} become 14.02 and 20.55, respectively, i.e.~global-leaderboard strategies remain uncompetitive.
This result confirms that \textit{M1-k1} is relatively robust against stochastic variability, and performance failures have more to do with the global leaderboard failing to represent a particular ML problem adequately.
There are, of course, other implications that aggregate results in this work suggest.
For instance, on the topic of \textit{MX-kn} strategies, the CD diagram in Fig.~\ref{fig:cd_diagram-strategy} suggests that the performance dip from \textit{M1-k4} to \textit{M1-k8} is substantially worse than going from \textit{M2-k4} to \textit{M2-k8}, despite only adding seemingly simple predictors to a search space, i.e.~NaiveBayes, SimpleLogistic, REPTree, and Logistic.
Perhaps this outcome reflects a circumstance where SMAC faces decision paralysis between appealing options, being slow to exclude options and thus to reach an optimum.
Nonetheless, we restrain from further commentary as it remains difficult with a sample size of 20 datasets to gauge how genuine such effects are.

\begin{table*}[htb!]

\vspace{-0.35cm}

\caption {Best pipeline structures found by strategy-assisted SMAC for dataset \textit{amazon}. Each search-space reduction strategy was run five times. Missing values denote the absence of a full 10-fold CV evaluation during a run.}
\label{tab:tab_pipelines_amazon}

\centering
\tiny
\setlength\tabcolsep{0.5pt}

\makebox[\linewidth]{
\begin{tabular}{|l|l|l|l|l|l|}
\hline
\textbf{Strategies} & \textbf{Run \#1}                                                                                                                                                                                                                               & \textbf{Run \#2}                                                                                                                                                                            & \textbf{Run \#3}                                                                                                                                                                            & \textbf{Run \#4}                                                                                                                                                                                                            & \textbf{Run \#5}                                                                                                                                                                                            \\ \hline
\textbf{r30}        & \begin{tabular}[c]{@{}l@{}}ReplaceMissingValues\\      $\rightarrow$ Normalize\\      $\rightarrow$ RandomSubset\\      $\rightarrow$ NaiveBayesMultinomial (P1)\\      $\rightarrow$ RandomSubSpace (P28)\end{tabular}                                                    & \begin{tabular}[c]{@{}l@{}}ReplaceMissingValues\\      $\rightarrow$ Normalize\\      $\rightarrow$ RandomSubset\\      $\rightarrow$ NaiveBayesMultinomial (P1)\\      $\rightarrow$ RandomSubSpace (P28)\end{tabular} & \begin{tabular}[c]{@{}l@{}}ReplaceMissingValues\\      $\rightarrow$ Normalize\\      $\rightarrow$ RandomSubset\\      $\rightarrow$ NaiveBayesMultinomial (P1)\\      $\rightarrow$ RandomSubSpace (P28)\end{tabular} & \begin{tabular}[c]{@{}l@{}}ReplaceMissingValues\\      $\rightarrow$ Normalize\\      $\rightarrow$ RandomSubset\\      $\rightarrow$ NaiveBayesMultinomial (P1)\\      $\rightarrow$ RandomSubSpace (P28)\end{tabular}                                 & \begin{tabular}[c]{@{}l@{}}ReplaceMissingValues\\      $\rightarrow$ Normalize\\      $\rightarrow$ RandomSubset\\      $\rightarrow$ NaiveBayesMultinomial (P1)\\      $\rightarrow$ RandomSubSpace (P28)\end{tabular}                 \\ \hline
\textbf{baseline}   & JRip (P12)                                                                                                                                                                                                                                     & -                                                                                                                                                                                           & -                                                                                                                                                                                           & NaiveBayes (P0)                                                                                                                                                                                                             & -                                                                                                                                                                                                           \\ \hline
\textbf{avatar}     & PART (P14)                                                                                                                                                                                                                                     & -                                                                                                                                                                                           & NaiveBayesMultinomial (P1)                                                                                                                                                                  & NaiveBayesMultinomial (P1)                                                                                                                                                                                                  & -                                                                                                                                                                                                           \\ \hline
\textbf{L1-k1}      & \begin{tabular}[c]{@{}l@{}}ReplaceMissingValues\\      $\rightarrow$ RemoveOutliers\\      $\rightarrow$ InterquartileRange\\      $\rightarrow$ Normalize\\      $\rightarrow$ PrincipalComponents\\      $\rightarrow$ ReservoirSample\\      $\rightarrow$ Logistic (P2)\end{tabular} & \begin{tabular}[c]{@{}l@{}}SpreadSubsample\\      $\rightarrow$ Center\\      $\rightarrow$ RandomSubset\\      $\rightarrow$ Resample\\      $\rightarrow$ Logistic (P2)\end{tabular}                                  & \begin{tabular}[c]{@{}l@{}}ClassBalancer\\      $\rightarrow$ RemoveOutliers\\      $\rightarrow$ InterquartileRange\\      $\rightarrow$ Center $\rightarrow$ Logistic (P2)\end{tabular}                               & \begin{tabular}[c]{@{}l@{}}ClassBalancer\\      $\rightarrow$ RemoveOutliers\\      $\rightarrow$ InterquartileRange\\      $\rightarrow$ Standardize\\      $\rightarrow$ RandomSubset\\      $\rightarrow$ Resample\\      $\rightarrow$ Logistic (P2)\end{tabular} & \begin{tabular}[c]{@{}l@{}}ReplaceMissingValues\\      $\rightarrow$ Standardize\\      $\rightarrow$ PrincipalComponents\\      $\rightarrow$ Logistic (P2)\end{tabular}                                                        \\ \hline
\textbf{L1-k4}      & NaiveBayesMultinomial (P1)                                                                                                                                                                                                                     & \begin{tabular}[c]{@{}l@{}}SMO (P6)\\      {[}NormalizedPolyKernel{]}\end{tabular}                                                                                                          & SMO (P6) {[}PolyKernel{]}                                                                                                                                                                   & \begin{tabular}[c]{@{}l@{}}SMO (P6)\\      {[}NormalizedPolyKernel{]}\end{tabular}                                                                                                                                          & SMO (P6) {[}PolyKernel{]}                                                                                                                                                                                   \\ \hline
\textbf{L1-k8}      & NaiveBayesMultinomial (P1)                                                                                                                                                                                                                     & -                                                                                                                                                                                           & -                                                                                                                                                                                           & NaiveBayesMultinomial (P1)                                                                                                                                                                                                  & -                                                                                                                                                                                                           \\ \hline
\textbf{L1-k10}     & -                                                                                                                                                                                                                                              & NaiveBayesMultinomial (P1)                                                                                                                                                                  & NaiveBayesMultinomial (P1)                                                                                                                                                                  & -                                                                                                                                                                                                                           & -                                                                                                                                                                                                           \\ \hline
\textbf{L1-k19}     & SimpleLogistic (P5)                                                                                                                                                                                                                            & NaiveBayesMultinomial (P1)                                                                                                                                                                  & NaiveBayesMultinomial (P1)                                                                                                                                                                  & NaiveBayes (P0)                                                                                                                                                                                                             & NaiveBayesMultinomial (P1)                                                                                                                                                                                  \\ \hline
\textbf{L2-k1}      & \begin{tabular}[c]{@{}l@{}}SMO (P6)\\      {[}NormalizedPolyKernel{]}\end{tabular}                                                                                                                                                             & SMO (P6) {[}PolyKernel{]}                                                                                                                                                                   & SMO (P6) {[}PolyKernel{]}                                                                                                                                                                   & \begin{tabular}[c]{@{}l@{}}SMO (P6)\\      {[}NormalizedPolyKernel{]}\end{tabular}                                                                                                                                          & SMO (P6) {[}PolyKernel{]}                                                                                                                                                                                   \\ \hline
\textbf{L2-k4}      & -                                                                                                                                                                                                                                              & -                                                                                                                                                                                           & -                                                                                                                                                                                           & -                                                                                                                                                                                                                           & -                                                                                                                                                                                                           \\ \hline
\textbf{L2-k8}      & NaiveBayesMultinomial (P1)                                                                                                                                                                                                                     & NaiveBayesMultinomial (P1)                                                                                                                                                                  & NaiveBayesMultinomial (P1)                                                                                                                                                                  & -                                                                                                                                                                                                                           & NaiveBayesMultinomial (P1)                                                                                                                                                                                  \\ \hline
\textbf{L2-k10}     & -                                                                                                                                                                                                                                              & -                                                                                                                                                                                           & -                                                                                                                                                                                           & NaiveBayesMultinomial (P1)                                                                                                                                                                                                  & -                                                                                                                                                                                                           \\ \hline
\textbf{L2-k19}     & NaiveBayesMultinomial (P1)                                                                                                                                                                                                                     & NaiveBayesMultinomial (P1)                                                                                                                                                                  & -                                                                                                                                                                                           & NaiveBayesMultinomial (P1)                                                                                                                                                                                                  & NaiveBayesMultinomial (P1)                                                                                                                                                                                  \\ \hline
\textbf{O1-k1}      & -                                                                                                                                                                                                                                              & -                                                                                                                                                                                           & -                                                                                                                                                                                           & -                                                                                                                                                                                                                           & -                                                                                                                                                                                                           \\ \hline
\textbf{O1-k4}      & -                                                                                                                                                                                                                                              & -                                                                                                                                                                                           & NaiveBayesMultinomial (P1)                                                                                                                                                                  & -                                                                                                                                                                                                                           & NaiveBayesMultinomial (P1)                                                                                                                                                                                  \\ \hline
\textbf{O1-k8}      & NaiveBayesMultinomial (P1)                                                                                                                                                                                                                     & -                                                                                                                                                                                           & NaiveBayesMultinomial (P1)                                                                                                                                                                  & NaiveBayesMultinomial (P1)                                                                                                                                                                                                  & NaiveBayesMultinomial (P1)                                                                                                                                                                                  \\ \hline
\textbf{O1-k10}     & -                                                                                                                                                                                                                                              & -                                                                                                                                                                                           & NaiveBayesMultinomial (P1)                                                                                                                                                                  & -                                                                                                                                                                                                                           & -                                                                                                                                                                                                           \\ \hline
\textbf{O1-k19}     & NaiveBayesMultinomial (P1)                                                                                                                                                                                                                     & NaiveBayes (P0)                                                                                                                                                                             & -                                                                                                                                                                                           & NaiveBayesMultinomial (P1)                                                                                                                                                                                                  & NaiveBayesMultinomial (P1)                                                                                                                                                                                  \\ \hline
\textbf{O2-k1}      & -                                                                                                                                                                                                                                              & -                                                                                                                                                                                           & -                                                                                                                                                                                           & -                                                                                                                                                                                                                           & -                                                                                                                                                                                                           \\ \hline
\textbf{O2-k4}      & -                                                                                                                                                                                                                                              & NaiveBayesMultinomial (P1)                                                                                                                                                                  & -                                                                                                                                                                                           & NaiveBayesMultinomial (P1)                                                                                                                                                                                                  & NaiveBayesMultinomial (P1)                                                                                                                                                                                  \\ \hline
\textbf{O2-k8}      & NaiveBayesMultinomial (P1)                                                                                                                                                                                                                     & NaiveBayesMultinomial (P1)                                                                                                                                                                  & -                                                                                                                                                                                           & NaiveBayesMultinomial (P1)                                                                                                                                                                                                  & NaiveBayesMultinomial (P1)                                                                                                                                                                                  \\ \hline
\textbf{O2-k10}     & NaiveBayesMultinomial (P1)                                                                                                                                                                                                                     & -                                                                                                                                                                                           & NaiveBayesMultinomial (P1)                                                                                                                                                                  & NaiveBayesMultinomial (P1)                                                                                                                                                                                                  & -                                                                                                                                                                                                           \\ \hline
\textbf{O2-k19}     & -                                                                                                                                                                                                                                              & -                                                                                                                                                                                           & -                                                                                                                                                                                           & -                                                                                                                                                                                                                           & -                                                                                                                                                                                                           \\ \hline
\textbf{M1-k1}      & \begin{tabular}[c]{@{}l@{}}ClassBalancer\\      $\rightarrow$ ReplaceMissingValues\\      $\rightarrow$ Standardize\\      $\rightarrow$ PrincipalComponents\\      $\rightarrow$ RandomForest (P19)\end{tabular}                                                          & \begin{tabular}[c]{@{}l@{}}SpreadSubsample\\      $\rightarrow$ ReplaceMissingValues\\      $\rightarrow$ RandomForest (P19)\end{tabular}                                                                 & RandomForest (P19)                                                                                                                                                                          & RandomForest (P19)                                                                                                                                                                                                          & \begin{tabular}[c]{@{}l@{}}ClassBalancer\\      $\rightarrow$ RemoveOutliers\\      $\rightarrow$ InterquartileRange\\      $\rightarrow$ Center\\      $\rightarrow$ PrincipalComponents\\      $\rightarrow$ RandomForest (P19)\end{tabular} \\ \hline
\textbf{M1-k4}      & J48 (P17)                                                                                                                                                                                                                                      & JRip (P12)                                                                                                                                                                                  & JRip (P12)                                                                                                                                                                                  & J48 (P17)                                                                                                                                                                                                                   & J48 (P17)                                                                                                                                                                                                   \\ \hline
\textbf{M1-k8}      & -                                                                                                                                                                                                                                              & -                                                                                                                                                                                           & -                                                                                                                                                                                           & -                                                                                                                                                                                                                           & -                                                                                                                                                                                                           \\ \hline
\textbf{M1-k10}     & SimpleLogistic (P5)                                                                                                                                                                                                                            & NaiveBayes (P0)                                                                                                                                                                             & NaiveBayes (P0)                                                                                                                                                                             & -                                                                                                                                                                                                                           & -                                                                                                                                                                                                           \\ \hline
\textbf{M1-k19}     & NaiveBayesMultinomial (P1)                                                                                                                                                                                                                     & NaiveBayesMultinomial (P1)                                                                                                                                                                  & -                                                                                                                                                                                           & -                                                                                                                                                                                                                           & -                                                                                                                                                                                                           \\ \hline
\textbf{M2-k1}      & \begin{tabular}[c]{@{}l@{}}Standardize\\      $\rightarrow$ RandomForest (P19)\end{tabular}                                                                                                                                                           & \begin{tabular}[c]{@{}l@{}}RemoveOutliers\\      $\rightarrow$ InterquartileRange\\      $\rightarrow$ Normalize\\      $\rightarrow$ RandomForest (P19)\end{tabular}                                            & RandomForest (P19)                                                                                                                                                                          & RandomForest (P19)                                                                                                                                                                                                          & RandomForest (P19)                                                                                                                                                                                          \\ \hline
\textbf{M2-k4}      & RandomForest (P19)                                                                                                                                                                                                                             & SimpleLogistic (P5)                                                                                                                                                                         & -                                                                                                                                                                                           & -                                                                                                                                                                                                                           & SimpleLogistic (P5)                                                                                                                                                                                         \\ \hline
\textbf{M2-k8}      & -                                                                                                                                                                                                                                              & -                                                                                                                                                                                           & SimpleLogistic (P5)                                                                                                                                                                         & -                                                                                                                                                                                                                           & -                                                                                                                                                                                                           \\ \hline
\textbf{M2-k10}     & SimpleLogistic (P5)                                                                                                                                                                                                                            & \begin{tabular}[c]{@{}l@{}}SMO (P6)\\      {[}NormalizedPolyKernel{]}\end{tabular}                                                                                                          & -                                                                                                                                                                                           & J48 (P17)                                                                                                                                                                                                                   & \begin{tabular}[c]{@{}l@{}}SMO (P6)\\      {[}NormalizedPolyKernel{]}\end{tabular}                                                                                                                          \\ \hline
\textbf{M2-k19}     & -                                                                                                                                                                                                                                              & -                                                                                                                                                                                           & J48 (P17)                                                                                                                                                                                   & NaiveBayes (P0)                                                                                                                                                                                                             & -                                                                                                                                                                                                           \\ \hline
\end{tabular}
}

\end{table*}

\begin{figure}[ht!]
\makebox[\linewidth][c]{%
    \subfloat[amazon\label{fig:stem_chart_amazon_cut}]{%
        \includegraphics[height=1.05\linewidth]{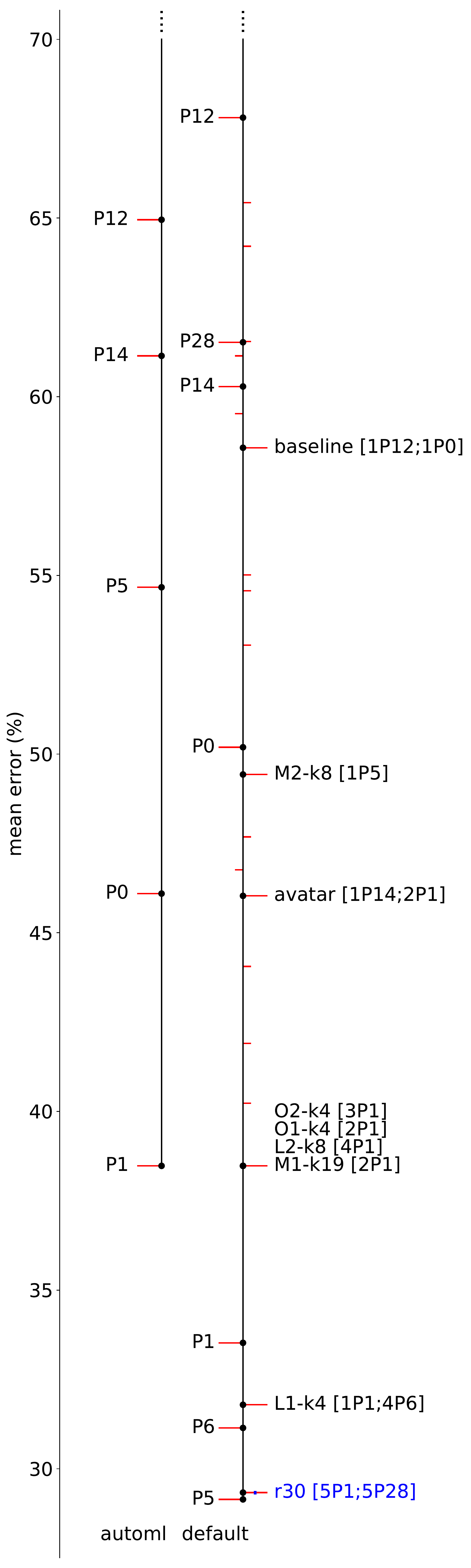}
        }%
    \hfill
    \subfloat[kddcup\label{fig:stem_chart_kddcup_cut}]{%
        \includegraphics[height=1.05\linewidth]{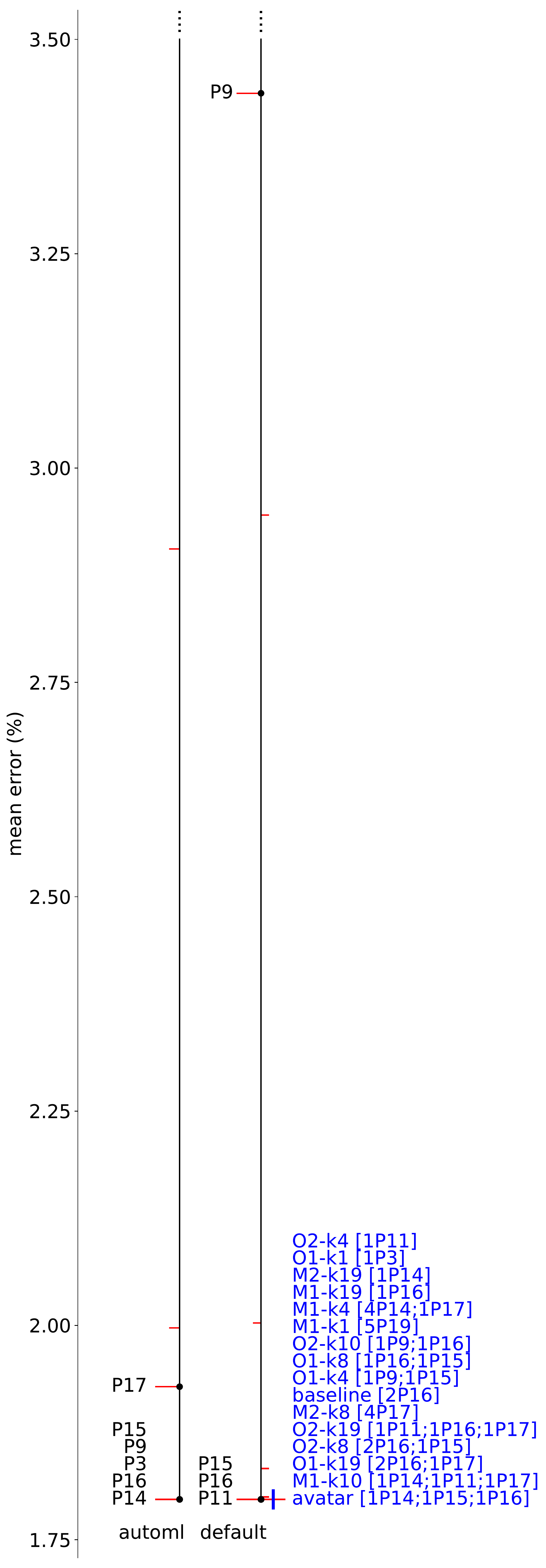}
        }%
    \hfill
    \subfloat[gcredit\label{fig:stem_chart_gcredit_cut}]{%
        \includegraphics[height=1.05\linewidth]{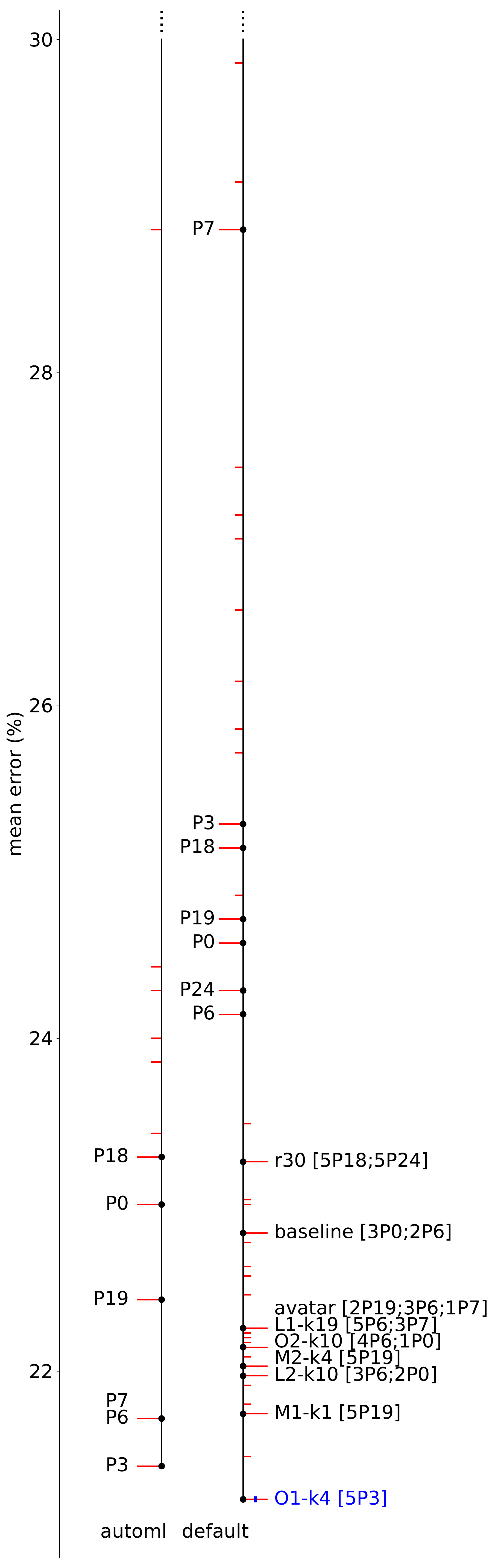}
        }
}
    \caption{Stem plot visualisation for the performance of predictors and search-space reduction strategies on different datasets. Notches to the left of the `automl' and `default' lines denote the optimal and solitary 10-fold CV error rates of predictors in \textit{automl-meta} and \textit{default-meta}, respectively, if they exist. Notches to the right of the `default' line denote the five-run-averaged error rates of ML solutions returned by strategy-assisted AutoML. Strategy notches are extended and labelled if they are statistically indistinguishable from the best (i.e.~blue) and/or represent \textit{baseline}, \textit{avatar}, \textit{r30}, or the best-performing approach per strategy type and meta-knowledge base. Predictor notches are extended and labelled if they associate with an optimal solution returned by a labelled strategy; see Table~\ref{tab:list_predictor} for symbols.}
    \label{fig:stem_charts_all_datasets_cut}
\end{figure}      

\begin{figure}[ht!]\ContinuedFloat 
\makebox[\linewidth][c]{%
    \subfloat[abalone\label{fig:stem_chart_abalone_cut}]{%
        \includegraphics[height=1.05\linewidth]{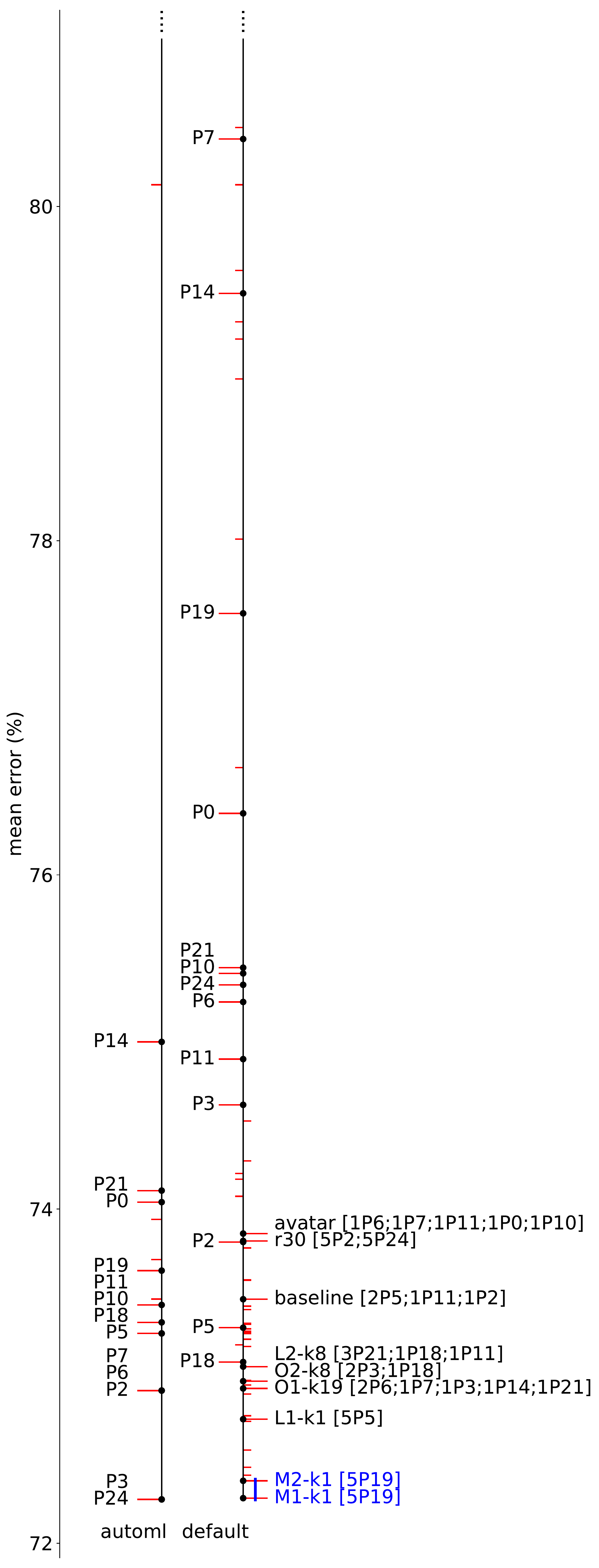}
        }%
    \hfill
    \subfloat[dexter\label{fig:stem_chart_dexter_cut}]{%
        \includegraphics[height=1.05\linewidth]{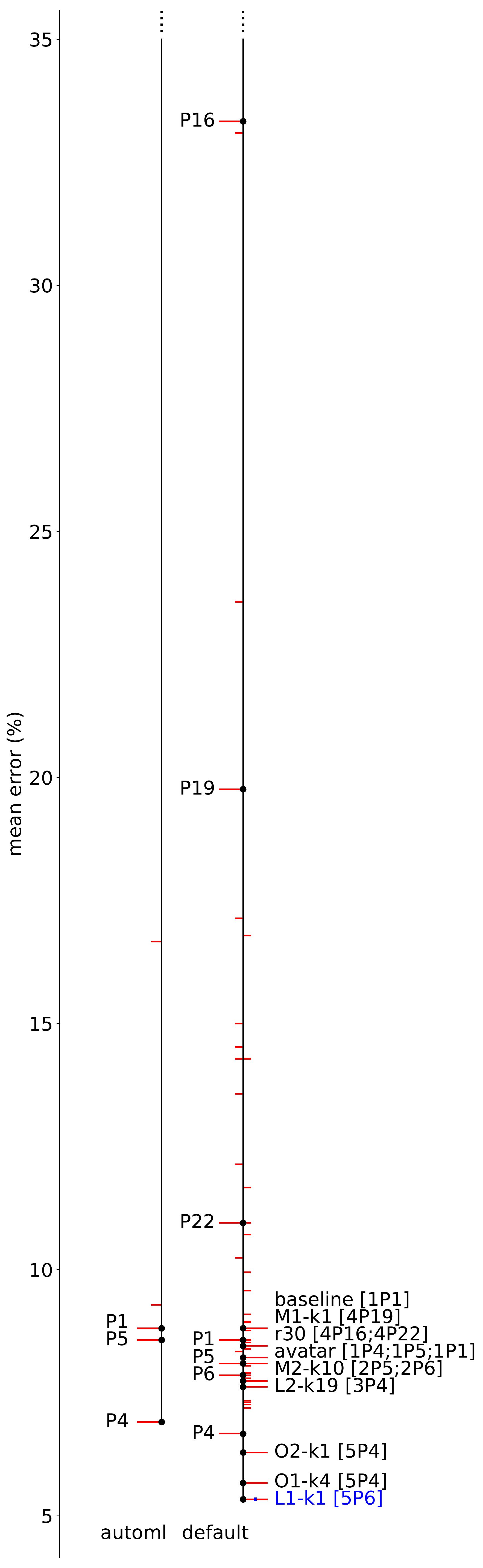}
        }%
    \hfill
    \subfloat[adult\label{fig:stem_chart_adult_cut}]{%
        \includegraphics[height=1.05\linewidth]{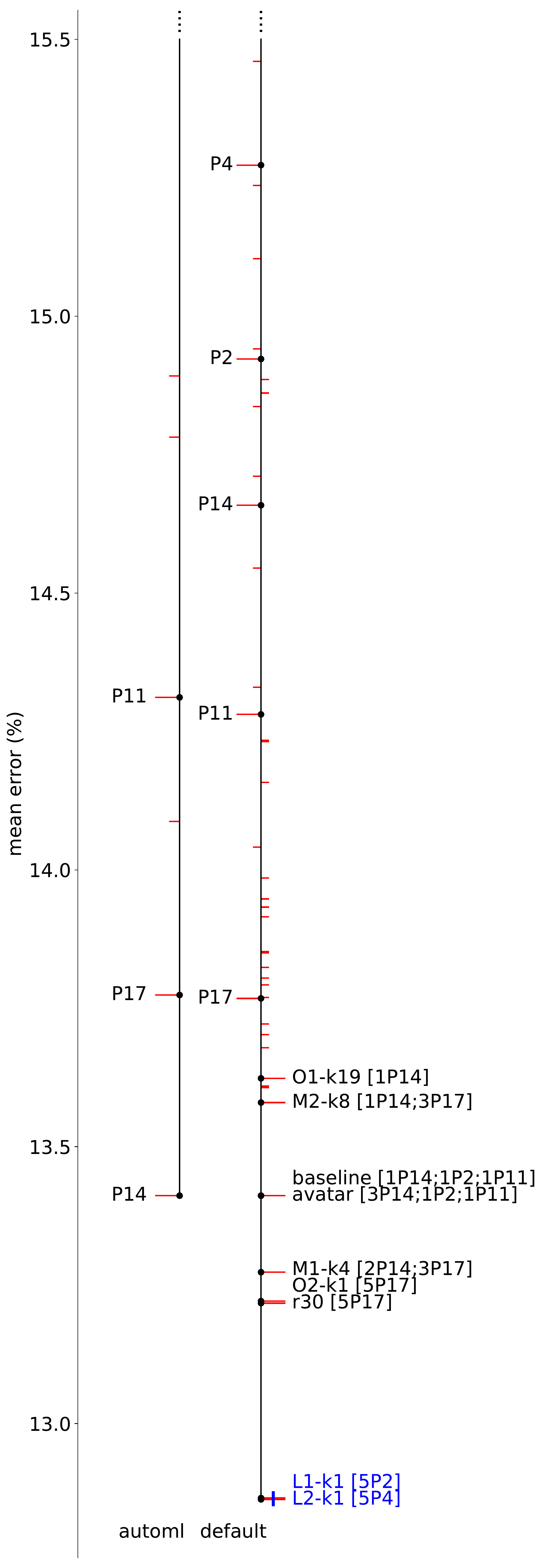}
        }
}
    \caption{Stem plot visualisation for the performance of predictors and search-space reduction strategies on different datasets. Notches to the left of the `automl' and `default' lines denote the optimal and solitary 10-fold CV error rates of predictors in \textit{automl-meta} and \textit{default-meta}, respectively, if they exist. Notches to the right of the `default' line denote the five-run-averaged error rates of ML solutions returned by strategy-assisted AutoML. Strategy notches are extended and labelled if they are statistically indistinguishable from the best (i.e.~blue) and/or represent \textit{baseline}, \textit{avatar}, \textit{r30}, or the best-performing approach per strategy type and meta-knowledge base. Predictor notches are extended and labelled if they associate with an optimal solution returned by a labelled strategy; see Table~\ref{tab:list_predictor} for symbols.}
\end{figure}

Ultimately, forming a generalised commentary on the performance of AutoML search-space reduction strategies is difficult, as such conclusions miss nuances specific to individual datasets.
Thus, we briefly highlight a few representative results here, beginning with the example of \textit{amazon}, a top-heavy `hard' dataset.
The optimal ML pipelines returned by every strategy for \textit{amazon} are listed in Table~\ref{tab:tab_pipelines_amazon}, where predictors are associated with the `P' symbols introduced in Table~\ref{tab:list_predictor}.
Additional information is provided by Fig.~\ref{fig:stem_charts_all_datasets_cut}, specifically Fig.~\ref{fig:stem_chart_amazon_cut} for \textit{amazon}, which visualises the performance of select strategies and compares them against the performance of predictors in \textit{automl-meta} and \textit{default-meta}.
Crucially, the value of each notch in the stem plot related to \textit{automl-meta} marks the \textit{best} 10-fold CV error rate for a predictor in the meta-knowledge base, if evaluated, \textit{not} the average across all single-fold evaluations that determined predictor rankings in this work.
In effect, Fig.~\ref{fig:stem_charts_all_datasets_cut} assesses whether ML pipelines returned by strategy-assisted AutoML manage to outperform both the best pipelines found in \textit{automl-meta} and the default hyperparameters tested in \textit{default-meta}.

For general context, it is first worth noting that the best strategies returned RandomForest (P19) in eight out of 20 datasets, exclusively for seven.
With the RBFKernel (P7) assisting once, SMO (P6) is a runner-up, exclusively optimising performance for four datasets.
Culling an AutoML search space to just RandomForest and SMO would thus cater to $60\%$ of the diverse benchmark set used in this work.
In fact, with SMO featuring as four out of five solutions for \textit{L1-k4}, Fig.~\ref{fig:stem_chart_amazon_cut} indicates that the \textit{amazon} dataset almost joins this collection.
However, \textit{amazon} is a computational heavyweight, with Table~\ref{tab:tab_pipelines_amazon} showing some strategies failing to complete at all, and, given that Fig.~\ref{fig:f2_default_time_of_evaluation_heatmap} suggests the Weka implementation of SMO is often relatively slow to run, it remains difficult to hone in on the perfect hyperparameters.
Instead, NaiveBayesMultinomial (P1) is evidently preferred for \textit{amazon}, uniquely among all 20 datasets, as an easier way to attain high performance, especially in large pools of options.
The table shows that the moment $k$ becomes large enough for any strategy to include the predictor, it will be prioritised.
This fact is particularly problematic for strategies based on the global leaderboard, with Fig.~\ref{fig:f4_automl_predictor_rankings_per_dataset_heatmap} and Fig.~\ref{fig:f5_default_predictor_rankings_per_dataset_heatmap} showing NaiveBayesMultinomial ranked 18th and 25th overall in \textit{automl-meta} and \textit{default-meta}, respectively.
Sure enough, Fig.~\ref{fig:stem_chart_amazon_cut} reveals the best \textit{M1-kn} strategy is for $k=19$, attaining a loss below $40\%$, while \textit{M2-kn} remains uncompetitive for all tested $k$ values; at an optimum of $k=8$, \textit{M2-kn} scores a loss of ${\sim}50\%$.

Importantly, \textit{amazon} is a good showcase for several eventualities, including being one of three datasets where \textit{r30} proved optimal; the other two are \textit{gisette} and \textit{krvskp}.
This result is unsurprising for the computational heavyweights of \textit{amazon} and \textit{gisette}, as, again, the reduced productivity of a two-hour optimisation leaves the 30 hours of prior pipeline search much more unbeatable than for the lightweight datasets.
It is also a reminder that multi-component ML pipelines can genuinely be worth their complexity.
For \textit{amazon}, most reduction strategies do not afford SMAC the time to explore this complicated space, yet, for \textit{r30} and several $k=1$ strategies, Table~\ref{tab:tab_pipelines_amazon} reveals that the inclusion of preprocessors and meta-predictors can improve the overall accuracy.
Solution optimality is often a matter of establishing an ML pipeline with appropriate expressive power, which is why universal approximators are so appealing, and the nonlinearities that components such as RandomSubset and RandomSubSpace (P28) introduce help nudge NaiveBayesMultinomial in this direction.

Another crucial observation is that, as Fig.~\ref{fig:stem_chart_amazon_cut} shows, the best solution-search strategy for \textit{amazon} is one that returns the predictor found to be optimal in \textit{automl-meta}.
This outcome is not universally the case, only holding for 13 datasets.
It is even less the case when considering alignment with \textit{default-meta}; optimal strategies return the best predictor in \textit{default-meta} for less than half of the 20 benchmark datasets.
For instance, despite SimpleLogistic (P5) with default hyperparameters performing remarkably well for \textit{amazon}, Table~\ref{tab:tab_pipelines_amazon} indicates that it is only returned by a few unimpressive \textit{M2-kn} strategies.
This result makes sense, as meta-knowledge is most relevant if the experimental context of its generation matches that of its use.
Regardless, the alignment of the best strategy and the best \textit{automl-meta} predictor is notable for another reason.
All four `top-heavy' datasets identified in Section~\ref{sec:chap4_dataset_understand}, i.e.~\textit{amazon}, \textit{convex}, \textit{madelon} and \textit{cifar10small}, feature within the set of 13 adherents.
Likewise, this means that an optimal strategy depended on locating the best \textit{automl-meta} predictor for only $56.25\%$ of the remaining `bottom-heavy' datasets.
Thus, admittedly limited by a small sample size, this observation reaffirms that intelligent predictor selection is most vital to a `hard' dataset where there is a distinguishable `best' predictor.
Accordingly, while our research did not find evidence that any \textit{specific} strategies were better suited for this notion of `challenge', its influence on AutoML outcome remains clear.

The remaining stem plots in Fig.~\ref{fig:stem_charts_all_datasets_cut} have been selected to demonstrate why singular guidelines for AutoML search-space reduction are so difficult to establish.
Clearly, reductions informed by meta-knowledge are desirable.
The best strategy for virtually every dataset outperforms the best predictors in \textit{automl-meta} and \textit{default-meta}; the exceptions to the rule, e.g.~default SimpleLogistic (P5) beating \textit{r30} for \textit{amazon}, appear insubstantial.
This is no surprise, especially for \textit{automl-meta}, which is effectively the amalgamation of five \textit{avatar} runs, i.e.~an expansive $k=30$ search.
Nonetheless, outcomes can vary immensely.
An ML problem such as \textit{kddcup} might be involved, which Fig.~\ref{fig:stem_chart_kddcup_cut} shows is sparsely covered by 10-fold CV evaluations that tend to perform indistinguishably well.
Essentially, anything goes \textit{if} it can run.
This `easy' computational heavyweight exemplifies the potential for an ML problem to distort aggregate predictor rankings and suggest spurious dataset similarities if included in a meta-knowledge base.
Likewise, it is a poor benchmarker for search-space reduction strategies.

Then there are more differentiable datasets like \textit{gcredit}, represented in Fig.~\ref{fig:stem_chart_gcredit_cut}, which uniquely benefits from being modelled by a MultilayerPerceptron (P3).
This predictor is not ranked highly overall in Fig.~\ref{fig:f4_automl_predictor_rankings_per_dataset_heatmap} and Fig.~\ref{fig:f5_default_predictor_rankings_per_dataset_heatmap}, so solving \textit{gcredit} efficiently would require leveraging meta-knowledge for a similar dataset.
Given that the landmarkers in this work have failed to provide, with Table~\ref{tab:tab_dataset_landmarking_similarity} pointing unhelpfully to \textit{amazon}, \textit{gcredit} is an example of a dataset relying instead on the theoretical extreme of similarity, i.e.~its own oracular rankings.
It also exemplifies the all-or-nothing nature of $k=1$ strategies induced by the unreliability of predictor rankings.
In \textit{automl-meta}, \textit{gcredit} found NaiveBayes (P0) to be the best based on averaging single-fold evaluations.
Unfortunately, this particular ranking approach misses that, in \textit{automl-meta}, P0 achieves no better than a 10-fold CV evaluation of ${\sim}23\%$ loss, while P3 approaches $21\%$.
Consequently, \textit{O1-k1} fails to perform well, but the overall benefit of an oracle remains, as seen by the dominance of the bet-hedging \textit{O1-k4}.

Of course, the oracle is not always correct, given all the stochastic factors involved in compiling opportunistic meta-knowledge or leaning on default hyperparameters.
The \textit{abalone} dataset exemplifies a case in Fig.~\ref{fig:stem_chart_abalone_cut} where \textit{OX-kn} is beaten not because of statistical variations in testing strategies but because, for whatever reason, \textit{automl-meta} and \textit{default-meta} did not give due credit to RandomForest (P19).
This circumstance allows \textit{MX-k1} to outperform all.
Granted, \textit{automl-meta} did seemingly find better-performing solutions with MultilayerPerceptron (P3) and Bagging (P24), but these predictors are complicated to tune with HPO, and, unsurprisingly, strategy-assisted SMAC did not repeatedly achieve the same fortune in finding them.

Finally, while we have stressed that the landmarking approach in this work is only demonstrative and often fails, it has occasionally proved optimal, even though we caution it may have simply stumbled into success.
Specifically, while using the average of single-fold evaluations appears to disincentivise the use of SMO (P6) overall, Fig.~\ref{fig:f4_automl_predictor_rankings_per_dataset_heatmap} shows that the sparsely evaluated \textit{cifar10small} is one where SVM methods do well.
Thus, purely by virtue of landmarker-based similarity, catalogued in Table~\ref{tab:tab_dataset_landmarking_similarity}, Fig.~\ref{fig:stem_chart_dexter_cut} shows that SMAC is able to find a surprise optimum for \textit{dexter}.
Likewise, while Logistic (P2) and SGD (P4) do not top the \textit{automl-meta} and \textit{default-meta} leaderboards, respectively, and are unexceptional within corresponding oracular rankings for \textit{adult}, their high performances in Fig.~\ref{fig:stem_chart_adult_cut} are unlocked by landmarked association between \textit{adult} and both \textit{waveform} and \textit{dexter}.

The takeaway from all this is that each dataset has its own narrative.
Indeed, for completeness, while RandomForest and SVM methods optimise loss for the bulk of datasets, the remainder -- excluding \textit{kddcup} -- spotlight NaiveBayes (P0), NaiveBayesMultinomial (P1), Logistic (P2), MultilayerPerceptron (P3), SGD (P4), SimpleLogistic (P5), VotedPerceptron (P8), JRip (P12), AdaBoostM1 (P22), Bagging (P24), and RandomSubSpace (P28).
Each of these is uniquely optimal for only one dataset, barring JRip and RandomSubSpace, which optimise two and three ML problems, respectively.
Notably, many highly ranked predictors in a global leaderboard, e.g.~J48 and PART in \textit{automl-meta}, do not feature in this list of best performers.
Their expertise is possibly redundant, suggesting that, in future work, it may be worth exploring reduced search spaces consisting of \textit{complementary} predictors, not simply strong performers.
Regardless, these results show just how complicated it is, in pursuit of general applicability, to form a singular search-space reduction approach that accommodates every ML problem-solving nuance.

\section{Conclusion}
\label{sec:chap4_conclusion}

One of the current frontiers in data science is AutoML, characterised by the endeavour to design/implement complex systems that automate high-level ML operations~\citep{kemu20,doke21,khke22}.
At its core is the ongoing effort to construct mechanisms that efficiently search for high-performance ML pipelines, i.e.~optimally accurate solutions to -- typically -- predictive ML problems.
Such a procedure is traditionally slow and computationally expensive.
Hence, this work investigated one approach for upgrading model-selection mechanisms: exploiting opportunistically/systematically curated records of past performance on various datasets in order to reduce the selection pool of ML predictors when encountering a new problem.
Crucially, as an expansion of previous work~\citep{ngke21}, much attention was placed on how dataset characteristics influence both the reliability of this meta-knowledge and the accuracy of assessing search-space reduction strategies.

We first briefly reviewed modern efforts at managing an AutoML search space in Section~\ref{sec:chap4_related_work}, additionally contemplating the broader topic of benchmarking such approaches.
Within Section~\ref{sec:chap4_problem_formulation}, we then mathematically formulated the problem of constraining a complex pipeline-inclusive hyperparameter space to an allowable pool of ML predictors, enabling a potentially efficient but restrictive form of solution optimisation.
Finally, in Section~\ref{sec:chap4_methodology}, we described the methodology of how this work explores using meta-knowledge to guide such a constrained optimisation.
Specifically, employing the AutoML package AutoWeka4MCPS and its inbuilt Bayesian optimiser named SMAC, we detailed the generation of two meta-knowledge bases for a diverse selection of 20 datasets and 30 predictors.
They were (1) an opportunistic collation of CV-related single-fold evaluations for possibly pipelined predictors, named \textit{automl-meta}, and (2) a systematic curation of 10-fold CV evaluations for predictors with default hyperparameters, named \textit{default-meta}.
The opportunism for \textit{automl-meta} refers to the evaluations arising as a by-product of five previous two-hour SMAC runs per dataset; see Section~\ref{sec:chap4_settings} for hardware and dataset/predictor details.
Additionally, meta-knowledge curation involved averaging all raw evaluations of error rates per dataset and predictor to provide performance rankings for the 30 ML algorithms.
We then listed 33 strategies for search-space management, including three `controls': (1) \textit{baseline}, a standard search involving 30 predictors, (2) \textit{avatar}, a \textit{baseline} upgrade that efficiently skips invalid pipelines, and (3) \textit{r30}, an extensive 30-hour pipeline search followed by a reset pipeline-locked two-hour HPO.
The remaining 30 strategies were specified by the meta-knowledge base -- index $X=1$ denotes \textit{automl-meta} and $X=2$ denotes \textit{default-meta} -- that they used, the size $k=n$ that they reduced the predictor pool to, and their detailed source of rankings.
Specifically, when solving a given dataset, \textit{OX-kn} referred to using oracular rankings from that dataset, \textit{MX-kn} referred to rankings averaged across all 20 datasets, and \textit{LX-kn} referred to rankings from the most `similar' dataset.
Similarity in this work was demonstratively defined by rank-based correlations between datasets for five `quick-to-evaluate' landmarker predictors.

Our main conclusions from the results in Section~\ref{sec:chap4_experiment}, with code and data available online\footnote{https://github.com/UTS-CASLab/autoweka}, are as follows.
From Section~\ref{sec:chap4_exp_metaknowledge} on the nature of the meta-knowledge:
\begin{itemize}

\item \textbf{One cannot expect to understand all datasets/predictors equally}.
Every ML problem and algorithm has intrinsic characteristics and meta-features that should give pause or at least bring caution to undue extrapolation.
This seemingly obvious fact is often ignored when it comes to benchmarking, e.g.~when espousing the general applicability of a deep learning architecture based on performance for a few popular datasets.
However, it is also often overlooked in the compilation of meta-knowledge, even when many meta-learning approaches are well aware of dataset/predictor individuality while seeking ways to define similarity.
Specifically, the issue revolves around assuming uniform quality across a meta-knowledge base.
In \textit{automl-meta}, sampling different datasets with an equal amount of time and memory results in an unequal number of evaluations, providing more coverage for computational lightweights versus heavyweights.
For an HPO mechanism like SMAC, different predictors will hold varying appeal, likewise prioritising coverage for some over others.
Even in \textit{default-meta}, where the absence of time constraints allows a uniform sampling across datasets/predictors, fixed memory constraints preclude the successful evaluation of some computationally complex ML algorithms for heavyweight datasets.
More importantly, despite results showing that default hyperparameters, i.e.~expert knowledge, are reasonably well-tuned~\citep{wemu20}, it is unlikely that each solitary 10-fold CV sampling captures an equivalent representation of predictor performance.
Of course, from an oracular perspective, any knowledge about a dataset/predictor may be better than nothing for understanding that specific dataset/predictor.
However, for real-world approaches, this variability in quality threatens to distort both aggregate rankings and any similarity metrics based on predictor performance, i.e.~\textit{MX-kn} and \textit{LX-kn}, respectively.
Designers/users of a meta-knowledge base must be cognisant of this fact. 

\item \textbf{Despite a mix of stochastic influences, any sampling of meta-knowledge may still provide a useful estimate of the ground truth.}
This assertion arises from two results.
First, \textit{automl-meta} predictor rankings that did and did not account for pipelines were well correlated.
Second, ignoring differences in meta-predictor performance between \textit{automl-meta} and \textit{default-meta} -- meta-predictors are not prioritised by the SMAC-based application of reduction strategies anyway -- the remaining predictors were reasonably well correlated between both meta-knowledge bases.
These results suggest that, despite the noise and statistical variability inherent in compiling meta-knowledge, associated assertions that some predictors are better than others do not appear entirely random and spurious.
Thus, the premise of this research, i.e.~harvesting meta-knowledge for a ground truth, buried as it is, to inform search-space reduction, is not invalidated from the outset.
\end{itemize}

From Section~\ref{sec:chap4_dataset_understand} on the challenge of the datasets:
\begin{itemize}

\item \textbf{The quality of a meta-knowledge base or a benchmark depends on the informativeness of its constituents.}
Again, the full implications of this seemingly obvious statement are often underappreciated.
In context, the information that is required to \textit{guide} search-space control relates to which \textit{ML components} are significantly better than others.
The information that is required to \textit{assess} search-space control relates to which \textit{reduction strategies} are significantly better than others.
Any datasets for which predictor performances cluster around certain error rates are thus complicating factors when both guiding and assessing approaches for reducing an AutoML predictor pool.
In fact, with variations in technical definition, this work defined a concept called `challenge', where, given a fixed experimental setting, a dataset is `hard' or `easy' depending on whether few or many predictors excel at solving it.
Consequently, despite the diversity of the 20 datasets used in this investigation, incorporated for the sake of research continuity~\citep{fekl15,sabu18}, only three or four datasets were evaluated to have `hard' top-heavy distributions of predictor performance.
They would later be identified as ML problems for which reduction strategies must be most intelligently selective to be optimal.
In contrast, `easy' datasets are much more amenable to random culling and arbitrary strategies due to a higher chance that one of their multiple best performers ends up in the reduced predictor pool.
The decreased ability of such datasets to informatively distinguish between predictors, combined with a greater susceptibility to stochastic ranking permutations, can potentially weaken the power of aggregate meta-knowledge and hide useful trends.
Therefore, although the scope of this work did not allow for the robust differentiation of strategies according to dataset challenge, nor did it enable advice on how to manage `easy' datasets best, it reaffirms that meta-learning systems should not thoughtlessly accumulate datasets.
Understanding the characteristics and associated informativeness of an ML problem~\citep{muma18} is crucial to pursuing/maintaining high-quality meta-knowledge bases and benchmarking sets.

\end{itemize}

From Section~\ref{Sec:EvalOracle} on evaluating reduction strategies against oracular expectation:
\begin{itemize}

\item \textbf{Leveraging dataset similarity is a highly complicated affair.}
This assertion is not unexpected, as it refers to an unsolved core problem within the field of meta-learning, namely how to identify similar problems that may provide relevant information.
Successful association requires both an appropriate/rigorous metric for similarity and for that association to genuinely exist within the meta-knowledge base.
In contrast, the landmarking approach used in this work is rudimentary and demonstrative, and a collection of 20 ML problems is not guaranteed to comprise substantial similarities, so the odds are against landmarked reduction strategies performing well.
Conveniently, conclusions can be drawn even before assessing actual outcomes, simply by treating oracular rankings as perfect and then analysing the expected minimum/mean performance of landmarked strategies based on the predictor pool they would leave post-culling.
It was confirmed that, for several datasets, even random culling is expected to outperform reduction strategies that leverage this form of landmarking, regardless of whether \textit{automl-meta} or \textit{default-meta} provides the predictor rankings.
This result contrasts with expectations for strategies that lean on the global leaderboard.
For instance, the expected average performance of \textit{M1-kn} improves upon random selection in all cases.
Admittedly, later results highlight notable exceptions where \textit{LX-kn} does aid SMAC in locating an optimal ML pipeline, but these appear to be cases where landmarking stumbles into success.
The takeaway point is that global leaderboards of strong generalist predictors provide plenty of mileage for little effort; implementing sophisticated similarity-based procedures has a very high bar to meet in order to be justified.

\end{itemize}

From Section~\ref{sec:results} on evaluating reduction strategies for their true performance:
\begin{itemize}

\item \textbf{The consistency of performance for reduction strategies depends on the computational tractability of a dataset.}
This expected conclusion is confirmed upon actually running five two-hour SMAC runs per dataset for each of the 33 strategies.
In effect, the fewer evaluations a dataset received in \textit{automl-meta} or, due to memory constraints, \textit{default-meta}, the less likely it was for SMAC to converge to a consistent error rate.
The proportion of runs that failed to complete for a strategy, either due to memory issues or insufficient time to generate a 10-fold CV evaluation, was also correlated with dataset intractability.
Notably, the other most evident correlation was between the size of a reduced search space and the level of performance inconsistency/failure.
For many datasets, two hours on the hardware used in this work is not enough time to find a reliable optimum amongst a large selection of predictors, e.g.~for $k=19$ or $k=30$.

\item \textbf{Informed search-space reduction is effective as long as the culling is not extreme.}
This conclusion mirrors the finding in previous work~\citep{ngke21}.
If \textit{automl-meta} is considered the by-product of five $k=30$ \textit{avatar} runs, the fact that the best strategy applied to every dataset is competitive with the best \textit{automl-meta} predictor, usually substantially outperforming it, implies that the efficiency boost of search-space reduction can be very beneficial.
However, an overly severe reduction can exclude strong performers, and results show that $k=1$ strategies are bimodally all-or-nothing.
With the experimental context of this work in mind, the sweet spot for reduction appears to be around $4\leq k\leq 10$.

\item \textbf{On average, solving an ML problem receives the most benefit from prior experience with that exact ML problem, but relying on strong general performers is also a good option.}
Essentially, despite the high level of statistical uncertainty arising from many stochastic influences, this investigation found that \textit{O1-k4} and \textit{O1-k8} scored significantly better than \textit{baseline}, according to a Nemenyi test, and reduction strategies leveraging oracular rankings performed the best on average.
Given that the oracle is an idealised limit of similarity-based meta-knowledge, i.e.~it is hoped that a sufficiently sized meta-knowledge base may contain similar datasets where predictor performances are ranked equivalently, there is certainly scope for an advanced meta-learning system to one day supplant the utility of a global leaderboard.
Nonetheless, it is worth stressing that a perfect oracle is a naive assumption, especially based on non-exhaustive meta-knowledge generation, and the extreme variability in \textit{O1-k1} performance attests to this; the strategy rarely fails because the oracle is generally ill-suited to a dataset, failing instead because an inaccurate ranking for a singular predictor can have a dramatic impact.
On the whole, such stochastic variations -- the `granularity' of oracular rankings is much more susceptible to noise -- are suppressed with a slightly larger $k$.
In contrast, the runner-up strategy type, i.e.~\textit{MX-kn}, is much more robust due to its averaging across 20 datasets.
When \textit{M1-k1} fails, it is often because the global leaderboard itself does not represent a dataset well.

\item \textbf{On balance, neither the opportunistic or systematic curation of meta-knowledge proves more valuable than the other.}
This unintuitive result must be approached with caution and perhaps further research.
Certainly, both \textit{automl-meta} and \textit{default-meta} have differing pros and cons.
For instance, the opportunistic harvest of meta-knowledge was compiled under the same experimental settings as the subsequent SMAC-based testing of reduction strategies.
Accordingly, unlike time-unconstrained \textit{default-meta}, it does not unhelpfully suggest evaluative outcomes for ML algorithms that cannot be replicated.
On the other hand, compiling \textit{automl-meta} provides no control for its number of predictor evaluations, whether a pipeline of preprocessors is tested, and so on.
Arguably, the way rankings are generated also matters.
In this work, predictor rankings for \textit{automl-meta} were based on an average of single-fold evaluations, which leaves their representational quality subject to how SMAC may have explored hyperparameter spaces.
It has not been investigated what happens when rankings are based on the best single-fold evaluations; these presumably represent optimal predictor performances better but, in doing so, ignore how unlikely and thus unrepeatable it is for SMAC to reach those optima.
Likewise, if \textit{automl-meta} only ranked predictors with a completed 10-fold CV evaluation, truly mirroring the assessment of reduction strategies, it is unclear what the impact of such a sparser but more reliable sampling would be.
As such, it is presently unclear how finely balanced the utility of both meta-knowledge compilation methods is.
Specifically, does the ground truth buried in both meta-knowledge bases ultimately come through with relative consistency despite all the surface-level statistical noise?
Or is this but a case of coincidence?

\item \textbf{A few predictors optimally solve many ML problems. Many predictors optimally solve a few ML problems.}
Specifically, $60\%$ of the studied datasets were optimally solved by reduction strategies that supported efficient optimisation paths to powerful universal approximators, i.e.~RandomForest and, with an appropriate kernel, SMO.
Almost without exception, the remaining datasets each had a unique preferred predictor, and these were often ML algorithms that were fast to compute within the constraints of the experimental settings.
These results indicate just how challenging it is to construct a mechanism for AutoML search-space management that universally caters to every nuanced ML problem that may be encountered.

\end{itemize}

Ultimately, while this work has teased out performance patterns for varying approaches to configuration-space reduction, there is plenty of scope for future research.
Statistical noise continues to be a fundamental issue, and it may be that future work should compile a much larger sample for meta-knowledge, perhaps drawn from ML pipeline evaluations on OpenML~\citep{vava14}.
Such an investigation may be a true test of how far opportunism can be taken, given the range of computational constraints associated with each evaluation.
In conjunction with this more extensive collection of datasets, there is also the potential to implement and test whether more sophisticated similarity-based methods can start to outperform strategies based on a global leaderboard.
Alternatively, another open question based on the conclusions of this work is whether predictors should actually be ranked based on whether they bring something new to the table, i.e.~whether they complement previous methods rather than whether they are simply overall solid performers.
Avoiding redundant ML components within a predictor pool may significantly speed up model selection and HPO without sacrificing nuanced ML problems.
Finally, while this research has focussed on static meta-knowledge bases, contexts behind real-world ML applications can evolve over time.
For instance, a dynamic data stream may resemble different datasets in a meta-knowledge base at varying times or, even for a static dataset, associations with historically encountered ML problems may become more apparent as AutoML evaluations continue.
Adaptively reconstructing an ML pipeline in both cases will benefit from intelligently managing a searchable predictor pool.
In short, this research has valuable insights to glean, and these are likely to inform the next wave of pipeline composition/optimisation improvements: \textit{dynamic} configuration-space control.

\section*{Acknowledgements}
This research is sponsored by CASLab, University of Technology Sydney (UTS).

\bibliography{references}

\end{document}